%% file: article.tex
\documentclass{article}

\usepackage{arxiv}

\usepackage[utf8]{inputenc} 
\usepackage[T1]{fontenc}    
\usepackage{hyperref}       
\usepackage{url}            
\usepackage{booktabs}       
\usepackage{amsfonts}       
\usepackage{amsrefs}       
\usepackage{amsmath}
\usepackage{nicefrac}       
\usepackage{microtype}      
\usepackage{lipsum}
\usepackage{listings}
\usepackage{graphicx}
\usepackage{textcomp}
\usepackage{soul}
\usepackage{mathtools,eqparbox}
\usepackage[dvipsnames]{xcolor}
\usepackage{tikz}
\usepackage{tabularx}
\usepackage{caption} 
\usepackage{dsfont} 
\usepackage{hhline}
\usepackage[figuresleft]{rotating}
\usepackage[section]{placeins}

\usepackage{lscape}

\usetikzlibrary{positioning}
\newcommand*\samethanks[1][\value{footnote}]{\footnotemark[#1]}
\renewcommand{\vec}[1]{\mathbf{#1}}

\title{What Matters in Reinforcement Learning for Tractography}
\author{
  Antoine Théberge \\
  Faculté des Sciences\\
  Université de Sherbrooke\\
  Sherbrooke, QC, CA, J1K 2R1 \\
  \texttt{antoine.theberge@usherbrooke.ca} \\
  
  \And
  
  Christian Desrosiers \\
  Département de génie logiciel et des TI\\
  École de technologie supérieure \\
  Montréal, QC, CA, H3C 1K3 \\
  \texttt{christian.desrosiers@etsmtl.ca} \\
    
  \And
    
  Maxime Descoteaux\thanks{Equal contributions, Co-senior authors} \\
  Faculté des Sciences\\
  Université de Sherbrooke\\
  Sherbrooke, QC, CA, J1K 2R1 \\
  \texttt{maxime.descoteaux@usherbrooke.ca} \\
  
  \And 
  
  Pierre-Marc Jodoin\samethanks \\
  Faculté des Sciences\\
  Université de Sherbrooke\\
  Sherbrooke, QC, CA, J1K 2R1 \\
  \texttt{pierre-marc.jodoin@usherbrooke.ca} \\

}

\begin{document}
\maketitle

\begin{abstract}
    Recently, deep reinforcement learning (RL) has been proposed to learn the tractography procedure and train agents to reconstruct the structure of the white matter without manually curated reference streamlines. While the performances reported were competitive, the proposed framework is complex, and little is still known about the role and impact of its multiple parts. In this work, we thoroughly explore the different components of the proposed framework, such as the choice of the RL algorithm, seeding strategy, the input signal and reward function, and shed light on their impact. Approximately 7,400 models were trained for this work, totalling nearly 41,000 hours of GPU time. Our goal is to guide researchers eager to explore the possibilities of deep RL for tractography by exposing what works and what does not work with the category of approach. As such, we ultimately propose a series of recommendations concerning the choice of RL algorithm, the input to the agents, the reward function and more to help future work using reinforcement learning for tractography. We also release the open source codebase, trained models, and datasets for users and researchers wanting to explore reinforcement learning for tractography.  
\end{abstract}

\vspace{-0.3cm}
\section{Introduction}
\vspace{-0.3cm}
\label{sec:intro}

    Tractography allows the user to reconstruct virtual connections in the from diffusion magnetic resonance~(dMR) images in the form of streamlines. Reconstructing enough streamlines provides an approximation of the architecture of white matter tissue. To do so, initial points are distributed in a mask representing white matter or the interface between grey and white matter. Then, steps are taken following directions extracted from local models of the dMR signal. Stepping in the white matter volume brings the tractography algorithm to a new point, where the process can be repeated until a termination criterion is met~\cite{basser_vivo_2000}. Since its inception, countless algorithms have been proposed, improving how directions are followed~\cites{fillard2009, girard2014}, which termination criterion are used~\cite{smith2012} or how the local model is computed~\cites{malcolm2010, rheault2019}.
    
    However, despite these advances, tractography is still plagued by several known limitations~\cite{rheault_common_2020}. In fact, tractography has been described as an \emph{ill-posed problem} as it attempts to recover global connectivity from local information~\cite{maier-hein_2017}. Global tractography algorithms, which minimize a global loss by aligning local directions with each other~\cite{fillard2009}, have been proposed, but have yet to solve the problem.
    
    Supervised machine learning~(ML) tractography algorithms have been developed as a way to circumvent the problems classical tractography methods have been suffering from.  Given a set of reference (aka {\em gold standard}) streamlines and associated diffusion MR images, ML methods are trained to learn the tractography procedure directly from raw diffusion data and, because they are shown the ``correct'' direction each step of the way, can hopefully outpace classical tractography algorithms. Their learning regime allows them to formulate a new, more appropriate local model~\cite{benou2019} or to propagate the streamlines directly~\cites{neher_machine_2015, neher_fiber_2017, poulin_learn_2017, wegmayr2018, wegmayr2021entrack}.
     
    Although supervised ML tractography algorithms have demonstrated superior reconstruction performance than their classical counterparts, they are severely limited by their training data. Training diffusion MR images with their reference streamlines can come from synthetic phantoms, such as the FiberCup~\cites{poupon2008, poupon2010, fillard2011} or the ISMRM 2015 WM Challenge dataset~\cite{maier-hein_2017} which, from their synthetic nature, do not fully represent real human brains. Reference streamlines can also come from semiautomatic clustering methods and / or manual segmentation performed by trained neuroanatomists. However, streamline segmentation is a time-consuming and highly variable process~\cites{rheault_tractostorm_2020, schilling2020} that may affect the resulting tractograms. Moreover, these tractograms are still generated by classical tractography algorithms which unavoidably lead to imperfect gold standard tractograms.
     
    A way to leverage the expressiveness of machine learning models for tractography without the need for reference streamlines is through reinforcement learning (RL) as recently proposed by Theberge {\em et al.}~\cite{theberge2021}. Their proposed {\em Track-to-Learn} is a framework whose agents learn by taking actions which influence the state of an environment and which, in return, rewards the agents if the actions taken propagate streamlines correctly. Agents trained with this procedure demonstrated highly competitive performance compared to their supervised and classical alternatives on \emph{in-silico} and \emph{in-vivo} datasets. 
     
    {\em Track-to-Learn} is the only fully operational deep-RL tractography method available today.  In our original paper, we reported results for only two RL methods, one seeding strategy, one signal representation, and one reward function. This limited exploration makes it unclear what contributes to the performance of the method. 
   
     
    In this work, we propose investigating different formulations of RL for tractography and shed light on their impacts. Specifically:
    
    \begin{enumerate}
        \item We provide a clear set of pitfalls one shall avoid in designing reinforcement learning methods for tractography as well as clear recommendations to train agents with the Track-to-Learn framework;
        \item We evaluate the performance of multiple RL algorithms applied to tractography;
        \item We reformulate {\em Track-to-Learn} to bring the framework closer to the standard RL context;
        \item We evaluate the performance of agents when using different input signals to measure the impact on reconstruction performance;
        \item We evaluate the reconstruction done by agents when trained using different reward functions to evaluate how suited they are for tractography;
        \item We release the complete codebase~\footnote{\url{https://github.com/scil-vital/TrackToLearn}}, datasets~\footnote{\url{https://zenodo.org/record/7853832}} and trained models~\footnote{\url{https://zenodo.org/record/7853590}} for researchers wanting to use the results of this work, or further analyze or improve upon the framework.
    \end{enumerate}

    The results of this investigation should provide strong foundations for researchers who want to use reinforcement learning for tractography or improve upon the Track-to-Learn framework.
     
   \section{Related Works}
        \label{sec:related}
        
        To the best of our knowledge, the first method to apply reinforcement learning (in its non-deep flavour) was presented at ISBI 2018 by Wanyan et al.~\cite{wanyan2018}. The method uses multiple agents to build expanding graphs that represent paths between the beginning and ending nodes of a bundle. Temporal Difference (TD) learning~\cite{sutton2018} is used to evaluate the nodes of the graphs, where agents are rewarded for successfully connecting the beginning and ending regions of a bundle. Streamlines can then be obtained by using the learnt policy to traverse the graph. Unfortunately, this method assumes that the starting and ending regions are provided, which is rarely true. It is also unclear how the graph was propagated and, most importantly, how it defines spatial positions as states (nodes).
        
        Recently, Track-to-Learn~\cite{theberge2021} was proposed as the first fully operational deep-RL tractography method. At the heart of the framework is a neural-network-driven agent which navigates in the brain like a robot in a maze and whose path along its journey forms a streamline. For Track-to-Learn,  the diffusion signal is the state space, and the reward function is the alignment between propagated streamline segments and the underlying peaks computed from the MR signal, as well as the alignment between streamline segments. This formulation allows agents to learn tractography independently of the underlying anatomy and thus generalise well to new datasets with competitive performance against state-of-the-art tractography algorithms.
        
        %
        
        Tractography can be seen as similar to the robotic control problem as defined in the Roboschool~(or its open-source successor, PyBullet Gymperium~\cite{benelot2018})~set of environments available through the OpenAI Gym software~\cite{brockman2016}. The suite of environments, where a set of robots (some taking a humanoid form) can be trained to accomplish a set of tasks (for example, learning to run or chase a moving waypoint), often formulates the state-space as the configuration of the joints of the robot and the actions as the torque applied to the joints. Similarly, one could picture tractography as a robot navigating the white matter, obtaining the current diffusion signal at its position and taking a step in a direction as a result. 

        Recent works have examined the intricacies of reinforcement learning for robotic control. For example, a recent study~\cite{andrychowicz2020matters} investigated the various formulations of on-policy reinforcement learning algorithms~(c.f. section~\ref{sec:rl}) to provide insight and practical recommendations on their training. A similar study was conducted to examine the training of reinforcement learning agents from human demonstrations~\cite{mandlekar2021matters}, where various RL algorithms, state formulations and hyperparameters were considered to guide future research on the topic.

\section{Method}
\label{sec:tractorl}

    \subsection{Preliminaries}
    \label{sec:prelim}
    
        To properly introduce the necessary concepts, here we present a brief overview of tractography and reinforcement learning and how they relate to each other.
    
        \subsubsection{Tractography}
        \label{sec:tractography}

            Tractography is the process of iteratively reconstructing white matter (WM) streamlines from diffusion MR images. From a starting position $p_0$, called a \emph{seed}, a step size $\Delta$, and a direction given a local model $v(p_0)$ of the diffusion MRI signal at the initial position, a ``forward'' step is taken to propagate the streamline further in the WM: 
            \begin{equation}
                p_{t+1} = p_t + \Delta v(p_t), \;\;\;\; t \in 0..M,
            \end{equation}           
            where $M$ is determined by the maximum streamline length or when a termination criterion is met. The set of points $p_{0..M}$ produces a \emph{half-streamline}. Because the diffusion signal is symmetric, the tracking procedure needs to be performed ``backward'' as well:
            \begin{equation}
                p_{t+1} = p_t - \Delta v(p_t), \;\;\;\; t \in 0..N,
            \end{equation}
            where $N$ is determined like $M$. This process produces another \emph{half-streamline} which is concatenated to the first half to produce a complete streamline $p_{N..0..M}$ in the WM~\cites{basser1994, basser_vivo_2000}. 
            
            Tracking can be initiated in the WM, where axonal fibres reside, or at the interface of the WM and grey matter (GM), where axonal fibres are known to emanate~\cite{girard2014}.

            Over the years, several improvements have been proposed to the tractography procedure. The Particle Filtering Tractography~(PFT)~\cite{girard2014} algorithm, for example, samples heuristic maps computed from partial volume maps to find the optimal path for streamlines and may ``backtrack'' if the streamline terminated early.
            
        \subsubsection{Reinforcement Learning}\label{sec:rl}
        
            \begin{figure}
                \centering
                \includegraphics[width=0.65\textwidth]{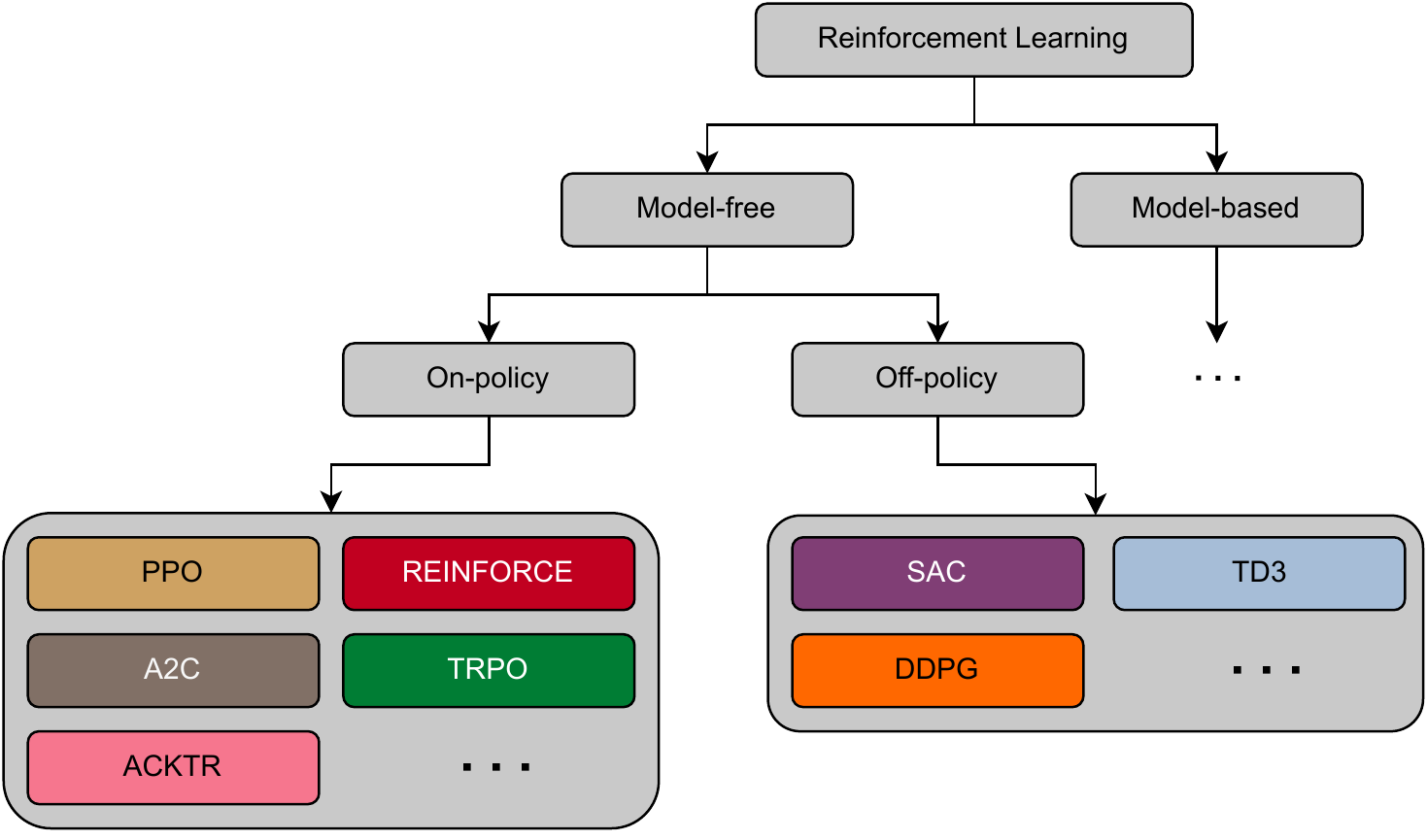}
                \caption{Taxonomy of some Reinforcement Learning algorithms, all of which are considered in this work.}
                \label{fig:rl_taxonomy}
            \end{figure}
            
            Reinforcement Learning~(RL) is the process of learning a policy $\pi$ that dictates the action $a$ taken by an agent based on the state $s$ of its environment.  In return, the environment provides the agent with a reward $r$, and the goal of the agent is to maximise its expected sum of future rewards which corresponds to solving the underlying goal.
            
            Formally, the reinforcement learning framework is based on the Markov Decision Process mathematical framework, a 4-tuple composed of $(S, A, p, R)$ where $s \in S$ is the set of all possible states, $a \in A$ is the set of all actions, $p(s'|s,a)$ defines the probability of going from state $s$ to state $s'$ by executing action $a$, and $r \in R(s,a)$ defines the reward obtained from executing action $a$ in state $s$. The subscript $\boldsymbol{\cdot}_t$ denotes the {\em timestep} at which an action $a_t$ is taken, a state $s_t$ is encountered, and a reward $r_t$ is obtained. As for an {\em episode} $\tau = (s_0, a_0, r_0, s_1, a_1, r_1, ..., s_T, a_T, r_T)$, it denotes the list of states, actions, and rewards encountered from the moment the environment is reset at timestep $0$ until a stopping criterion is encountered at timestep $T$. 
            
            The goal of the learning agent is to maximize the RL objective, i.e. the overall expected reward:
            \begin{equation}
             \label{eq:rl_objective}
                J \, = \, \mathbb{E}_{s_t,a_t \sim \pi}[R(s_t,a_t)], \; \forall t \in 0..T,
            \end{equation}
            and to learn the optimal policy $\pi$ which will maximize the above objective:
             \begin{equation}
             \label{eq:rl_policy}
                \pi^* \, = \, \operatorname*{argmax}_\pi \, J, \; \forall s_t, \; \forall a_t \sim \pi(s_t), \; \forall t \in 0..T.
            \end{equation} 

            To learn which action is most desirable, the \emph{state value function} $V^\pi$ (often referred to as the \emph{value function}) can be trained to evaluate the expected return $G$ at each state $s_t$ according to the behaviour of the policy $\pi$:
            \begin{equation}
                V^\pi(s_t) \, = \, \sum_{t}^{T} \mathbb{E}_{s_t,a_t \sim \pi}[\gamma^{T-t} r_t],
            \end{equation}
            where the discount parameter $\gamma$ controls the greediness of the agent and prevents the expected sum of rewards from growing to infinity. In the same vein, the \emph{state-action value function} $Q^\pi$ (often referred to as the $Q$-function) can be trained to evaluate the expected return by executing action $a_t$ at state $s_t$ and then following the policy $\pi$ for subsequent states:
            \begin{equation}
                Q^{\pi}(s_t,a_t) \, = \, r_t + {\gamma} V^\pi(s_{t+1}).
            \end{equation}

            Since its inception, many algorithms have been developed to learn the optimal policy defined in Eq.~(\ref{eq:rl_policy}). These can be classified into two broad categories: model-based and model-free. Model-based algorithms learn a model of the environment and plan the optimal course of action using the learnt model. On the other hand, model-free algorithms directly learn a policy to maximise the objective described in~(\ref{eq:rl_objective}). Model-free algorithms can be further classified into \emph{on-policy} and \emph{off-policy} algorithms: on-policy algorithms will typically learn as they go, training their policy on all newly encountered tuples of $(s_t, a_t, r_t)$, whereas off-policy algorithms will instead store all the encountered experience in a buffer, which will be uniformly sampled at training time. Figure~\ref{fig:rl_taxonomy} provides an overview of some RL algorithms and their categories.

        \subsubsection{Track-to-Learn}
        \label{sec:tracktolearn}

            \begin{figure}
                \centering
                \includegraphics[width=1.\textwidth]{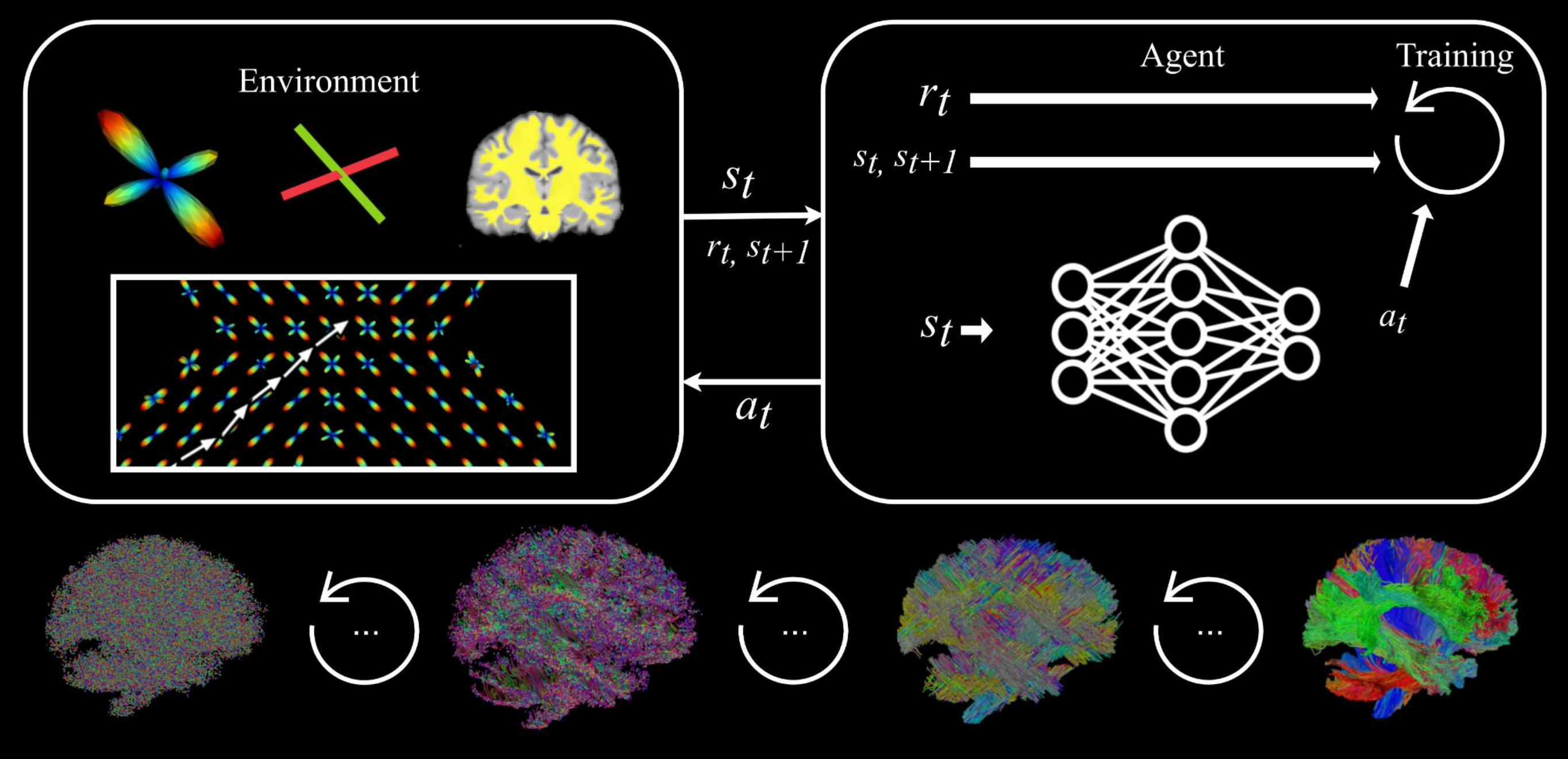}
                \caption{The Track-to-Learn framework. The environment sends the state $s_t$ (i.e. fODFs at the streamline position $p_t$) to the learning agent, who produces an action $a_t$ (i.e., a 3D vector representing the tracking direction). The action $a_t$ is sent to the environment, which propagates the streamline to a new point $p_{t+1}$, and returns a new state $s_{t+1}$ as well as a reward $r_t$ associated with the last pair of input signal and action. The stepping process repeats until a streamline is tracked, where it starts over until learning is complete.}
                \label{fig:tracktolearn}
            \end{figure}
        
            Track-to-Learn\cites{theberge2021} is a tractography-orientated deep-RL method that serves as the cornerstone of this work. Its overarching goal is to learn the tractography procedure without a pre-annotated dataset as in most other ML methods. RL agents are trained to navigate in a diffusion MRI volume and reconstruct tractograms. The proposed framework requires a signal volume (e.g., the raw diffusion-weighted signal or the fODFs), a tracking mask and a peak volume (e.g., the local maxima of the fODF glyphs). The input diffusion signal acts as the state-space, and the environment provides the signal at the tip of the streamline as the state. To help the agents better navigate the WM, the signal from the six immediate neighbours (up, down, left, right, front, back) as well as the WM mask value at the streamline's tip and its neighbours are also appended to the state. Finally, to add directional information to the signal, the last four streamline segments are appended to the state.
            
            In return, the agent produces an action that represents the orientation of the tracking step. The environment first takes the action and scales it to the tracking step size to obtain a new streamline segment $\vec{u}_t$: 
            \begin{equation}
                \vec{u}_t = \, \Delta\frac{\vec{a_t}}{||\vec{a_t}||}.   
            \end{equation}
            The new streamline segment is used to propagate the streamline and obtain $p_{t+1}$, which can be used by the environment to produce $s_{t+1}$. The environment then computes the reward as:
           \begin{eqnarray}
           \label{eq:ttl_reward}
                r_t \, = \, |\max_{\vec{v}(p_t)} \, \langle \vec{v}(p_t), \vec{u}_{t} \rangle| \cdot \langle \vec{u}_t, \vec{u}_{t-1}\rangle,
            \end{eqnarray}
            where $\vec{v}(p_t)$ are the fODF peaks at position $p_t$ and $\langle \cdot \rangle$ is the dot product between two normed vectors.
            
            Episodes are defined as the tracking of a streamline and tracking stops if one of the following conditions occurs: 1) tracking goes out of the tracking mask, 2) the angle between the current tracking step and the last streamline segment is too high, 3) the streamline is too long, or 4) the cumulative angle between streamline segments exceeds a threshold. The framework allows any RL algorithm to be used to train the agents.
            Unlike the standard RL framework, the formulation in \cite{theberge2021} uses two environments: one to perform  \emph{forward} tracking, and the other to perform \emph{backward} tracking. While forward tracking works as described in Section \ref{sec:tractography}, the \emph{ backward} tracking starts from the end of the half-streamline, which is \emph{retracked} until the tracking arrives at the original seed, where it can begin to track freely in the other direction. To preserve the integrity of the half-streamline computed during the forward phase, actions computed by the agents during retracking do not replace the half-streamline segments but are still rewarded and used for training. Figure~\ref{fig:tracktolearn} presents an overview of the method.
            
            The framework was implemented with two RL algorithms, namely TD3~\cite{fujimoto2018} and SAC~\cite{haarnoja2018a}. A concatenation of the fODFs and the WM mask was used as input signal, and the peaks were computed from the fODF volume. To help better fill the WM volume, Gaussian noise proportional to the fractional anisotropy at the tip of the streamline was added to the actions output by the agents. All neural networks involved were implemented as two-layer fully connected nets with either ReLU or Tanh activations and a width of 1024.

            Agents trained with this method exhibit performances comparable to competing supervised-learning tractography algorithms which are trained using reference streamlines. Track-to-Learn agents also show excellent generalisation to \emph{unseen} datasets. However, Track-to-Learn agents tend to reconstruct many anatomically implausible bundles and produce tractograms that generally have lower overlap to reference bundles than supervised or classical methods.
    
    \subsection{Benchmarking \& Experiments}
    \label{sec:benchmarks}
    
        In this section, we present the datasets used as well as performed experiments.
    
        \subsubsection{Datasets}
            \paragraph*{FiberCup}
            
                The synthetic FiberCup dataset~\cites{neher_machine_2015, neher_fiber_2017} is a recreation of the original physical FiberCup phantom acquisitions~\cites{poupon2008, poupon2010, fillard2011} using the Fiberfox tool~\cite{neher2014}. The phantom features challenging fiber configurations such as crossings, kissings and fannings. The dataset features a 64\,$\times$\,64 matrix grid of 3mm isometric voxels over three axial slices like the original dataset. The synthetic ``acquisition'' is done at a b-value of 1000 s/mm$^2$ over 30 gradient directions and features no artifacts, giving an estimated SNR around 40. No further post-processing is performed as the dataset is already free of artifacts.
                
                To ensure that the learning agents do not overfit the training dataset, we propose a coronally and sagitally flipped version of the synthetic FiberCup described above, which we dub ``Flipped''. The dataset therefore features the same diffusion signal and the same fiber configurations, but rearranged differently.
            
            \paragraph*{ISMRM2015}
                
                The ISMRM2015 dataset~\cite{maier-hein_2017} is a synthetic virtual phantom mimicking a real human brain. The dataset was generated by manually segmenting 25 bundles from a collection of subjects from the Human-Connectome Project dataset~\cite{glasser2016} using definitions by Stieltjes et al.~\cite{stieltjes2013}, and then generating the diffusion volume and a structural T1 image using Fiberfox~\cite{neher2014}. The phantom features a 2mm isometric diffusion volume generated with 32 gradient directions at a b-value of 1000 s/mm$^2$. Artifacts are added by the creators of the dataset to make it resemble clinical data. 
                
                The ground-truth WM, GM and CSF masks are used for the experiments.
                
            %
            %
            
            \paragraph{TractoInferno}
                The TractoInferno dataset~\cite{poulin2022tractoinferno} is a collection of 284 manually quality-controlled \emph{in-vivo} acquisitions from multiple sites, each including T1-weighted images and diffusion data, which was then processed through TractoFlow\cite{theaud2020} to denoise the DWI, resample the diffusion resolution to 1mm isometric, segment the tissues and compute the fODF volume. Each acquisition includes reference-curated streamlines computed from an ensemble of classical tractography algorithms, which can be used to compare new tractograms on the same subject. For this project, we one subject of the dataset (ID: 1006) which is the same that was used by the authors of TractoInferno to display qualitative results.
                
            For every dataset, the fODFs and derived metrics (such as fODF peaks) are computed using the \texttt{descoteaux07} basis~\cite{descoteaux2007} at order 6 using scilpy\footnote{\url{https://github.com/scilus/scilpy}}.
             
        \subsubsection{Evaluation metrics}
        \label{sec:metrics}
        
            We use the Tractometer tool to evaluate the performance of trained agents on the FiberCup, Flipped and ISMRM2015 datasets. The Tractometer is a tool which allows the user to score generated tractograms against ground bundles. The Tractometer reports the rate of valid-connections~(VC) streamlines (connecting two regions that should be connected), the rate of invalid-connection~(IC) streamlines (connecting two regions that should not be connected) and the rate of no-connections~(NC) streamlines (not connecting two regions). Valid and invalid connections are grouped into valid and invalid bundles~(VB, IB), which are also reported. The Tractometer also reports voxel-wise measures such as the overlap~(OL), i.e. the proportion of ground-truth voxels that have at least one streamlines passing through them, and the overreach~(OR), i.e. the fraction of voxels which have at least one streamline passing through them but should not. Finally, we report the F1 score, which is the voxel-wise Dice coefficient between ground-truth and reconstructed bundles.
            

            To evaluate the generalization performances of trained agents on the TractoInferno dataset, we use the TractoEval pipeline~\footnote{~\url{https://github.com/scil-vital/TractoInferno/}} to compute OL and OR, as well as the Dice with respect to the provided reference streamlines.
        
        \subsubsection{Training}
            As opposed to the original \emph{Track-to-Learn} training procedure where the best episode was chosen, all agents are now trained for 1000 episodes with no early stopping. An episode consists of tracking $n$ streamlines until a termination criterion is met for each of them. In all experiments, $n$ is set to $4096$. A grid search, per algorithm, per dataset and per experiment, over the learning rate $\eta$, discount factor $\gamma$ as well as algorithm-specific hyperparameters is performed to determine the best set of hyperparameters for each agent. The hyperparameters are chosen according to which set gave the best VC rate. If multiple sets give similar VC rates, the OL is used as tie-breaker. Once the best hyperparameters are found, the agents are trained using five different random seeds. Unless specified, the rest of the training procedure and implementation details are identical to those in \cite{theberge2021}. 
            
             Agents are trained on several types of computers equipped with different GPUs, such as a desktop 1080Ti GPU, an RTX8000 GPU (with GNU Parallel~\cite{tange2011a}) or nodes of the Compute Canada clusters equiped with V100 GPUs.
            
        \subsubsection{Tracking parameters}
            Unless specified, the tracking parameters are the same for all experiments and agents. Agents are trained using 10 seeds per voxel on the FiberCup and 2 seeds per voxel on the ISMRM2015 datasets. At test time, tracking is performed at 33 seeds per voxel on the FiberCup and Flipped datasets, 7 seeds per voxel on the ISMRM2015 dataset. We initialize tracking at 10 seeds per voxel on the TractoInferno dataset to compare ourselves to the results presented in ~\cite{poulin2022tractoinferno}. The tracking step size is set to 0.75mm on the FiberCup and ISMRM2015 datasets and 0.375mm on the TractoInferno dataset (c.f. section~\ref{sub:voxel_vs_step}). Streamlines longer than 200mm and shorter than 20mm are discarded at test time.

        \subsubsection{Experiment 1: Agents}

            Since the rise in popularity of deep RL, a plethora of algorithms based on this approach have been proposed, each with their own improvements and drawbacks. In the first experiment, we  benchmark the performance of many algorithms and determine which are best suited for tractography. The algorithms considered are REINFORCE~\cite{williams1992} (also known as Vanilla Policy Gradient, VPG), Advantage Actor-Critic (A2C)~\cite{mnih2016}, Trust-Region Policy Optimization (TRPO)~\cite{schulman2015trust}, Actor-Critic using Kronecker-Factored Trust Region (ACKTR)~\cite{wu2017}, Proximal Policy-Optimization (PPO)~\cite{schulman2017a}, Deep Deterministic Policy Gradient~(DDPG)~\cite{lillicrap2015}, Twin Delayed Deep Deterministic policy gradient~(TD3)~\cite{fujimoto2018}, Soft Actor-Critic~(SAC)~\cite{haarnoja2018a} and its variant with automatic entropy tuning~\cite{haarnoja2018b} which we dub SAC Auto. A description of the implemented algorithms is available in Appendix \ref{app:agents}. We perform a hyperparameter sweep for all algorithms to ensure the best possible performance  (all hyperparameter ranges are available in Appendix \ref{app:exp1_hyperparameters}). Once the best hyperparameters are found, we again train the agents using five random seeds and report average results.
            
            To assess the generalization capabilities of our agents, we also perform tracking on the TractoInferno dataset using our agents trained on the ISMRM2015 dataset. Whole-brain tractograms are segmented using RecoBundlesX~\cites{garyfallidis2018recognition, rheault2020analyse} using the provided ReconBundlesX atlas~\cite{rheault2021atlas}.
            
            To provide a baseline performance, we compare our trained agents to the PFT algorithm. The PFT algorithm was initialized using the same tracking parameters as the RL agents: 0.75mm step size (0.375mm on the TractoInferno dataset), WM seeding, streamlines shorter than 20mm or longer than 200mm are discarded, 33 seeds per voxel on the FiberCup and Flipped datasets, 7 seeds per voxel on the ISMRM2015 dataset and 10 seeds per voxel on the TractoInferno dataset).
            
        \subsubsection{Experiment 2: Seeding}
            
            The two-environment context (c.f. subsection~\ref{sec:tracktolearn}) that WM seeding imposes could be a source of instability for the training procedure and prohibits the usage of established RL algorithms implementations, limiting the ease of development. As such, in this experiment, we explore the impact of the seeding strategy on the training procedure and reconstructed tractograms.
            
            We adapt the Track-to-Learn framework to allow for seeding from the WM-GM interface, removing the need for the \emph{backwards} environment. In this formulation, only one environment is needed, which will produce complete streamlines. To help the learning procedure, we flip the first action if it immediately goes out of the tracking mask, akin to what was previously proposed in ~\cite{wegmayr2018}. To thoroughly compare the impacts of interface seeding versus WM seeding, we again perform a hyperparameter sweep for all algorithms to ensure the highest possible performance. Once the best hyperparameters are found, we again train the agents using five random seeds and report average results.
            
            We perform training on the FiberCup and the ISMRM2015 datasets and report results on the FiberCup, Flipped and ISMRM2015 datasets. Because interface masks typically contains fewer voxels than WM masks, agents are trained with 100 seeds per voxel on the FiberCup dataset and 20 seeds per voxel on the ISMRM2015 dataset to obtain roughly the same number of seeds. Testing was performed at 300 seeds per voxel on the FiberCup and Flipped datasets, 60 seeds per voxel on the ISMRM2015 dataset. The tracking step size was set to 0.75mm. Streamlines longer than 200mm and shorter than 20mm are discarded at test time. We again compare ourselves to the PFT algorithm, initialized to match the parameters of our agents. 
            
            As for the previous experiment, we perform tracking on the TractoInferno dataset to evaluate the generalization capabilities of our agents trained on the ISMRM2015 dataset. Tracking was initialized from the interface at 20 seeds per voxel with a step-size of 0.375mm for all agents and algorithms, and streamlines shorter than 20mm and longer than 200mm are discarded.

        \subsubsection{Experiment 3: Retracking}

            \begin{figure}[thbp]
                \centering
                \includegraphics[width=1.\textwidth]{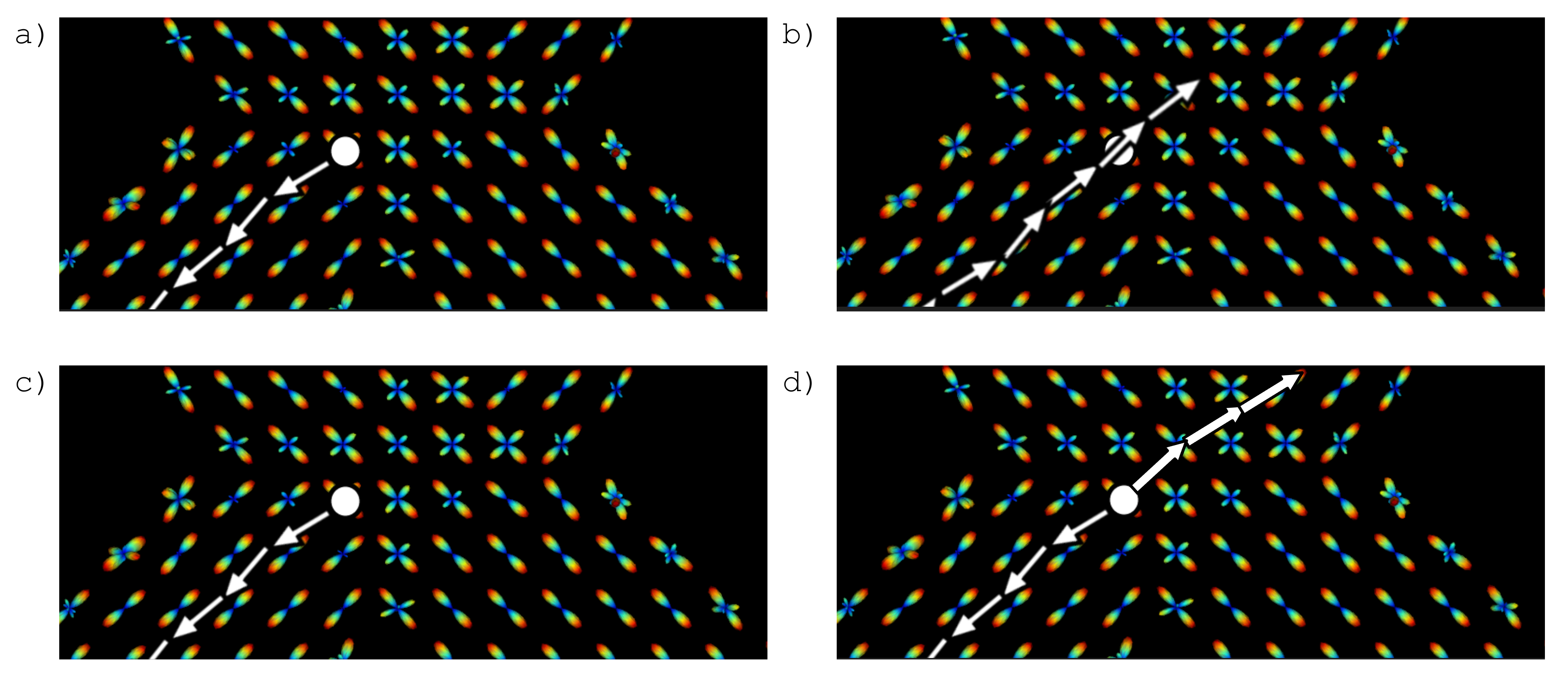}
                \caption{Visualization of the retracking procedure and the alternative procedure proposed in Experiment 3. a) Tracking is performed ``forward'' until a termination criteria is hit. b) the streamline is flipped and ``retracked'' up until the seed point, then the tracking continues ``backward''. c) without the retracking procedure, the forward tracking is performed as usual. d) the ``backwards'' tracking starts at the seed point and tracks in the other direction, without retracking the first half of the streamline.}
                \label{fig:retracking}
            \end{figure}
            
            As described in \ref{sec:tracktolearn}, the WM tracking training procedure has agents retrack their own steps in hope of learning the \emph{backwards} tracking procedure more easily. When ``retracking'', actions generated by the agent are discarded until the retracking procedure is over, and then tracking proceeds as usual~\footnote{This is in spirit of teacher forcing when training recurrent neural networks~\cite{lamb2016professor}.}. However, we suspect this complexifies the learning procedure and could be a cause of instability. As such, in this experiment, we explore the impact of the retracking procedure on the training process and reconstructed tractograms.

            We train agents to perform tractography from WM seeding but without the retracking regime described in section \ref{sec:tracktolearn}. As such, when \emph{backwards} tracking begins, the initial state is simply computed from the input signal at the seed's position and using the first four directions of the reversed half-streamlines computed during the \emph{forward} phase. Figure~\ref{fig:retracking} offers a visual representation of the two tracking procedures.
            
            We track on the FiberCup, Flipped and ISMRM2015 datasets in this experiment. We repeat the hyperparameter search and training process presented in Experiment 1 for all algorithms.

        \subsubsection{Experiment 4: State}
            
            The input signal for the last experiments was the same as in \cite{theberge2021}: the fODF concatenated with the white matter mask and the last four tracking steps. This imposes a prior on the learning procedure and requires specific processing of the diffusion data. In this experiment, we explore the different components of the state to measure their impact on the training procedure and the reconstructed tractograms.
                
            \paragraph*{Raw diffusion signal}
                
                First, we explore the usage of the raw diffusion signal, resampled to 100 directions (akin to~\cite{neher_fiber_2017}) and min-max normalized, instead of the fODFs fed to the agents. The peaks used for the reward function still come from the fODFs used in Experiment 1. The rest of the input signal is left as previously defined. 
        
            \paragraph*{Previous directions}
            
                We then vary the number of previous directions given to the agents at each step, by giving 0 and 2 previous directions instead of 4. The rest of the input signal is again left as previously defined in \emph{Track-to-Learn}. This allows us to assess if fewer directions can still provide enough directional information to the agents, of if directional information is even needed at all.
                
            \paragraph{WM Mask}
                We finally train the SAC Auto agent using fODFs and the four previous directions, but {\em without} the WM mask as previously included. This allows us to determine whether agents can still stay in the WM even if they're not told exactly where it is and where it ends.
              
            For simplicity, we only train agents using the SAC Auto algorithm. We chose this algorithm for its robustness to different ranges of hyperparameters as well as its overall good performance. We repeat the hyperparameter search and training process presented in Experiment 1 for the SAC Auto agent.
              
            We repeat the hyperparameter search and training process for the SAC Auto agent described in Experiment 1 (final hyperparameter values are available in Appendix~\ref{app:exp4_hyperparameters}). 
            
        \subsubsection{Experiment 5: Reward function}

            As defined in eq.~\ref{eq:ttl_reward}, the environment computes the reward $r_t$ as the dot product between the tracking step $\vec{u}_{t}$ and the most aligned peak $\vec{v}(p_t)$ multiplied by the dot product between the tracking step $\vec{u}_{t}$ and the last streamline segment $\vec{u}_{t-1}$.
            
            We argue this reward function forces the learning agents to reproduce the behaviour of classical algorithms, with an added constraint on the streamlines' smoothness. In this experiment, we explore the addition of different components to the existing reward function in the hope of improving upon the results obtained previously.
        
            \paragraph{Length}
                
                We first explore the addition of a \emph{length bonus} component to the reward function so that it becomes:
                \begin{eqnarray}
                    r_t \, = \, \big(|\max_{\vec{v}(p_t)} \, \langle \vec{v}(p_t), \vec{u}_{t} \rangle| \cdot \langle \vec{u}_t, \vec{u}_{t-1}\rangle\big) + \alpha_{\text{length}} \big(l(\vec{u}) / l_\text{max}\big),
                \end{eqnarray}
                where $l(\vec{u})$ is the length of the streamline being tracked (because the step size is constant, the length in this case is simply the number of points in the streamline), $l_\text{max}$ is the maximum streamline length and $\alpha_{\text{length}}$ is a hyperparameter controlling the amplitude of the bonus. We hypothesize that adding a length component to the reward function will lead to longer streamlines and fewer ``broken'' streamlines (streamlines not terminating in GM). We explore $\alpha_{\text{length}}$ values of 0.01, 0.1, 1 and 5.
                
            \paragraph{Reaching GM}
                We then explore the addition of a \emph{GM-termination bonus} component to the reward function so that it becomes:
                \begin{eqnarray}
                    r_t \, = \, \big(|\max_{\vec{v}(p_t)} \, \langle \vec{v}(p_t), \vec{u}_{t} \rangle| \cdot \langle \vec{u}_t, \vec{u}_{t-1}\rangle\big) + \alpha_{\text{GM}} \mathds{1}_\text{GM}(\vec{u}),
                \end{eqnarray}
                where $\mathds{1}_\text{GM}(\vec{u}) = 1$ if the streamline has reached GM and $0$ otherwise and $\alpha_{\text{GM}}$ is a hyperparameter controlling the amplitude of the bonus. We hypothesize that this bonus component will encourage agents to reach GM and reconstruct fewer broken or looping streamlines. We explore $\alpha_{\text{GM}}$ values of 1, 10 and 100. 
                
                To minimize variability and allow for a fair comparison, we keep the same experiment procedure as in Experiment 1 for the SAC Auto agent.
        
\section{Results}
\label{sec:results}
    
    Here we present the results obtained for all experiments presented in section \ref{sec:benchmarks}.

    \subsection{Experiment 1 - Agents}\label{sec:exp1_results}

        \begin{figure}[!t]
            \centering
            \includegraphics[width=1.\textwidth]{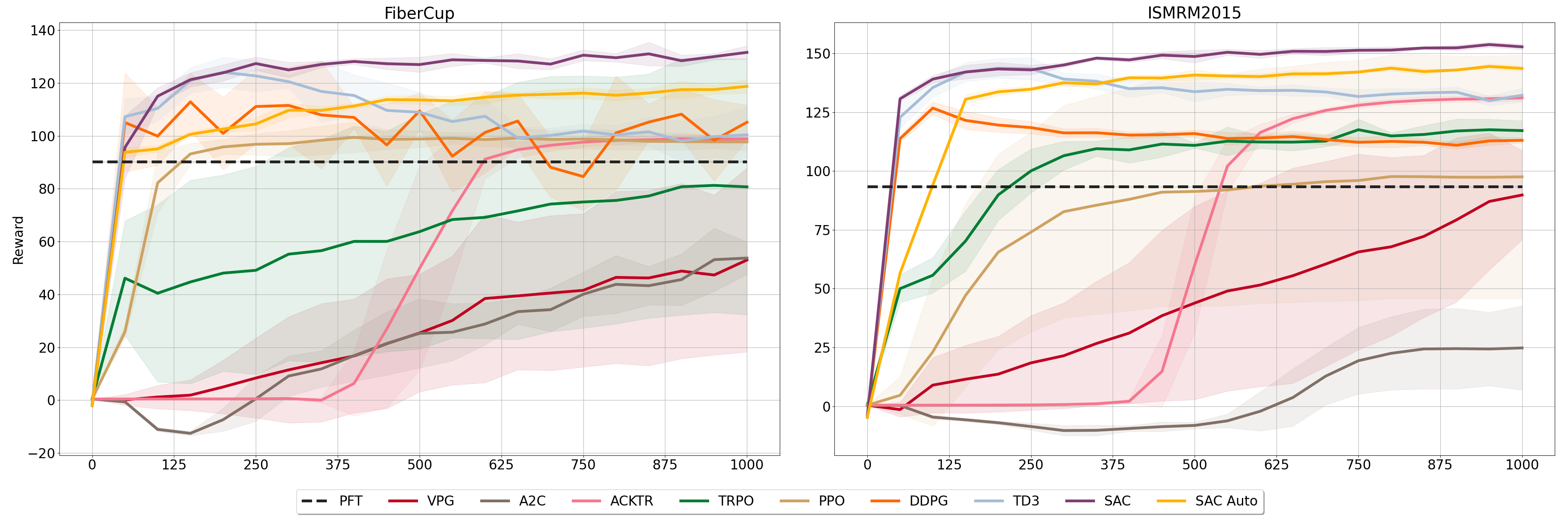}
            \caption{Progression of the average sum of reward per streamline obtained by agents during training using WM seeding on the FiberCup and ISMRM2015 datasets. Dashed line represents the reward ``obtained'' by PFT by rewarding its tractograms \emph{a posteriori} as reference. Middle line represents mean reward from 5 training runs with different random seeds. Shading represents $\pm$ standard deviation.}\label{fig:wm_seeding_reward}
            
            \includegraphics[width=1.\textwidth]{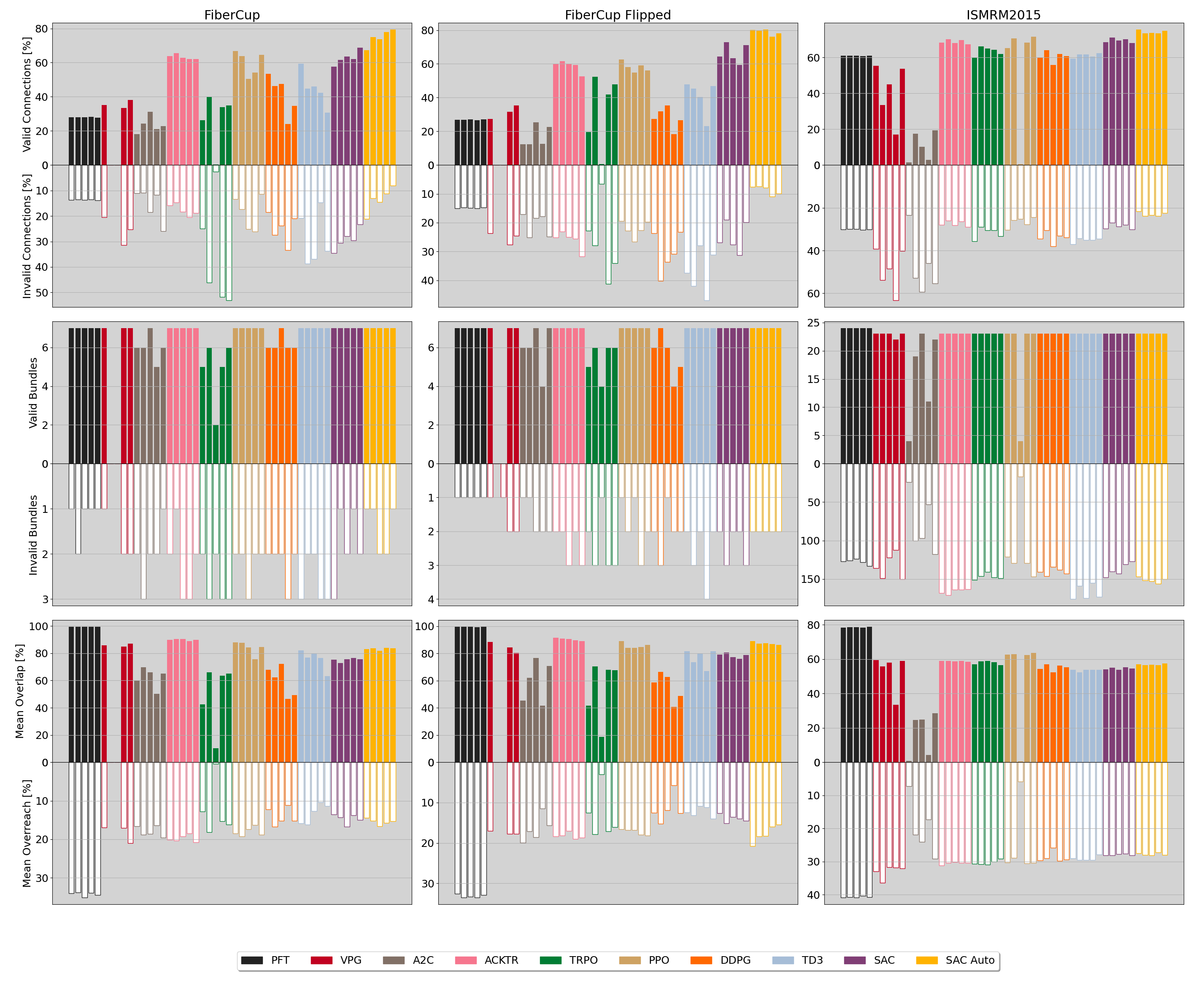}
            \caption{Valid connections (VC) vs invalid connections (IC), valid bundles (VB) vs. invalid bundles (IB), mean overlap (OL) vs mean overreach (OR) for tractograms generated on the FiberCup, Flipped and ISMRM2015 datasets for all agents during Experiment 1 with 5 training runs. }\label{fig:wm_seeding_bundles}
        \end{figure}
       
        \begin{sidewaysfigure}[!t]
            \centering
            \includegraphics[width=1.\textwidth]{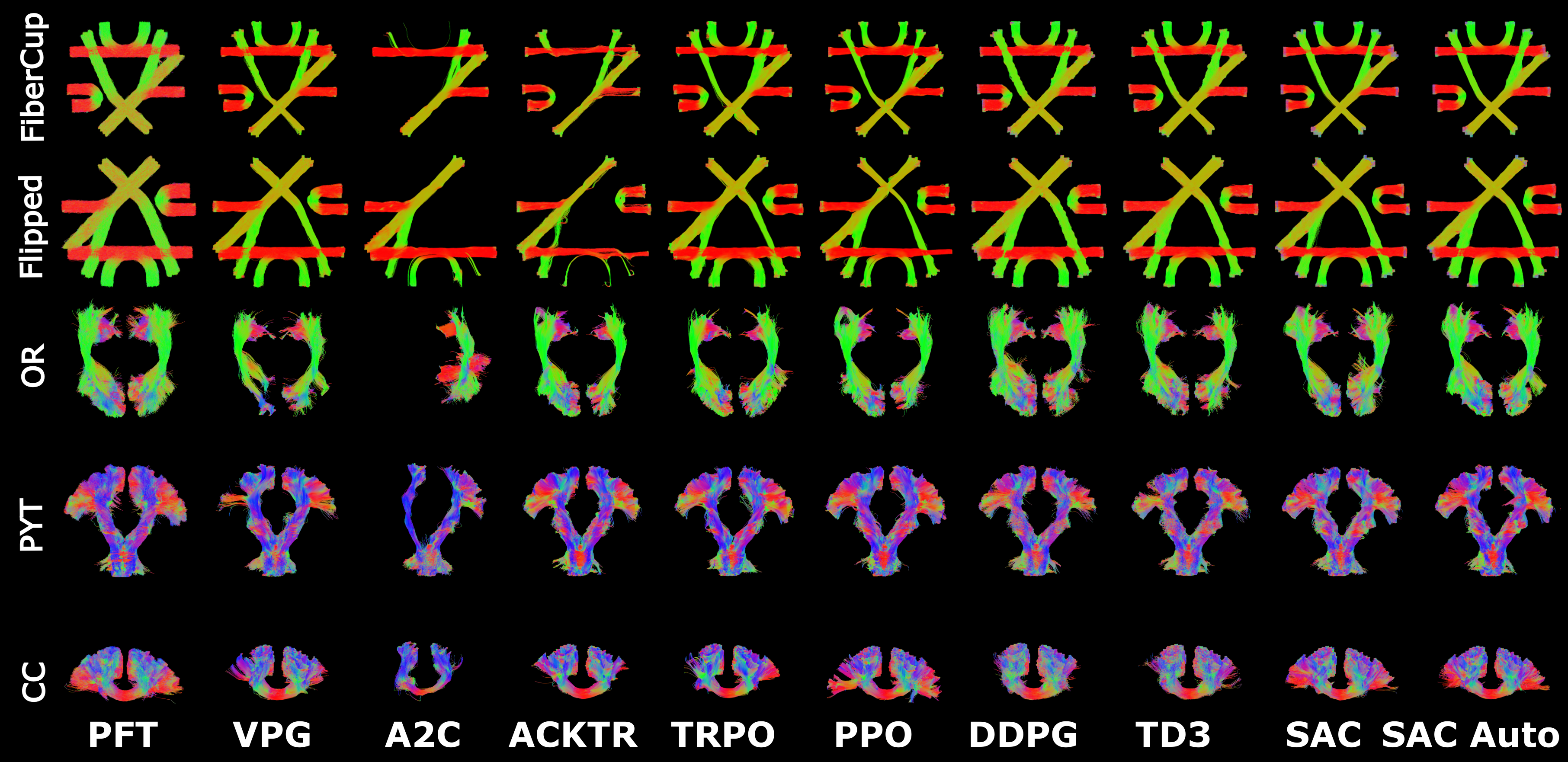}
            \caption{FiberCup, Flipped FiberCup, Optic Radiations (OR), Pyramidal Tracts (PYT) and Corpus Callosum (CC) bundles reconstructed by all algorithms using WM seeding.}
            \label{fig:wm_seeding_viz}
        \end{sidewaysfigure} 
 
        \begin{table}[!th] 
            \caption{Mean $\pm$ std. Dice scores for the OR, PYT and CC bundles on subject `sub-1006` of the TractoInferno dataset with tracts initialized from the WM. Supervised learning scores are reported from ~\cite{poulin2022tractoinferno}, which did not include std. \textbf{Bold} in the first section indicates the best performing supervised method as reported by~\cite{poulin2022tractoinferno}. \textbf{Bold} in the last section indicates which method matched or surpassed the best supervised method.}
            \label{tab:exp1_tractoinferno_results}
            \renewcommand{\arraystretch}{1.2}
            \begin{tabularx}{\columnwidth}{>{\hsize=.1\hsize}X|*3{>{\centering\arraybackslash}X}|}
            \cline{2-4}
            \multicolumn{1}{l}{}
            & \multicolumn{1}{|c }{OR}
            & \multicolumn{1}{c }{PYT}
            & \multicolumn{1}{c |}{CC} \\ \hline
            \multicolumn{ 1 }{|l|}{ Det-SE } & 0.569 & 0.665 & 0.658  \\
            \multicolumn{ 1 }{|l|}{ Det-Cosine } & 0.598 & 0.708 & 0.646  \\
            \multicolumn{ 1 }{|l|}{ Prob-Sphere } & \textbf{0.599} & 0.695 & 0.648  \\
            \multicolumn{ 1 }{|l|}{ Prob-Gaussian } & 0.542 & \textbf{0.723} & \textbf{0.668}  \\
            \multicolumn{ 1 }{|l|}{ Prob-Mixture } & 0.436 & 0.522 & 0.614  \\ \hhline{|=|===|}
            \multicolumn{ 1 }{|l|}{ PFT } & 0.644 $\pm$ 0.136 & 0.753 $\pm$ 0.010 & 0.827 $\pm$ 0.008  \\ \hline
            \multicolumn{ 1 }{|l|}{ VPG } & 0.369 $\pm$ 0.135 & 0.434 $\pm$ 0.128 & 0.428 $\pm$ 0.182  \\
            \multicolumn{ 1 }{|l|}{ A2C } & 0.225 $\pm$ 0.108 & 0.323 $\pm$ 0.082 & 0.222 $\pm$ 0.025  \\
            \multicolumn{ 1 }{|l|}{ ACKTR } & 0.397 $\pm$ 0.171 & 0.559 $\pm$ 0.028 & 0.584 $\pm$ 0.054  \\
            \multicolumn{ 1 }{|l|}{ TRPO } & 0.330 $\pm$ 0.154 & 0.498 $\pm$ 0.062 & 0.594 $\pm$ 0.048  \\
            \multicolumn{ 1 }{|l|}{ PPO } & 0.440 $\pm$ 0.187 & 0.619 $\pm$ 0.042 & 0.650 $\pm$ 0.028  \\
            \multicolumn{ 1 }{|l|}{ DDPG } & \textbf{0.612 $\pm$ 0.063} & 0.630 $\pm$ 0.045 & \textbf{0.731 $\pm$ 0.006}  \\
            \multicolumn{ 1 }{|l|}{ TD3 } & 0.555 $\pm$ 0.097 & 0.603 $\pm$ 0.045 & \textbf{0.688 $\pm$ 0.035}  \\
            \multicolumn{ 1 }{|l|}{ SAC } & 0.598 $\pm$ 0.098 & 0.658 $\pm$ 0.028 & \textbf{0.753 $\pm$ 0.010}  \\
            \multicolumn{ 1 }{|l|}{ SAC Auto } & \textbf{0.608 $\pm$ 0.088} & 0.655 $\pm$ 0.032 & \textbf{0.747 $\pm$ 0.019}  \\ \hline
            \end{tabularx}       
        \end{table}
        
        \begin{figure}[!t]
            \centering
            \includegraphics[width=1\textwidth]{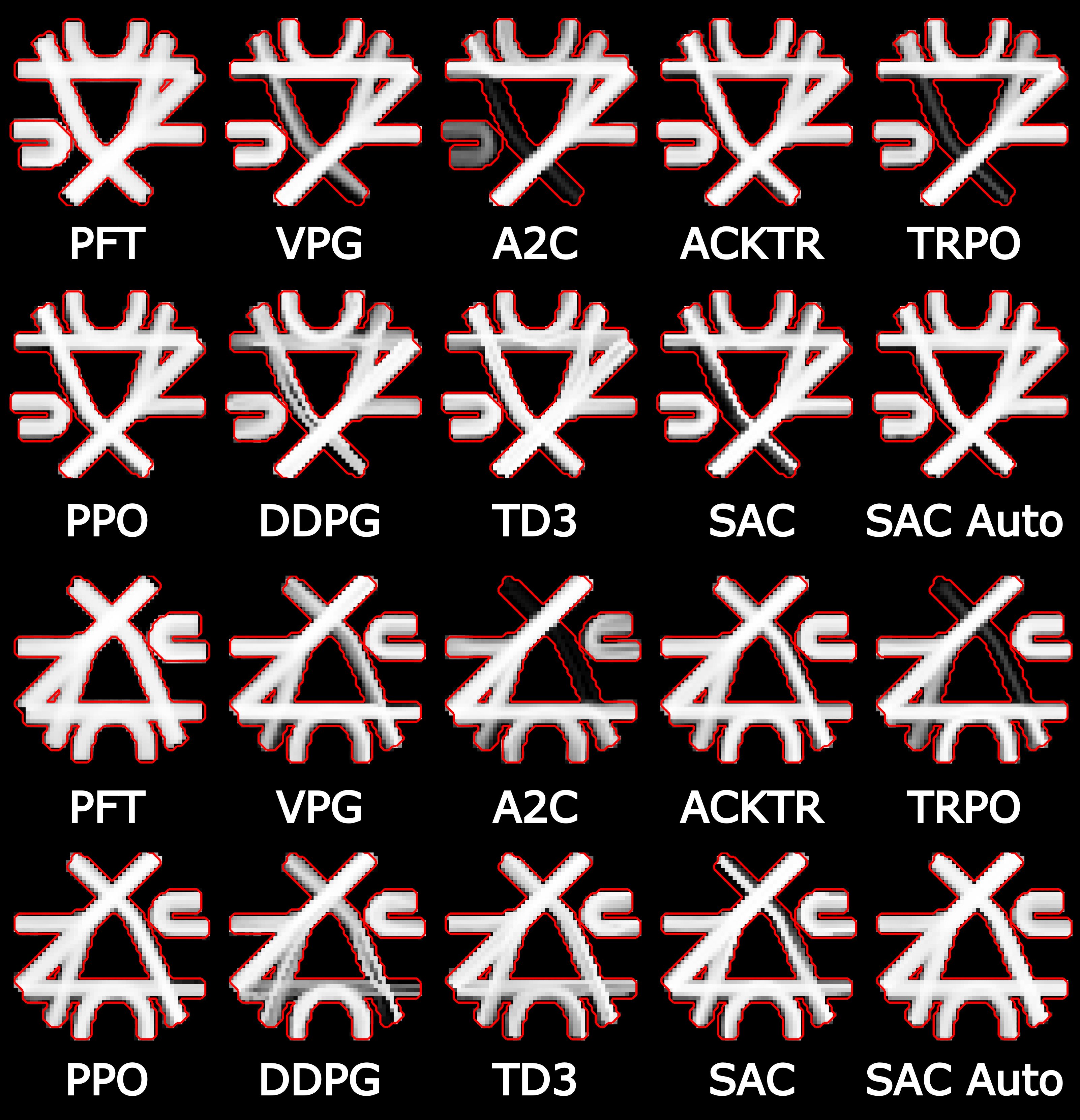}
            \caption{Streamline density maps of the valid connections reconstructed on the FiberCup (top two rows) and Flipped (bottom two rows) datasets during Experiment 1 for all tracking algorithms. The WM tracking mask boundary is colored red. Intensities are log-scaled independently for each map. }
            \label{fig:streamline_density}
        \end{figure}

        Figure \ref{fig:wm_seeding_reward} presents the progression of the average sum of reward (ASR) per streamline obtained during training on the FiberCup and ISMRM2015 datasets for all agents. The dotted line presents the ASR calculated on the tractograms from PFT \emph{a posteriori}. As can be seen, SAC and SAC Auto achieve the best ASR on both dataset, while VPG and A2C achieve the worst. DDPG, TD3, SAC and SAC Auto quickly reach a high ASR, but the performances of DDPG and TD3 decrease throughout the epochs, whereas ACKTR is slower to learn a good policy.
        
        Figure \ref{fig:wm_seeding_bundles} shows some Tractometer metrics for all agents after tracking on the FiberCup, Flipped, and ISMRM2015 datasets (see Appendix \ref{app:exp1_results} for full scores). While PPO generally displays good performance, it fails to learn for one training run on the ISMRM2015 dataset. Moreover, ACKTR, PPO (except the one agent), SAC and SAC Auto obtain the best VC rates on all datasets. Although PFT obtains better OL than RL algorithms, it also produces a much higher OR. Furthermore, ACKTR, PPO (except for the one agent), SAC and SAC Auto reconstructed 7/7 bundles on the FiberCup and its flipped version, and 23 out of 25 bundles on ISMRM2015. Only PFT managed to reconstruct a 24th bundle on the ISMRM2015 dataset.
        
        Figure~\ref{fig:wm_seeding_viz} presents tractograms of valid connections extracted by the Tractometer on the FiberCups and three sets of bundles on the TractoInferno dataset: Optic Radiations (OR), Pyramidal Tracts (PYT) and part of the Corpus Callosum (CC), left and right instances when applicable, segmented from whole-brain tractography on the TractoInferno dataset by all agents and the PFT algorithms.

        From this figure, we see that the best performing agents on the FiberCups (ACKTR, PPO, SAC and SAC Auto) according to the Tractometer also reconstruct visually appealing tractograms, whereas poorer performing methods such as A2C, TRPO and DDPG tend to miss parts of some bundles. We also observe that, per algorithm, the reconstructions of the flipped version of the FiberCup tend to be similar and even exhibit the same ``quirks'' (i.e., skinnier endpoints of parts of bundles) to their nonflipped counterparts. This could mean that the algorithms do not ``overfit'' to the FiberCup's bundle configuration during training. 
        

        We also observe in figure~\ref{fig:wm_seeding_viz} that,  similarly to the FiberCup and Flipped case, agents that perform well \emph{ in silico} also reconstruct visually appealing bundles \emph{ in vivo}. Although the fanning at the endpoints of the bundles reconstructed by RL agents is generally less voluminous than in reconstructions by PFT, the ``body'' of the bundles is well reconstructed and streamlines are generally ``smoother''.

        Table~\ref{tab:exp1_tractoinferno_results} presents the Dice scores computed on the TractoInferno datasets for the same three bundles. DDPG and SAC Auto surpass the Dice score of the best performing supervised agents for the OR and DDPG, TD3, SAC, and SAC Auto surpass the best performing supervised agent on the CC, while not even being trained in-vivo. No agent, RL or supervised, matches the Dice scores of PFT on all three bundles.
        
        Finally, Figure~\ref{fig:streamline_density} displays streamline density maps for tractograms reconstructed during Experiment 1 on the FiberCup and Flipped dataset by the PFT algorithm and the RL agents. From it we can see that the streamline density of the tractograms generated by RL agents is less uniform and lower (often zero) near the WM boundary. This behavior may lead to a lower OL with the reference bundles (cf.~\ref{sub:overlap}).

        All selected hyperparameters for this experiment are available in Appendix~\ref{app:exp1_hyperparameters}.
        
    \subsection{Experiment 2 - Interface seeding}\label{sec:exp2_results}

        \begin{figure}[!t]
            \centering
            \includegraphics[width=1\textwidth]{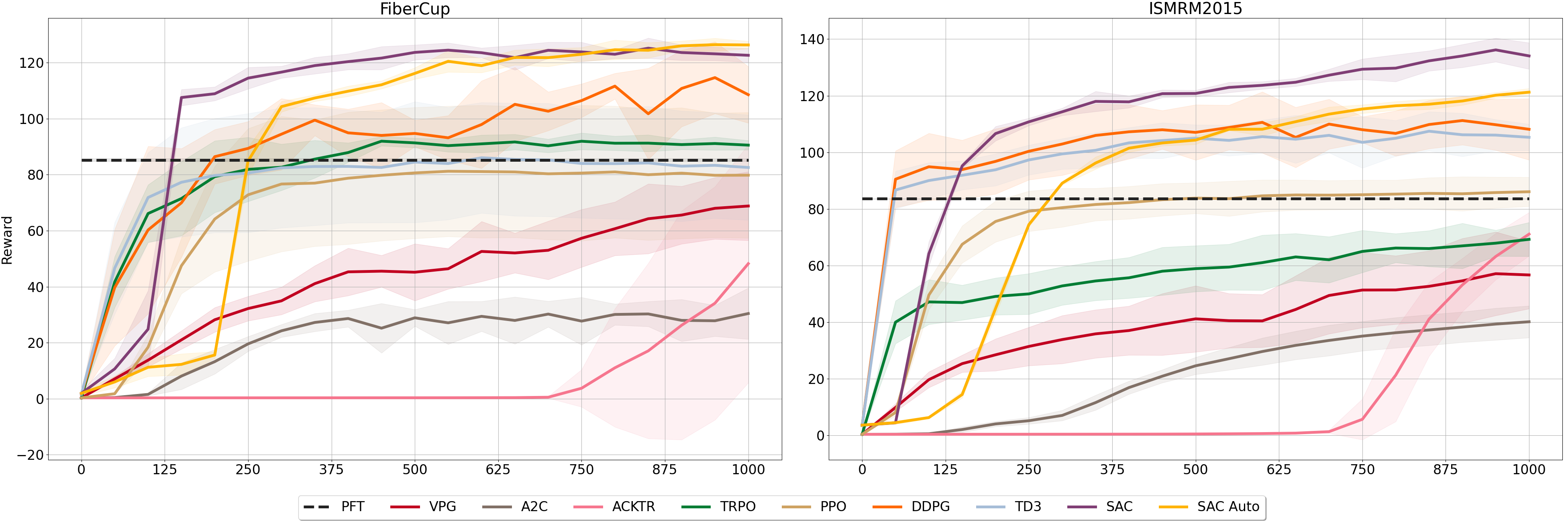}
            \caption{Progression of the average sum of reward per streamline obtained by agents during training using interface seeding. Dashed line represents the reward ``obtained'' by PFT by rewarding its tractograms \emph{a posteriori}. Middle line represents mean reward from 5 training runs with different random seeds. Shading represents standard deviation.}
            \label{fig:interface_seeding_reward}

            \includegraphics[width=1.0\textwidth]{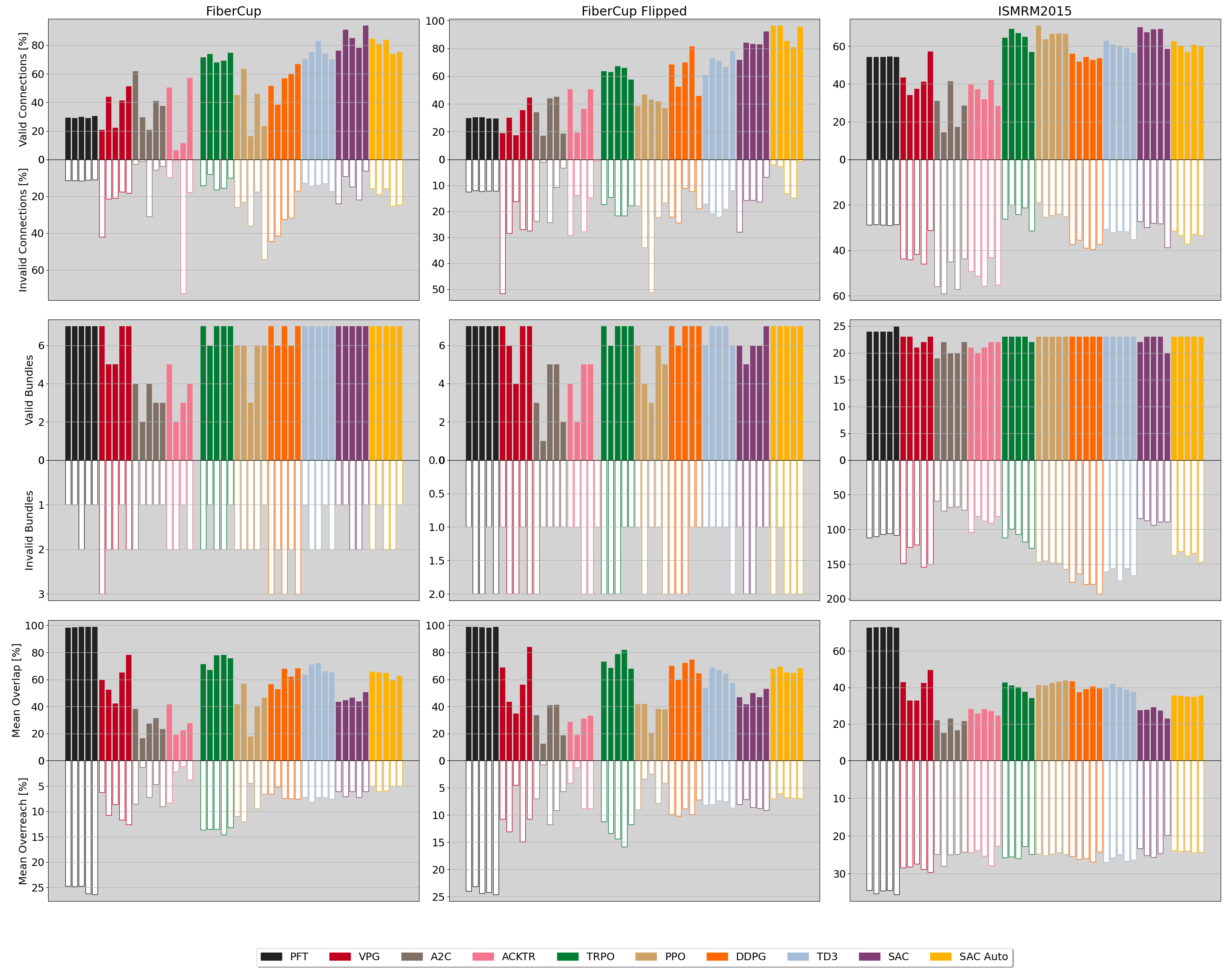}
            \caption{Valid connections (VC) vs Invalid connections (IC), Valid bundles (VB) vs. Invalid bundles (IB), mean overlap (OL) vs mean overreach (OR) for tractograms generated on the FiberCup, Flipped and ISMRM2015 datasets for all agents during Experiment 2.}
            \label{fig:interface_seeding_bundles}
        \end{figure} 
        
        \begin{sidewaysfigure}[!t]
            \centering
           
            \includegraphics[width=1.\textwidth]{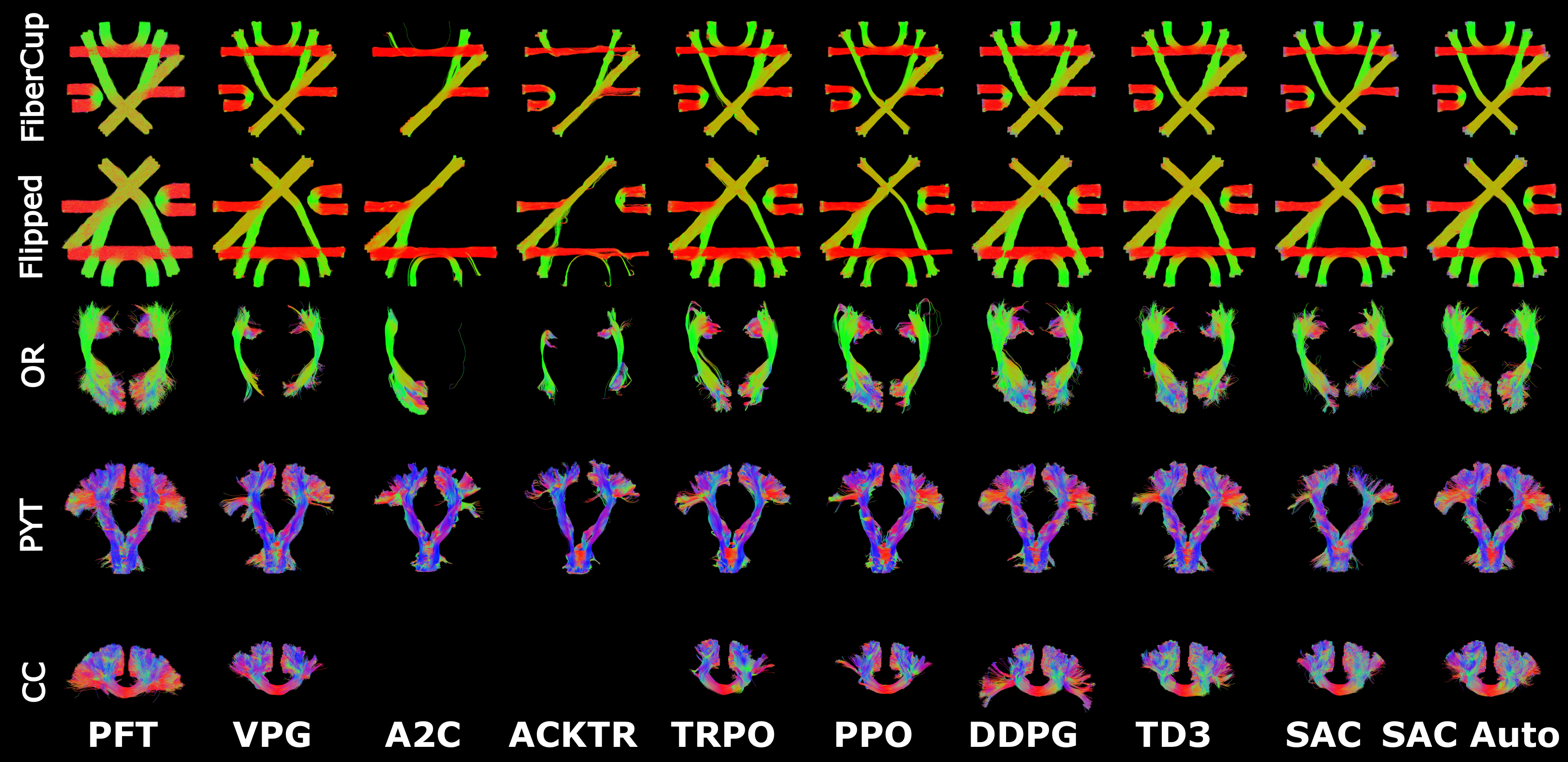}
            \caption{FiberCup, Flipped FiberCup, Optic Radiations (OR), Pyramidal Tracts (PYT) and Corpus Callosum (CC) bundles reconstructed by all algorithms using interface seeding.}
            \label{fig:interface_seeding_viz}
        \end{sidewaysfigure} 
  
        \begin{table}[!th] 
            \caption{Mean $\pm$ std. Dice scores for the OR, PYT and CC bundles on subject `sub-1006` of the TractoInferno dataset from agents tracking from the WM/GM interface. Supervised learning scores are reported from ~\cite{poulin2022tractoinferno}, which did not include std. \textbf{Bold} in the first section indicates the supervised method that performs best, as reported by~\cite{poulin2022tractoinferno}. \textbf{Bold} in the last section indicates which method matched or surpassed the best supervised method.}
            \label{tab:exp2_tractoinferno_results}
            \renewcommand{\arraystretch}{1.2}
            \begin{tabularx}{\columnwidth}{>{\hsize=.1\hsize}X|*3{>{\centering\arraybackslash}X}|}
            \cline{2-4}
            \multicolumn{1}{l}{}
            & \multicolumn{1}{|c }{OR}
            & \multicolumn{1}{c }{PYT}
            & \multicolumn{1}{c |}{CC} \\ \hline
            \multicolumn{ 1 }{|l|}{ Det-SE } & 0.569 & 0.665 & 0.658  \\
            \multicolumn{ 1 }{|l|}{ Det-Cosine } & 0.598 & 0.708 & 0.646  \\
            \multicolumn{ 1 }{|l|}{ Prob-Sphere } & \textbf{0.599} & 0.695 & 0.648  \\
            \multicolumn{ 1 }{|l|}{ Prob-Gaussian } & 0.542 & \textbf{0.723} & \textbf{0.668}  \\
            \multicolumn{ 1 }{|l|}{ Prob-Mixture } & 0.436 & 0.522 & 0.614  \\ \hhline{|=|===|}
            \multicolumn{ 1 }{|l|}{ PFT } & 0.586 $\pm$ 0.060 & 0.671 $\pm$ 0.021 & 0.768 $\pm$ 0.030  \\ \hline
            \multicolumn{ 1 }{|l|}{ VPG } & 0.139 $\pm$ 0.096 & 0.296 $\pm$ 0.138 & 0.374 $\pm$ 0.186  \\
            \multicolumn{ 1 }{|l|}{ A2C } & 0.228 $\pm$ 0.222 & 0.264 $\pm$ 0.101 & 0.000 $\pm$ 0.000  \\
            \multicolumn{ 1 }{|l|}{ ACKTR } & 0.099 $\pm$ 0.074 & 0.188 $\pm$ 0.132 & 0.000 $\pm$ 0.000  \\
            \multicolumn{ 1 }{|l|}{ TRPO } & 0.216 $\pm$ 0.137 & 0.262 $\pm$ 0.096 & 0.251 $\pm$ 0.144  \\
            \multicolumn{ 1 }{|l|}{ PPO } & 0.296 $\pm$ 0.162 & 0.284 $\pm$ 0.097 & 0.258 $\pm$ 0.092  \\
            \multicolumn{ 1 }{|l|}{ DDPG } & 0.428 $\pm$ 0.171 & 0.446 $\pm$ 0.073 & 0.576 $\pm$ 0.034  \\
            \multicolumn{ 1 }{|l|}{ TD3 } & 0.433 $\pm$ 0.144 & 0.458 $\pm$ 0.054 & 0.558 $\pm$ 0.014  \\
            \multicolumn{ 1 }{|l|}{ SAC } & 0.257 $\pm$ 0.122 & 0.267 $\pm$ 0.122 & 0.408 $\pm$ 0.140  \\
            \multicolumn{ 1 }{|l|}{ SAC Auto } & 0.505 $\pm$ 0.139 & 0.520 $\pm$ 0.077 & \textbf{0.668 $\pm$ 0.026}  \\ \hline
            \end{tabularx}       
        \end{table}      
              
        \begin{figure}[!t]
            \centering
            \includegraphics[width=1\textwidth]{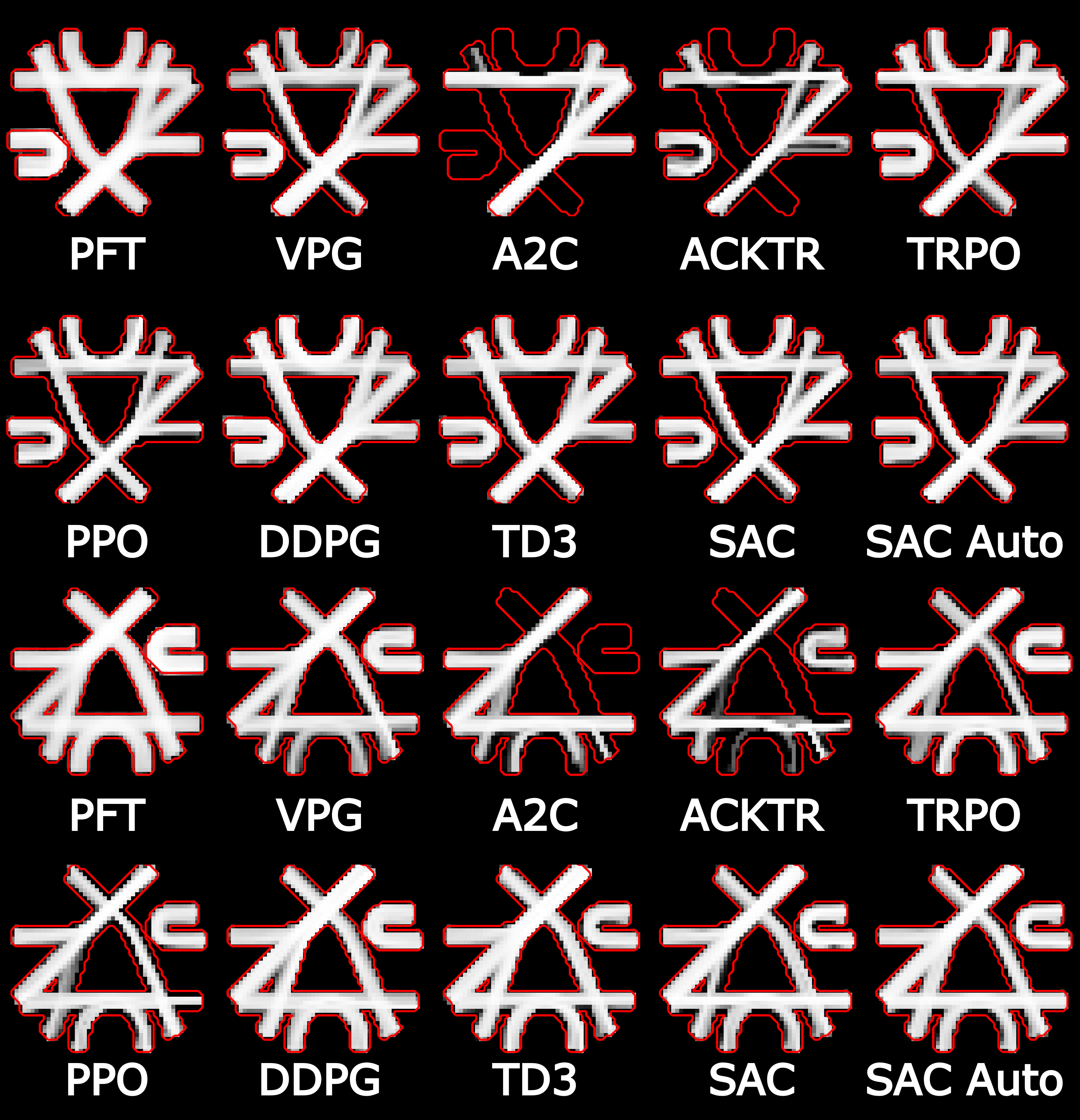}
            \caption{Streamline density maps of the valid connections reconstructed on the FiberCup (top two rows) and Flipped (bottom two rows) datasets during Experiment 2 for all tracking algorithms. The WM tracking mask boundary is shown in red. Intensities are log-scaled independently for each map. }
            \label{fig:interface_streamline_density}
        \end{figure} 
        
        Figure \ref{fig:interface_seeding_reward} presents the progression of ASR obtained during training using interface seeding on the FiberCup and ISMRM2015 datasets for all agents. The SAC and SAC Auto agents are again able to obtain the best ASR, while VPG and A2C again fail to learn how to produce actions that are well rewarded. Contrary to the previous experiment, ACKTR also cannot learn a decent policy during the training regime. Although it would seem that TRPO also fails to obtain good performance, figure~\ref{fig:interface_seeding_bundles} paints a different picture. 

        Figure \ref{fig:interface_seeding_bundles} reports some Tractometer metrics for all agents after tracking on the FiberCup, Flipped, and ISMRM2015 datasets using interface seeding (see Appendix \ref{app:exp2_results} for full scores). Again, the SAC and SAC Auto agents outperform PFT in terms of VC on all datasets, with TRPO and PPO also obtaining better VC. Moreover, PFT is able to recover 25 bundles in one instance and 24 in others on the ISMRM2015 dataset, whereas only one TRPO agent is able to recover a 24th bundle, most recovering only 23. We can observe consistency between the performances of the agents on the FiberCup and Flipped datasets. PFT again obtains better OL on all datasets, at the cost of much worse OR. 

        Figure \ref{fig:interface_seeding_viz} shows the valid connections extracted by the Tractometer on the FiberCup and Flipped datasets as well as OR, PYT and CC bundles from the TractoInferno dataset as reconstructed by all agents. The poor overlap and bundle detection of A2C and ACKTR shown in figure~\ref{fig:interface_seeding_bundles} is reflected here, as several bundles are either missing or skinny. Other RL agents reconstruct all bundles on all datasets, but produce noticeably skinnier bundles than PFT. Fanning at the endpoints of bundles is still lacking in some algorithms, but is mostly present. 

        Table~\ref{tab:exp2_tractoinferno_results} presents the Dice scores of the same three bundles for agents trained with interface seeding. Although the number of seeds per voxel is adapted to obtain roughly the same total number of seeds, the resulting Dice scores decrease such that RL agents no longer match the scores of supervised learning algorithms, which use WM seeding and are trained on the same dataset. Only the SAC Auto agent is able to match the Dice score of the best performing supervised agent on the CC. However, we also notice that the scores of PFT initialized via interface seeding also decrease to a point where it is surpassed even by supervised learning algorithms on the OR and PYT bundles.

        Figure~\ref{fig:interface_streamline_density} again presents the streamline density of the valid-connection tractograms reconstructed by all agents during Experiment 2. We can see that while PFT fills most of the WM volume, RL agents instead tend to mainly fill the ``middle'' of the bundles and avoid the edges. 

        All selected hyperparameters for this experiment are available in Appendix~\ref{app:exp2_hyperparameters}.
        
    \subsection{Experiment 3 - No retracking}\label{sec:exp3_results}
    
            \begin{table}[!thbp] 
            \caption{Difference between Tractometer scores for experiments 3 and 1 for the SAC Auto agent for training $\rightarrow$ testing datasets. \textbf{Bold} indicates a statistically different mean using a Welch t-test with p < 0.005.}
            \label{tab:exp3_sac_auto_results}
            \renewcommand{\arraystretch}{1.2}
            \begin{tabularx}{\columnwidth}{>{\hsize=.1\hsize}X|*5{>{\centering\arraybackslash}X}|}
            \cline{2-6}
            \multicolumn{2}{|l|}{Training $\rightarrow$ Testing}
            & \multicolumn{1}{c }{VC $\% \uparrow$}
            & \multicolumn{1}{c }{IC $\% \downarrow$}
            & \multicolumn{1}{c }{NC $\downarrow$}
            & \multicolumn{1}{c |}{OL $\uparrow$} \\ \hline

            \multicolumn{ 2 }{|l|}{ FiberCup $\rightarrow$ FiberCup } & \textbf{-60.90} $\pm$ 2.36& \textbf{16.24} $\pm$ 4.62& \textbf{44.66} $\pm$ 2.65& \textbf{-27.97} $\pm$ 2.59 \\

            \multicolumn{ 2 }{|l|}{ FiberCup $\rightarrow$ Flipped } & \textbf{-69.44} $\pm$ 3.85& 2.09 $\pm$ 3.67& \textbf{67.35} $\pm$ 7.22& \textbf{-28.86} $\pm$ 3.88 \\ 

            \multicolumn{ 2 }{|l|}{ ISMRM2015 $\rightarrow$ ISMRM2015 } & \textbf{-31.23} $\pm$ 4.12& \textbf{18.15} $\pm$ 1.45& \textbf{13.07} $\pm$ 3.01& \textbf{-14.84} $\pm$ 1.10 \\ \hline

            \end{tabularx}       
            \end{table}      

            Table~\ref{tab:exp3_sac_auto_results} displays the difference between some Tractometer scores of Experiment 3 and Experiment 1 for the SAC Auto agent. For all the metrics and algorithms considered, the exclusion of the retracking procedure has catastrophic results on the reconstructed tractograms. The VC rate for all agents on all datasets goes down. Even if the IC rate goes down for some agents, the NC rate goes up, indicating that more broken streamlines were reconstructed overall. Although the OR generally decreases, the OL decreases more significantly. 
            
            Full results for all algorithms are available in Appendix~\ref{app:exp3_full_results}). All selected hyperparameters for this experiment are available in Appendix~\ref{app:exp3_hyperparameters}.

    \subsection{Experiment 4 - State}\label{sec:exp4_results}

        \begin{figure}[!t]
            \centering
            \includegraphics[width=1\textwidth]{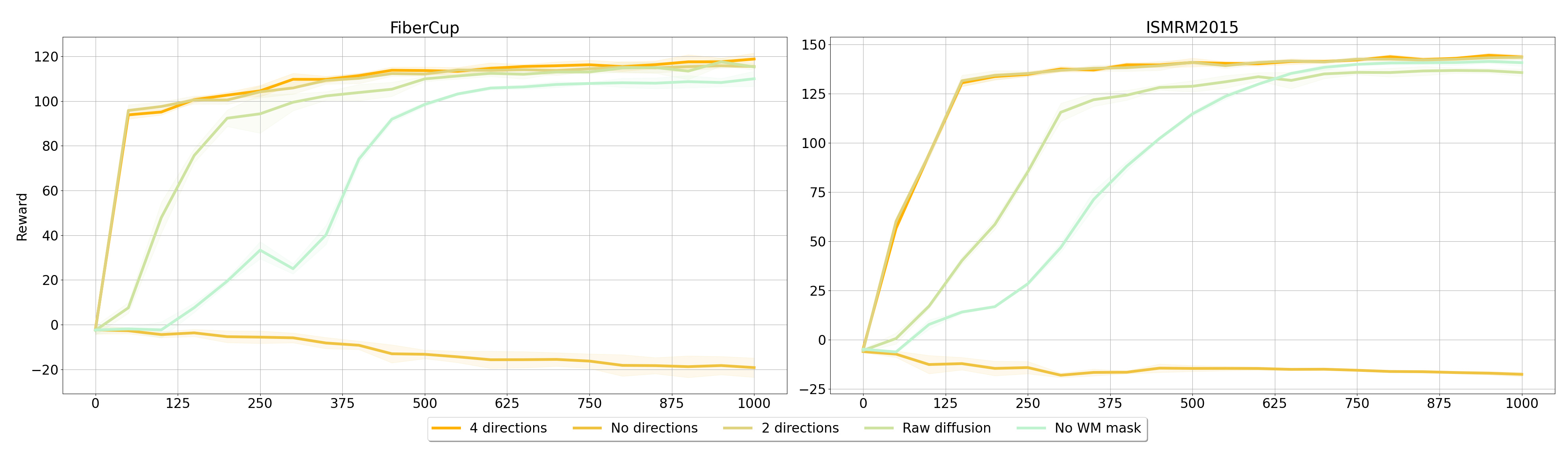}
            \caption{Progression of the sum of reward obtained by agents during training for Experiment 4. ``Base'' represents the SAC Auto agents trained for Experiment 1. Middle line represents mean reward from 5 training runs with different random seeds. Shading represents standard deviation.}
            \label{fig:exp4_reward}

            \includegraphics[width=1.\textwidth]{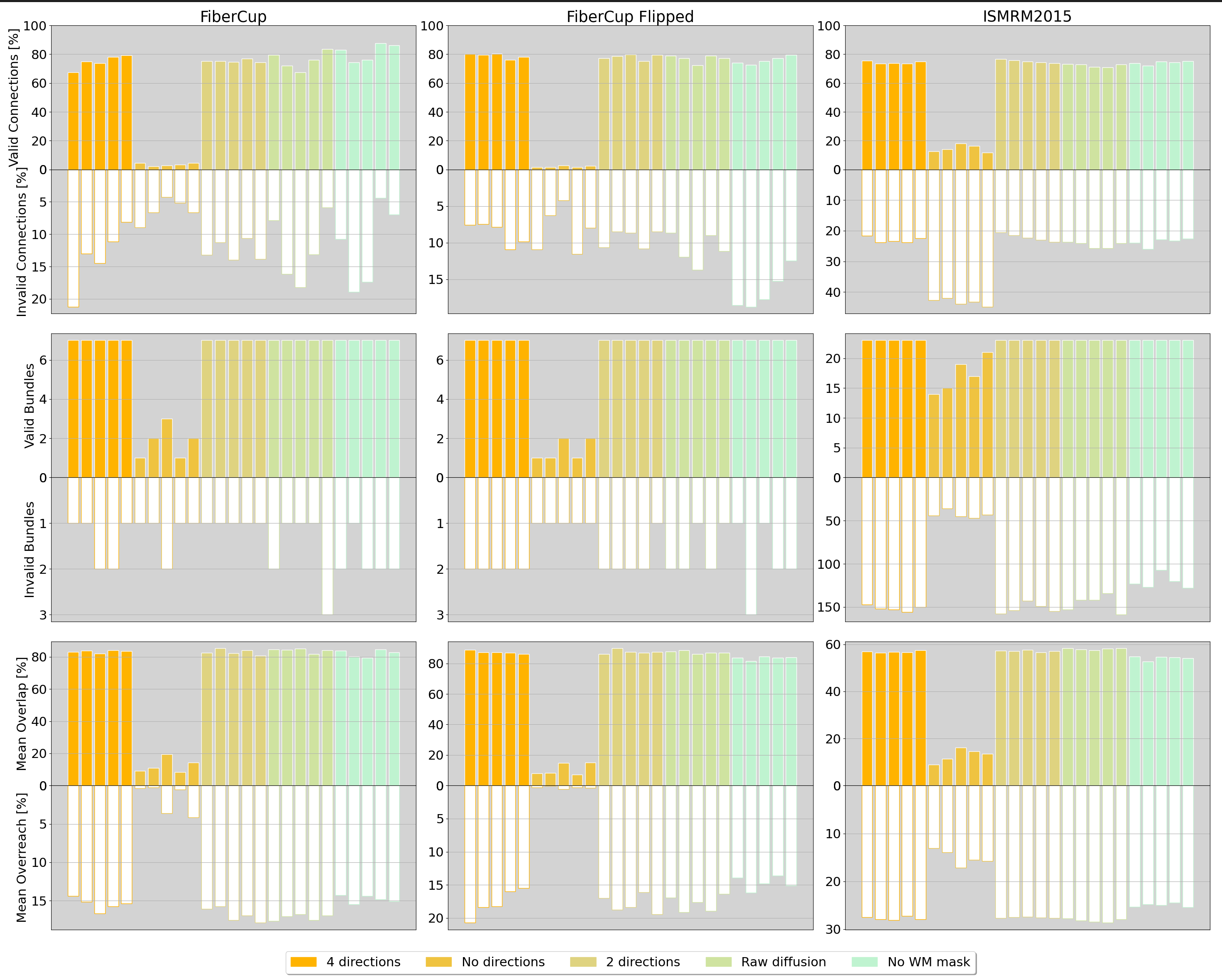}
            \caption{Valid connections (VC) vs Invalid connections (IC), Valid bundles (VB) vs. Invalid bundles (IB), mean overlap (OL) vs mean overreach (OR) for tractograms generated on the FiberCup, Flipped and ISMRM2015 datasets for all agents during Experiment 4.}
            \label{fig:exp4_bundles}
        \end{figure}

        Figure~\ref{fig:exp4_reward} plots the average reward per streamline obtained during training for all agents trained for Experiment 4: state variations. As can be seen, the agents trained with two directions closely follow the progression of reward accumulation of the reference agents. The ``raw'' and ``no WM'' agents are slower to learn a policy that accumulates the same amount of reward as the reference, but arrive at a similar return by the end of training. Agents trained with no directions as part of the state fail to learn a meaningful policy, even ending with a negative average reward per streamline.

        Figure~\ref{fig:exp4_bundles} presents some Tractometer scores for agents with the state variations considered in Experiment 4 on the FiberCup, Flipped and ISMRM2015 datasets. We observe that agents trained without any previous directions as part of the state fail to learn the tractography procedure completely. Moreover, agents instead trained with two directions do not exhibit different scores than the reference agents from Experiment 1. The same applies to agents trained on the raw diffusion signal. However, agents trained in fODFS but without a WM mask display a slightly lower OL and also lower OR across datasets. 
        
         All selected hyperparameters are available in Appendix~\ref{app:exp4_hyperparameters}. Full Tractometer scores are available in Appendix~\ref{app:exp4_results}.           
    \subsection{Experiment 5 - Reward function}\label{sec:exp5_results}

        \begin{figure}[!thbp]
            \centering
            \includegraphics[width=1\textwidth]{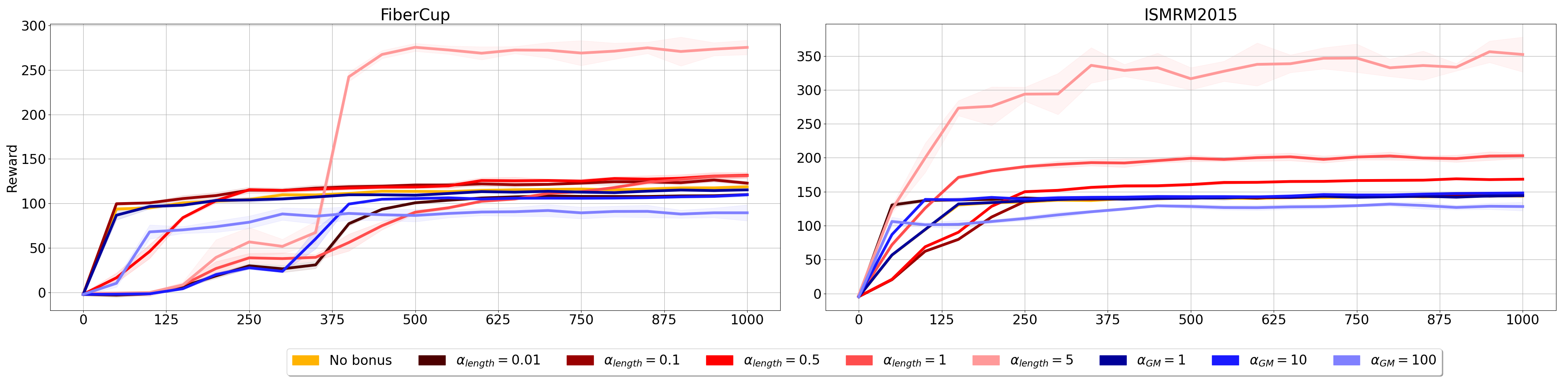}
            \caption{Progression of the sum of reward obtained by agents during training for Experiment 5. ``Base'' represents the SAC Auto agents trained for Experiment 1. Middle line represents mean reward from 5 training runs with different random seeds. Shading represents standard deviation.}
            \label{fig:exp5_reward}

            \includegraphics[width=1.\textwidth]{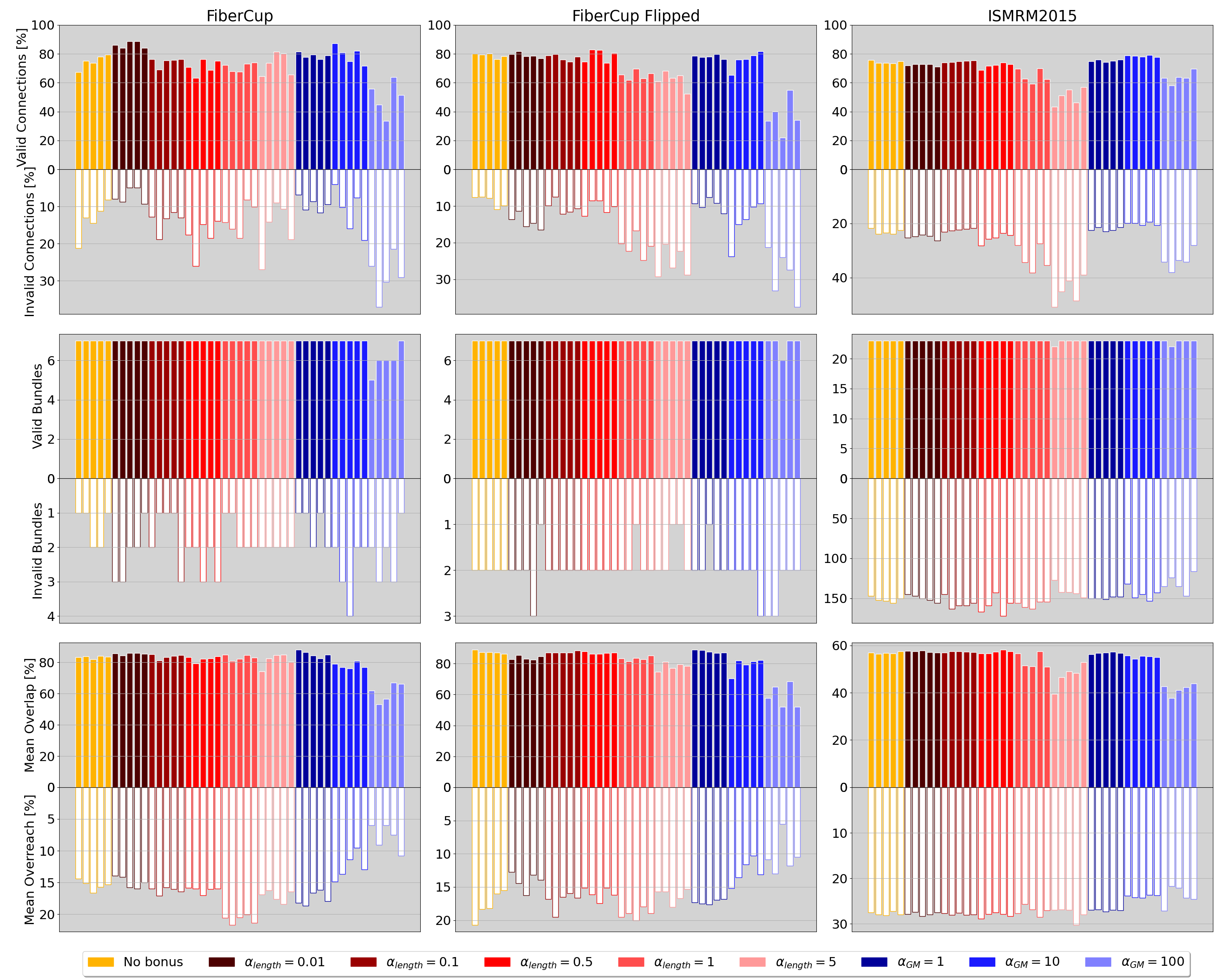}
            \caption{Valid connections (VC) vs Invalid connections (IC), Valid bundles (VB) vs. Invalid bundles (IB), mean overlap (OL) vs mean overreach (OR) for tractograms generated on the FiberCup, Flipped and ISMRM2015 datasets for all agents during Experiment 5.}
            \label{fig:exp5_bundles}
        \end{figure} 
            
            Figure~\ref{fig:exp5_reward} traces the progression of ASR for agents trained during Experiment 5. Noticeably, the agent trained with a length bonus of 5 quickly obtains a large amount of reward per episode, which is not surprising considering the bonus. However, more surprisingly, the agent trained with a bonus of 100 when reaching GM not only does not reach a high ASR during training but even obtains the worse ASR of all agents for this experiment. Most of the other agents obtain a similar ASR to the reference agent from Experiment 1.

            Figure~\ref{fig:exp5_bundles} displays some Tractometer metrics for agents trained with reward bonuses. Having a length bonus of 5 or a GM bonus of 100 seems to hamper the performances of the trained agents in terms of valid connections and overlap. On the contrary, the use of a GM bonus of 10 seems to increase the VC rate of agents on the FiberCup and ISMRM2015 datasets, but at the cost of reduced OL. A slight length bonus of 0.01 seems to provide mixed results, increasing VC and OL rates on the FiberCup, but not on the other datasets. Other bonuses either do not increase the performances of agents significantly or decrease them.
            
            All selected hyperparameters are available in Appendix~\ref{app:exp5_hyperparameters}. Full Tractometer scores are available in Appendix~\ref{app:exp5_results}.
            
\section{Discussion}
\label{sec:discussion}
    \subsection{Variations on the components of \emph{Track-to-Learn}}\label{sec:components}
    
        From the results presented in sections~\ref{sec:exp1_results}, \ref{sec:exp2_results}, \ref{sec:exp3_results}, \ref{sec:exp4_results} and \ref{sec:exp5_results}, we can infer a few key points about the~\emph{Track-to-Learn} framework.

        \paragraph{Off-policy algorithms are best-suited for tractography:} We underline in Sections~\ref{sec:exp1_results} and ~\ref{sec:exp2_results} that VPG and A2C consistently fail to obtain decent tracking performance, and ACKTR, TRPO, and PPO fail on some datasets and experiments. On the contrary, TD3, SAC and SAC Auto consistently achieve good or best results in all relevant experiments. Moreover, from figures~\ref{fig:wm_seeding_reward} and~\ref{fig:interface_seeding_reward}, we see that DDPG, TD3, SAC and SAC Auto are much more stable during training, whereas the other algorithms have more variance in their obtained reward throughout the training process. Overall, off-policy algorithms tend to perform better than on-policy algorithms, with SAC Auto generally performing best.
        
        \paragraph{Q-functions vs. Value functions:} While the use of a replay buffer is typically what separates on-policy and off-policy algorithms, the latter algorithms have another distinctive feature. The critics of the off-policy methods make use of a $Q$-function, whereas the critics of the on-policy methods use a value function (cf. Section ~\ref{sec:rl}) to predict the ``desirability'' of states. We hypothesize that the $Q$-function might provide an advantage to learning agents in the context of tractography, where choosing the next tracking step is a risky process. If the next tracking step is ill-chosen, the angle between the new streamline segment and the last might cause the tracking to end prematurely. Even if the angle is within acceptable bounds, the step might lead the streamline outside of the WM volume, terminating the tractography process. As such, the range of desirable actions is very limited in each state, causing all states to have a low \emph{expected} value. An algorithm only ranking the desirability of a particular state without considering the possible actions (which are mostly undesirable with only a select few allowing the tracking process to continue) might not be able to differentiate good states from bad ones. 
        
        \paragraph{\emph{Directionality} is crucial for the agents:} The retracking procedure, if seeding from the WM, is necessary and so are previous directions in the state. We hypothesize that this is because the input signal (i.e. the diffusion signal and the WM mask) by itself has no information about the directionality of steps taken. This is problematic because tractography is a sequential and oriented process: the previous step taken will influence the next one by, among other things, restricting the cone of possible steps. As such, without the retracking procedure, the agent has to learn to go ``in the other direction'' directly at the seeding point without having previously encountered what ``the other direction'' is. Worse, without any previous directions as part of the state, the agent cannot know where it came from, and therefore where it should go.
        
        \paragraph{The seeding strategy is left to the user's preference:} Comparing the results of Experiments 1 and 2, we see that interface seeding is not necessary, but may provide advantages to the user. For example, we can notice an increase in VC rates and a decrease in NC rates for most algorithms on the FiberCup datasets but also a decrease in OL. This is further validated by comparing the visualized bundles in Experiments 1 and 2, where the bundles in the latter experiment tend to be thinner. As such, the choice of seeding strategy can be left to the user, for example, depending on whether bundle coverage or connection accuracy is preferred. However, empirically, we did find that the retracking procedure takes time and that seeding from the WM/GM interface removes the need for a second environment, which speeds up training and tracking significantly.
       
        \paragraph{The state formulation can be simplified:} Replacing the fODF as part of the signal with the raw diffusion signal, reducing the number of previous directions to two, or removing the WM mask do not have a significant impact on the reconstructed tractograms and the computed Tractometer scores. This means that the state formulation can be reduced by omitting the WM mask and two directions, cutting down the number of parameters in the agents' networks. Removing the need for the WM mask might also reduce the complexity pipelines upstream: if the seeding and tracking masks are based on other criteria than tissue segmentation, much simpler pipelines could be used to feed data to Track-to-Learn.
        
        \paragraph{A bonus for reaching GM may be beneficial:} However, bonuses to the reward function should be added with care. Considered bonuses to the reward function often do not provide significant improvements to the agents' performance, even reducing it in many cases (cf. Section~\ref{sub:reward_analysis}).
    
    \subsection{Reward vs. performance}
    \label{sub:reward_analysis}

        Looking at figures~\ref{fig:wm_seeding_reward} and ~\ref{fig:interface_seeding_reward}, we notice a disparity between the final results achieved and the performances of agents measured by the Tractometer. While the SAC and SAC Auto agents achieve the best performance and the highest final ASR, the DDPG agent often achieves the third best ASR without being the third best performing agent. Similarly, TRPO on the ISMRM2015 dataset using interface seeding achieves a low ASR but still performs well. This phenomenon was first observed by Theberge et al.~\cite{theberge2021} where it was observed that agents achieving the best return did not always yield the highest VC or OL rates. This is because agents are rewarded on a \emph{per step} basis. Therefore, agents who take more steps and produce longer streamlines obtain a larger sum of average reward, whereas these streamlines are not always more plausible anatomically. This reiterates that the reward function as formulated so far is a poor proxy for the agents' performance and needs to be adapted to reward streamlines based on their anatomical plausibility.

        Looking back on the results of Experiment 5~(c.f. section~\ref{sec:exp5_results}), we can infer a few key points on the learning process and the reward function. Adding a length component to the reward function does not seem to provide any consistent value to the agents. We hypothesize that, as agents already get rewarded for stepping in the WM, with more steps leading to higher returns, the bonus only serves as a scaling factor on the accumulated reward with no correlation to the desirability of actions produced. The scaling then only impacts the critics who have to evaluate the expected returns, with a bigger scaling inducing larger estimation errors, hampering the learning process.

        The GM bonus only provided improvements when agents got +10 for entering GM. As entering GM signals a stop to the tracking process, the bonus is obtained very sparsely (at most twice per streamline). Therefore, a low bonus may not have much of an impact on the learning process. Conversely, a very high bonus may again confuse critics because of the sudden reward spike. Empirically, we found that larger reward bonuses incur higher training losses for critics during training, which may negatively impact the learning process.
        
    \subsection{Voxel size vs step size}
    \label{sub:voxel_vs_step}
        
        When first tracking on the testing datasets for Experiments 1 and 2, we initially noticed agents performing well on the Flipped FiberCup dataset but extremely poorly on the TractoInferno dataset. After experimenting with the tracking parameters, we realized that the step size must be adjusted so that agents ``traverse'' the same number of voxels per step between the training and testing dataset. For example, the ISMRM2015 dataset (used for training) has a voxel size of 2mm isotropic whereas the TractoInferno dataset has a voxel size of 1mm isotropic (after processing through Tractoflow). This was not an issue when going from the FiberCup to the Flipped dataset as they have the same voxel size. As such, the step size on the TractoInferno dataset was halved from 0.75mm to 0.375mm.
        
        Previous machine learning-based tractography approaches, such as the Random-Forest by Neher et al.~\cites{neher_machine_2015, neher_fiber_2017} and \emph{Learn-to-Track}~\cite{poulin_learn_2017} did not encounter this problem, as they formulated their step-size \emph{per voxel}. In this work, we define the step-size in millimetres to be in-line with the classical formulation~\cite{neher2015strengths}.

    \subsection{Comparison with \emph{Track-to-Learn}}
    \label{sub:original_ours}
        Experiment 1 of this work roughly follows the same experimental procedure as in the original \emph{Track-to-Learn} paper. However, discrepancies can be noted in results mainly due to the following reasons: First, the number of episodes during the training procedure was variable in Track-to-Learn (ranging from 80 to 150) whereas it is fixed to 1000 in this work. As such, the agents may be able to fine-tune their policy better than in previous works. The selection of hyperparameters is also different in both works, where a Bayesian hyperparameter tuner was used in Track-to-Learn and a simple grid search is employed here. The Bayesian hyperparameter tuner considered smaller value ranges than the grid-search but could select more precise hyperparameter values.  
        
        The considered step-size might also have an effect. Although the framework was initially implemented using a step-size in voxels (much like prior work from Neher et al.~\cites{neher_machine_2015, neher_fiber_2017} and Poulin et al.~\cite{poulin_learn_2017}), the current work considers a step-size in millimetres. As such, in this work, tracking on the FiberCup (for example) is performed at a step-size a third of that in the original work, which might put agents at a disadvantage when navigating the crossings and turns of the dataset.   

        However, while some results are different from previously reported ones, some general conclusions still hold. From Experiments 1 and 2, we can infer that the good generalization performances of RL agents are still valid. For example, most agents perform similarly well on the FiberCup and Flipped dataset (which can be considered as a generalization to new bundle configurations) despite the Flipped dataset not being considered when training the agents. Likewise, we also see that agents who perform well on the ISMRM2015 dataset during training also reconstruct visually appealing bundles on the TractoInferno datasets, and agents that perform poorly on the ISMRM2015 phantom also give poor results \emph{in-vivo}. 

    \subsection{Discount factor}

        Looking at the selected hyperparameters throughout the experiments (cf. Appendices ~\ref{app:exp1_hyperparameters},~\ref{app:exp2_hyperparameters},~\ref{app:exp3_hyperparameters},~\ref{app:exp4_hyperparameters} and~\ref{app:exp5_hyperparameters}), we notice that the discount factor selected for most agents is quite low (from 0.5 to 0.85) compared to what is standard in the reinforcement learning literature (0.9 to 0.99)~\cites{fujimoto2018, haarnoja2018a, schulman2017a}. This is somewhat contradictory to what was reported by Theberge et al.~\cite{theberge2021}, as it was suggested that a higher discount factor led to generally higher VC rates. A few hypotheses can explain the differences. 

        First, as noted in section~\ref{sub:original_ours}, the experimental procedure is different in the presented work. Most importantly, the hyperparameters considered are limited and selected via grid-search, whereas~\cite{theberge2021} chose hyperparameters via Bayesian optimization and obtained arbitrary values. Furthermore, the lowest discounted factor considered was 0.85, instead of 0.5. Finally, only the TD3 algorithm was considered when plotting the discount parameter vs. different metrics.

        Lower discount factors can be beneficial when considering the reward-hacking propensity of agents. As noted by Theberge et al.~\cite{theberge2021}, agents may loop around in the WM in the hope of not reaching GM, making more steps, and therefore accumulating more reward. When looping, the agents take a ``reward hit'' temporarily, as the steps produced no longer align with the fODF peaks. However, this hit is balanced by the future rewards the agents will accumulate when it starts tracking again in the opposite directions. A lower discount factor, making the agents more greedy and less caring of the future reward, may make looping (and its temporary reward hit) a deterrent big enough to prevent reward hacking.

    \subsection{Bundle coverage}
    \label{sub:overlap}

        Experiments 1 and 2 present a comparison between the performances of RL agents and the PFT algorithm. While RL agents generally exhibit competitive or superior performance with respect to connection rates, they also exhibit a much lower OL. Streamline density maps~(c.f. figures~\ref{fig:streamline_density} and~\ref{fig:interface_streamline_density}) also show that agents tend to avoid the ``outside edges'' of the WM tracking mask, undoubtedly lowering the resulting OL.
        
        Because agents want to maximize their return, and therefore implicitly want to track for as long as possible (cf. Section~\ref{sec:components}), the hard constraint on streamline termination may encourage agents to ``avoid'' the WM boundary so as to minimize the risks of stepping out the WM mask and end the tracking procedure. 

        From the same two figures (cf. figures~\ref{fig:streamline_density} and~\ref{fig:interface_streamline_density}), one might notice that the streamlines density maps are non-zero outside the WM mask for tractograms produced by RL agents. This is because the last step produced by the agents (which might exit WM) is kept, contrary to classical tractography algorithms. This might increase the overreach computed by the Tractometer for RL agents. 

    \subsection{Future works}
    \label{sub:future_work}
        
        Several components of the \emph{Track-to-Learn} framework could be improved. For example, the input signal and the reward function could benefit from using asymmetric ODFs (aODF)~\cites{bastiani2017improved, karayumak2018asymmetric}. Using peaks computed from aODFs would remove the need for the absolute value function in the reward function, as peaks would now have a direction, and the asymmetric signal could help agents learn and disentangle the tracking direction.
        
        Partial volume maps, used as is~\cite{smith2012} or with a heuristic~\cite{girard2014}, could be integrated in the environment to ``soften'' the termination criteria and / or integrated into the reward function to help agents fill the volume of the WM better. Similarly, surface normals~\cite{st2018surface} could be used to reward agents by providing a termination bonus based on alignment with the last tracking step and the surface normal.
        
        Reinforcement learning agents typically sample actions from a multivariate normal distribution using their policy as mean and a parameterized standard deviation. However, in the context of tractography, actions taken by agents would be much better represented on a sphere than on an unconstrained 3D space. As such, a von Mises-Fischer distribution could be a sound alternative to a Normal distribution, as demonstrated by Wegmayr et al.~\cite{wegmayr2021entrack}. An alternative would be to make the agents output the polar and azimuthal angles of vectors in spherical coordinates, with the radius determined by the tractography step size.
        
        In a similar vein, all RL algorithms considered produce actions in a continuous domain. However, \emph{Deep Q-Learning}-based methods~\cites{mnih2015human, van2016deep, hessel2018rainbow} are very popular and successful methods that consider and produce actions in a discrete domain. These methods could be leveraged for tractography by, for example, selecting directions on a discrete sphere instead of producing a 3D vector directly.

        As noted in section~\ref{sub:voxel_vs_step}, the step size is constrained by the voxel size of the volume used during training. This enforces a specific step size that may not be desirable to the user when tracking on arbitrary data. This limitation can be addressed either by training the agents with multiple step sizes so that they can generalize to the user's preference, or by doing away with the step size as a whole and letting the agents make steps of arbitrary sizes.
        
        Most importantly, it is crucial that the reward function be improved in future work to truly represent what is expected from learning agents. The reward function as presented in \emph{Track-to-Learn} only rewards agents locally, whereas the metrics used to evaluate agents are global in nature. This makes the current reward function a poor proxy for evaluating the desirability of the reconstructed tractograms. Rewarding agents not for their alignment with local peaks, but instead for connecting two anatomically plausible regions would much better suit the tractography problem and prevent agents from abusing the reward function to track infinitely. 

    \section{Conclusion}
    \label{sec:conclusion}
        
        In this work, we evaluated what works and what does not work when training RL agents to reconstruct WM streamlines. To achieve this goal, we tested various implementations and assessed their impact on the performance of learning agents. We found several components that are deemed critical, while others can be modified to best fit the user's preference. As such, we offer a formulation of \emph{Track-to-Learn} that follows the results of the experiments presented in this work: 
        \begin{enumerate}
        
        \item \textbf{Choose SAC Auto} as an RL algorithm to train the agents. 
        
        \item \textbf{Use fODFs, no WM mask, two previous directions} as a state formulation.
        
        \item \textbf{Seed from the WM/GM interface} if bundle coverage is not your priority.
        
        \item \textbf{Train on the ISMRM2015 dataset} to generalize to \emph{in-vivo} subjects.
        
        \item \textbf{Perform extensive hyperparameter search} before training to select the agent that best fits your needs.
        
        \item \textbf{Vary the step size} after training to accommodate differences in voxel sizes.

        \item \textbf{Consider adding a slight termination bonus} to the reward function to improve the performance of your agents.
        \end{enumerate}
        We hope future researchers eager to try reinforcement learning methods for tractography will make use of the information presented above so as not to waste their time with ill-formed implementations of the Track-to-Learn framework. 
       
\section*{Acknowledgments}

This research was enabled in part by the support provided by Calcul Québec (\url{https://www.calculquebec.ca/}) and Compute Canada (\url{www.computecanada.ca}). 

\bibliographystyle{unsrt}
\bibliography{references}
\include{appendix}
\end{document}

%% file: appendix.tex
\appendix
\section{Reinforcement learning algorithms}
\label{app:agents}
    We present below all the algorithms considered for the experiments. To fit the tractography formulation of RL, every algorithm was reimplemented by the authors using Pytorch~\cite{paszke2019pytorch}. To validate their implementation, each algorithm was trained until convergence on the 'InvertedPendulumMuJoCoEnv-v0' environment of PyBullet-Gym~\cite{benelot2018}. Hyperparameters with single values in further tables were chosen empirically. 
        
    \paragraph*{REINFORCE} 
        \begin{table}[!tbhp] 
            \caption{Hyperparameters considered for REINFORCE. Single values were found empirically.}
            \label{tab:reinforce_hyperparams}
            \renewcommand{\arraystretch}{1.2}
            \begin{tabularx}{\linewidth}{X|*4{>{\arraybackslash}X}}
                Hyperparameter                  & Values                              \\ \hline 
                Learning rate $\eta$            & [1e-5, 5e-5, 1e-4, 5e-4, 1e-3, 5e-3] \\ 
                Discount ($\gamma$)                & [0.5, 0.75, 0.85, 0.90, 0.95, 0.99] \\
                Entropy bonus                   & 0.001 \\ \hline
            \end{tabularx}       
        \end{table}
    
        The use of neural networks in reinforcement learning has allowed the advent of policy gradient methods, the first being REINFORCE~\cite{williams1992} (also called Vanilla Policy Gradient, VPG) , where the policy $\pi_\theta$ is directly parametrized by $\theta$. The algorithm uses the policy-gradient theorem~\cite{sutton2018} to update its weights using gradient ascent to maximize the future expected reward:
        \begin{equation}
            \theta = \frac{1}{T}\sum_{t=0}^T\theta + \eta{G_t}\log\pi_\theta(a_t|s_t),
        \end{equation}
        where $\pi(a|s)$ is the probability of the policy $\pi$ taking action $a$ at state $s$ and $G_t$ is the expected return. The neural network representing the policy outputs the mean of a Gaussian distribution and a separate parameter handles the standard deviation. Table~\ref{tab:reinforce_hyperparams} displays the hyperparameters considered for REINFORCE.
        
    \paragraph*{A2C} 
        \begin{table}[thbp]
        \caption{Hyperparameters considered for A2C. Single values were found empirically.}
        \label{tab:a2c_hyperparams}
        \renewcommand{\arraystretch}{1.2}
        \begin{tabularx}{\linewidth}{X|*2{>{\arraybackslash}X}}
            Hyperparameter            & Values \\ \hline 
            Learning rate $\eta$      & [1e-5, 5e-5, 1e-4, 5e-4, 1e-3, 5e-3] \\ 
            Discount ($\gamma$)          & [0.5, 0.75, 0.85, 0.90, 0.95, 0.99] \\
            Entropy bonus             & 0.001 \\
            Lambda ($\lambda$)        & 0.95 \\ \hline
        \end{tabularx}       
        \end{table}

        The Advantage Actor-Critic (A2C) is a synchronous variant of the Asynchronous-Advantage Actor-Critic (A3C) algorithm~\cite{mnih2016} built upon the Actor-Critic framework. The Actor-Critic framework improves on policy-gradient methods by using a learned baseline to reduce the variability in the policy-gradient update. Specifically, the framework uses a critic network to predict the value function for the policy network. The update rule then becomes:
        \begin{equation}
            \theta = \frac{1}{T}\sum_{t=0}^T\theta + \eta{A_t}\log\pi_\theta(a_t|s_t),
        \end{equation}
        where $A_t$ corresponds to the \emph{advantage function}:
        \begin{equation}
            A_t = Q^\pi_\phi(s_t,a_t) - V^\pi_\phi(s_t).
        \end{equation}
         $A_t$ then measures how better (or worse) was the reward obtained by the agent compared to the expected value of the state, which leads to less variable updates than using the actual returns. $V^\pi_\phi$ can be trained used the mean squared error between $G_t$ and $V^\pi_\phi(s_t)$.
        
        We use separate networks for the policy and the critic and the policy is again represented as the mean of a Gaussian distribution, with the standard deviation being a separate parameter. Generalized Advantage Estimation (GAE)~\cite{schulman2015} is used to compute the advantage function. We also use an entropy bonus for the policy update and the RMSProp optimizer is again used to update the weights of the networks. The same learning rate is used for the actor and the critic networks. Table~\ref{tab:a2c_hyperparams} presents the hyperparameter ranges considered for A2C.
       
     \paragraph*{TRPO} 
        \begin{table}[thbp]
        \caption{Hyperparameters considered for TRPO. Single values were found empirically.}
        \label{tab:trpo_hyperparams}
        \renewcommand{\arraystretch}{1.2}
        \begin{tabularx}{\linewidth}{X|*4{>{\arraybackslash}X}}
            Hyperparameter            & Values \\ \hline 
            Learning rate $\eta$      & [1e-5, 5e-5, 1e-4, 5e-4, 1e-3, 5e-3] \\ 
            Discount ($\gamma$)          & [0.5, 0.75, 0.85, 0.90, 0.95, 0.99] \\
            Entropy bonus             & 0.001 \\
            Lambda ($\lambda$)        & 0.95 \\
            Nb. backtracks            & 10 \\
            Backtracking coef.        & 0.5 \\
            Nb. of epochs per batch ($K$)  & 3 \\
            Delta ($\delta$)          & [0.001, 0.01, 0.1] \\ \hline
        \end{tabularx}      

        \end{table}
    
        The Trust-Region Policy Optimization (TRPO) algorithm improves on A2C by taking the biggest gradient step possible while constraining the new policy to a trust-region around the old one. The trust-region constraint is enforced by adding a Kullback-Liebler divergence constraint between the old and new policies. The steepest gradient step is obtained by computing the \emph{natural gradient}~\cite{kakade2001} instead of the standard policy gradient and by performing line-search~\cite{armijo1966} on the gradient step. Natural Gradient methods rely on computing the Fischer matrix to reshape the gradient landscape to perform updates in the "policy-space" instead of the "parameter space". Because computing the inverse of the Fischer matrix for the natural gradient is computationally expensive, TRPO instead uses the conjugate gradient~\cite{hestenes1952} algorithm and automatic differentiation packages to obtain an approximation.
                        
        We again use separate networks for the policy and the critic, the policy is formulated the same as previous policy gradient methods and GAE is again used to compute the advantage. The critic is updated using the Adam~\cite{kingma2014} optimizer and an entropy bonus is again used in the policy's update. A "delta" parameter controls how much the policy can diverge from the previous at each optimization step. The policy is updated using a backtracking line-search applied for a maximum number of steps. 
        
        Table~\ref{tab:trpo_hyperparams} offers the hyperparameters considered for the TRPO algorithm.
                      
    \paragraph*{ACKTR} 
        \begin{table}[thbp]
            \caption{Hyperparameters considered for ACKTR. Single values were found empirically.}
            \label{tab:acktr_hyperparams}
            \renewcommand{\arraystretch}{1.2}
            \begin{tabularx}{\linewidth}{X|*4{>{\arraybackslash}X}}
                Hyperparameter            & Values \\ \hline 
                Learning rate ($\eta$)    & [0.01, 0.1, 0.15, 0.2, 0.25] \\ 
                Discount ($\gamma$)          & [0.5, 0.75, 0.85, 0.9, 0.95, 0.99] \\
                Entropy bonus             & 0.001 \\
                Lambda ($\lambda$)        & 0.95 \\
                Delta ($\delta$)          & [0.0001, 0.0005, 0.001, 0.005, 0.01] \\\hline
            \end{tabularx}    
        \end{table}
        Much like TRPO, the Actor-Critic using Kronecker-Factored Trust Region (ACKTR)~\cite{wu2017} algorithm also builds on A2C and improves it by optimizing the policy using the natural gradient instead of the standard policy gradient. ACKTR uses Kronecker-factored approximated curvature (K-FAC) optimization~\cite{martens2015optimizing} to efficiently compute the natural gradient step. Moreover, the gradient step is also constrained by the Kullback-Liebler divergence between the old policy and the new, preventing the training process from diverging too much according to a hyperparameter ($\delta$). The algorithm is implemented the same way as A2C, but the K-FAC optimizer is used for both the actor and the critic networks. Table~\ref{tab:acktr_hyperparams} presents the hyperparameters considered for ACKTR.
   

    \paragraph*{PPO}
        \begin{table}[thbp]
        \caption{Hyperparameters considered for PPO. Single values were found empirically.}
        \label{tab:ppo_hyperparams}
        \renewcommand{\arraystretch}{1.2}
        \begin{tabularx}{\linewidth}{X|*2{>{\arraybackslash}X}}
            Hyperparameter            & Values \\ \hline 
            Learning rate $\eta$      & [1e-5, 5e-5, 1e-4, 5e-4, 1e-3, 5e-3] \\ 
            Discount ($\gamma$)          & [0.5, 0.75, 0.85, 0.9, 0.95, 0.99] \\
            Entropy bonus             & 0.001 \\
            Lambda ($\lambda$)        & 0.95 \\
            Clipping parameter ($\epsilon$) & [0.05, 0.1, 0.2] \\
            Nb. of epochs per batch ($K$)   & 30 \\ \hline
        \end{tabularx} 
            
        \end{table}
        
        The Proximal Policy-Optimization (PPO) family of algorithms~\cite{schulman2017a} improves on TRPO by removing the computationally expensive second-order optimization step while maintaining its core idea: making the biggest gradient step possible constrained by a "trust-region" around the old policy. In its most popular form (PPO-Clip), PPO clips its objective function "pessimistically" so that it does not diverge too much from its old policy. PPO also improves on TRPO by training multiple times on the same transitions, improving sample efficiency and by training on partial episodes. Our implementation is otherwise similar to A2C but the Adam optimizer is used for both the actor and the critic. Table~\ref{tab:ppo_hyperparams} presents the hyperparameters considered for PPO.
       
    \paragraph*{DDPG} 
        \begin{table}[thbp]
        \caption{Hyperparameters considered for DDPG.}
        \label{tab:ddpg_hyperparameters}
        \renewcommand{\arraystretch}{1.2}
        \begin{tabularx}{\linewidth}{X|*2{>{\arraybackslash}X}}
            Hyperparameter            & Values \\ \hline 
            Learning rate ($\eta$)    &  [1e-5, 5e-5, 1e-4, 5e-4, 1e-3, 5e-3] \\ 
            Discount ($\gamma$)          & [0.5, 0.75, 0.85, 0.9, 0.95, 0.99] \\
            Action noise ($\sigma$)   & [0.2, 0.25, 0.3, 0.35, 0.4] \\ \hline
        \end{tabularx}
        
        \end{table}
        
        Diverging from the policy gradient methods described previously, the Deep Deterministic Policy Gradient~(DDPG)~\cite{lillicrap2015} algorithm instead builds on Q-learning and proposes a extension for continuous control. Because Q-Learning relies on evaluating the Q-function for all actions at every state encountered, it would be intractable to apply it to environment with continuous actions. Instead, the algorithm relies on the Deterministic Policy Gradient theorem~\cite{silver2014} to update its policy, which stipulates that since actions are continuous, the Q-function is differentialble w.r.t to the actions of the policy.

        The policy then updates its weights using $Q_\phi$, the critic, and the update becomes
        \begin{equation}
            \theta = \theta + \eta\nabla_{a}Q_{\phi}(s,a)\nabla_{\theta}\pi_{\theta}(s)|_{a = \pi(s)},
        \end{equation}
        where $\pi$ is now a deterministic function $a \leftarrow \pi(s)$. Noise $\sim \mathcal{N}(0, \sigma)$ is added to actions at training time to allow the algorithm to explore its environment. Because DDPG is an \emph{off-policy} algorithm (contrary to previous methods which were considered \emph{on-policy}), it uses a persistent replay buffer to store past transitions and randomly samples the buffer . The policy now outputs the actions directly instead of being formulated as a Gaussian distribution. All neural networks are trained using the Adam optimizer. Hyperparameters considered are available in table~\ref{tab:ddpg_hyperparameters}.
   
    \paragraph*{TD3} 
        \begin{table}[thbp]
        \caption{Hyperparameters considered for TD3.}
        \label{tab:td3_hyperparameters}
        \renewcommand{\arraystretch}{1.2}
        \begin{tabularx}{\linewidth}{X|*2{>{\arraybackslash}X}}
            Hyperparameter            & Values \\ \hline 
            Learning rate ($\eta$)    &  [1e-5, 5e-5, 1e-4, 5e-4, 1e-3, 5e-3] \\ 
            Discount ($\gamma$)          & [0.5, 0.75, 0.85, 0.9, 0.95, 0.99] \\
            Action noise ($\sigma$)   & [0.2, 0.25, 0.3, 0.35, 0.4] \\ \hline
        \end{tabularx}       

        \end{table}
        The Twin Delayed Deep Deterministic policy gradient~(TD3)~\cite{fujimoto2018} algorithm builds on DDPG and adds a second Q network to fight overestimation of the value of pairs $(s,a)$ which leads to suboptimal policies. TD3 also adds noise to the update of the Q-function to fight overfitting to certain actions, as the noise will smooth the update and force similar actions to produce similar value. Finally, the TD3 algorithm delays the update of the policy to allow the Q-function to be trained properly before its target moves. Our implementation of TD3 is otherwise identical to our implementation of DDPG. Hyperparameter ranges considered are available in table~\ref{tab:td3_hyperparameters}.
   
    \paragraph*{SAC and SAC Auto}
        \begin{table}[thbp]
        \caption{Hyperparameters considered for SAC.}
        \label{tab:sac_hyperparameters}
        \renewcommand{\arraystretch}{1.2}
        \begin{tabularx}{\linewidth}{X|*2{>{\arraybackslash}X}}
            Hyperparameter            & Values \\ \hline 
            Learning rate ($\eta$)    &  [1e-5, 5e-5, 1e-4, 5e-4, 1e-3, 5e-3] \\ 
            Discount ($\gamma$)          & [0.5, 0.75, 0.85, 0.9, 0.95, 0.99] \\
            Alpha ($\alpha$)          & [0.075, 0.1, 0.15, 0.2, 0.3] \\ \hline
        \end{tabularx}      
        

        \caption{Hyperparameters considered for SAC Auto.}
        \label{tab:sac_auto_hyperparameters}
        \renewcommand{\arraystretch}{1.2}
        \begin{tabularx}{\linewidth}{X|*4{>{\arraybackslash}X}}
            Hyperparameter            & Values \\ \hline 
            Learning rate ($\eta$)    &  [1e-5, 5e-5, 1e-4, 5e-4, 1e-3, 5e-3] \\ 
            Discount ($\gamma$)          & [0.5, 0.75, 0.85, 0.9, 0.95, 0.99] \\ \hline
        \end{tabularx}
        
        \end{table}
        The Soft Actor-Critic~(SAC)~\cite{haarnoja2018a} algorithm also builds on DDPG but uses the Maximum-Entropy Reinforcement Learning framework~\cite{eysenbach2021}, which adds an entropy coefficient to the standard RL objective:
        \begin{equation}
            J = \mathbb{E}_{s,a \sim \pi}[R(s,a) + \alpha\mathcal{H}(\pi_\theta(\cdot|s)],
        \end{equation}

        where $\alpha$ is a hyperparameter controlling the importance of the entropy. Maximizing the entropy alongside the standard RL objective allows the agent to explore better and be robust to situations where multiple strategies may be optimal. Like TD3, SAC also uses two critics to fight overestimation. SAC formulates the policy as a neural network which outputs the mean and the standard deviation of a Gaussian distribution. Otherwise, our implementation is identical to TD3 and DDPG.
       
        SAC can be sensible to the $\alpha$ hyperparameter. In a subsequent paper~\cite{haarnoja2018b}, the authors improved the algorithm by automatically tuning the $\alpha$ instead of fixing it for the whole training regime. We call this algorithm SAC Auto in our tests. Hyperparameters considered for SAC and SAC Auto are available in tables~\ref{tab:sac_hyperparameters} and \ref{tab:sac_auto_hyperparameters}.
       
\section{Hyperparameter search for experiment 1}\label{app:exp1_hyperparameters}

    Tables \ref{tab:exp1_reinforce_hyperparams}, \ref{tab:exp1_a2c_hyperparams}, \ref{tab:exp1_trpo_hyperparams}, \ref{tab:exp1_acktr_hyperparams}, \ref{tab:exp1_ppo_hyperparams}, \ref{tab:exp1_ddpg_hyperparams}, \ref{tab:exp1_td3_hyperparams}, \ref{tab:exp1_sac_hyperparams}, \ref{exp1_tab:exp1_sac_auto_hyperparams} hyperparameters selected for each agent in experiment 1. 
 
    \begin{table}[!tbhp] 
        \caption{Hyperparameters chosen for REINFORCE in experiment 1}
        \label{tab:exp1_reinforce_hyperparams}
        \renewcommand{\arraystretch}{1.2}
        \begin{tabularx}{\linewidth}{X|*3{>{\arraybackslash}X}}
            Hyperparameter                  & FiberCup & ISMRM2015 \\ \hline 
            Learning rate $\eta$            & 0.0005    & 0.0001   \\ 
            Discount ($\gamma$)                & 0.75     &  0.5      \\ \hline
        \end{tabularx}       

        \caption{Hyperparameters selected for A2C in experiment 1}
        \label{tab:exp1_a2c_hyperparams}
        \renewcommand{\arraystretch}{1.2}
        \begin{tabularx}{\linewidth}{X|*4{>{\arraybackslash}X}}
            Hyperparameter           & FiberCup & ISMRM2015 \\ \hline 
            Learning rate ($\eta$)   & 0.00001 & 0.00001    \\ 
            Discount ($\gamma$)         & 0.5     & 0.95        \\ \hline
        \end{tabularx}       

        \caption{Hyperparameters selected for TRPO in experiment 1}
        \label{tab:exp1_trpo_hyperparams}
        \renewcommand{\arraystretch}{1.2}
        \begin{tabularx}{\linewidth}{X|*4{>{\arraybackslash}X}}
            Hyperparameter            & FiberCup & ISMRM2015 \\ \hline 
            Learning rate ($\eta$)    & 0.001  & 0.0005 \\ 
            Discount ($\gamma$)          & 0.75   & 0.5    \\
            Delta ($\delta$)          & 0.001  & 0.01   \\ \hline
        \end{tabularx}       

        \caption{Hyperparameters selected for ACKTR in experiment 1}
        \label{tab:exp1_acktr_hyperparams}
        \renewcommand{\arraystretch}{1.2}
        \begin{tabularx}{\linewidth}{X|*4{>{\arraybackslash}X}}
            Hyperparameter            & FiberCup & ISMRM2015 \\ \hline 
            Learning rate ($\eta$)    & 0.01   & 0.01 \\ 
            Discount ($\gamma$)          & 0.5    & 0.5 \\
            Delta ($\delta$)          & 0.001  & 0.01 \\ \hline
        \end{tabularx}       

        \caption{Hyperparameters selected for PPO in experiment 1}
        \label{tab:exp1_ppo_hyperparams}
        \renewcommand{\arraystretch}{1.2}
        \begin{tabularx}{\linewidth}{X|*4{>{\arraybackslash}X}}
            Hyperparameter                  & FiberCup & ISMRM2015   \\ \hline 
            Learning rate ($\eta$)          & 0.00005  & 0.005 \\ 
            Discount ($\gamma$)                & 0.5      & 0.5  \\
            Clipping parameter ($\epsilon$) & 0.05     & 0.2  \\ \hline
        \end{tabularx}       
    \end{table}
    
    \begin{table}[!tbhp]

        \caption{Hyperparameters selected for DDPG in experiment 1}
        \label{tab:exp1_ddpg_hyperparams}
        \renewcommand{\arraystretch}{1.2}
        \begin{tabularx}{\linewidth}{X|*4{>{\arraybackslash}X}}
            Hyperparameter            & FiberCup  & ISMRM2015 \\ \hline 
            Learning rate ($\eta$)    & 0.0005  & 0.00001 \\ 
            Discount ($\gamma$)          & 0.95    & 0.5     \\
            Action noise ($\sigma$)   & 0.35    & 0.35    \\ \hline
        \end{tabularx}


        \caption{Hyperparameters selected for TD3 in experiment 1}
        \label{tab:exp1_td3_hyperparams}
        \renewcommand{\arraystretch}{1.2}
        \begin{tabularx}{\linewidth}{X|*4{>{\arraybackslash}X}}
            Hyperparameter            & FiberCup & ISMRM2015\\ \hline 
            Learning rate ($\eta$)    & 0.0005 & 0.00001 \\ 
            Discount ($\gamma$)          & 0.9 & 0.85 \\
            Action noise ($\sigma$)   & 0.4 & 0.4 \\ \hline
        \end{tabularx}       

        
        \caption{Hyperparameters selected for SAC in experiment 1}
        \label{tab:exp1_sac_hyperparams}
        \renewcommand{\arraystretch}{1.2}
        \begin{tabularx}{\linewidth}{X|*4{>{\arraybackslash}X}}
            Hyperparameter            & FiberCup & ISMRM2015 \\ \hline 
            Learning rate ($\eta$)    & 0.0005 & 0.0001 \\ 
            Discount ($\gamma$)          & 0.85   & 0.75 \\
            Alpha ($\alpha$)          & 0.15   & 0.1 \\ \hline
        \end{tabularx}      
        

        \caption{Hyperparameters selected for SAC Auto in experiment 1}
        \label{exp1_tab:exp1_sac_auto_hyperparams}
        \renewcommand{\arraystretch}{1.2}
        \begin{tabularx}{\linewidth}{X|*4{>{\arraybackslash}X}}
            Hyperparameter            & FiberCup & ISMRM2015 \\ \hline 
            Learning rate ($\eta$)    & 0.0005 & 0.0001      \\ 
            Discount ($\gamma$)          & 0.5   & 0.5          \\ \hline
        \end{tabularx}       
        
        
    \end{table}
    
\section{Full results for experiment 1 - WM seeding}\label{app:exp1_results}

         Table \ref{tab:agents_fibercup} presents the Tractometer scores for the agents trained and tested on the FiberCup dataset. Table \ref{tab:agents_flipped} presents the Tractometer scores for the agents trained on the FiberCup dataset and tested on the Flipped dataset. Table \ref{tab:agents_ismrm} presents the Tractometer scores for the agents trained and tested on the ISMRM2015 dataset.
        
        \begin{table}[!tbhp] 
        \caption{Tractometer scores for agents trained and tested on the FiberCup dataset using WM seeding.}
        \label{tab:agents_fibercup}
        \renewcommand{\arraystretch}{1.2}
        \begin{tabularx}{\columnwidth}{>{\hsize=.1\hsize}X|*4{>{\centering\arraybackslash}X}|}
        \cline{2-5}
        
        & \multicolumn{1}{c }{VC $\% \uparrow$}
        & \multicolumn{1}{c}{VB $\uparrow$} 
        & \multicolumn{1}{c }{IC $\% \downarrow$}
        & \multicolumn{1}{c|}{IB $\downarrow$} \\ \hline
        
        
        \multicolumn{ 1 }{|l|}{ PFT } & 27.95 $\pm$ 0.19 & 7.00 $\pm$ 0.00 & 13.65 $\pm$ 0.14 & 1.20 $\pm$ 0.40 \\ \hline 
        \multicolumn{ 1 }{|l|}{ VPG } & 21.31 $\pm$ 17.46 & 4.20 $\pm$ 3.43 & 15.42 $\pm$ 13.05 & 1.00 $\pm$ 0.89  \\
        \multicolumn{ 1 }{|l|}{ A2C } & 23.51 $\pm$ 4.42 & 6.00 $\pm$ 0.63 & 15.64 $\pm$ 5.88 & 2.00 $\pm$ 0.63  \\
        \multicolumn{ 1 }{|l|}{ ACKTR } & 63.32 $\pm$ 1.36 & 7.00 $\pm$ 0.00 & 17.65 $\pm$ 2.13 & 2.20 $\pm$ 0.75  \\
        \multicolumn{ 1 }{|l|}{ TRPO } & 27.03 $\pm$ 14.08 & 4.80 $\pm$ 1.47 & 35.69 $\pm$ 19.40 & 2.60 $\pm$ 0.49  \\
        \multicolumn{ 1 }{|l|}{ PPO } & 60.00 $\pm$ 6.43 & 7.00 $\pm$ 0.00 & 18.65 $\pm$ 6.01 & 2.20 $\pm$ 0.40  \\
        \multicolumn{ 1 }{|l|}{ DDPG } & 41.17 $\pm$ 10.49 & 6.20 $\pm$ 0.40 & 24.81 $\pm$ 5.23 & 2.20 $\pm$ 0.40  \\
        \multicolumn{ 1 }{|l|}{ TD3 } & 44.67 $\pm$ 9.15 & 7.00 $\pm$ 0.00 & 28.98 $\pm$ 9.46 & 2.60 $\pm$ 0.49  \\
        \multicolumn{ 1 }{|l|}{ SAC } & 62.73 $\pm$ 3.55 & 7.00 $\pm$ 0.00 & 29.18 $\pm$ 3.67 & 1.80 $\pm$ 0.75  \\
        \multicolumn{ 1 }{|l|}{ SAC Auto } & 74.72 $\pm$ 4.19 & 7.00 $\pm$ 0.00 & 13.59 $\pm$ 4.36 & 1.40 $\pm$ 0.49  \\
        \hline
        
        &\multicolumn{1}{c}{OL $\% \uparrow$}
        &\multicolumn{1}{c}{OR $\% \downarrow$}
        &\multicolumn{1}{c}{F1 $\% \uparrow$}
        &\multicolumn{1}{c|}{NC $\downarrow$} \\ \hline   
        
        
        \multicolumn{ 1 }{|l|}{ PFT } & 99.41 $\pm$ 0.02 & 34.26 $\pm$ 0.49 & 78.94 $\pm$ 0.37 & 58.40 $\pm$ 0.13 \\ \hline
        \multicolumn{ 1 }{|l|}{ VPG } & 51.61 $\pm$ 42.14 & 11.00 $\pm$ 9.10 & 49.53 $\pm$ 40.45 & 63.27 $\pm$ 30.14  \\
        \multicolumn{ 1 }{|l|}{ A2C } & 62.21 $\pm$ 6.73 & 18.00 $\pm$ 1.28 & 61.52 $\pm$ 7.16 & 60.84 $\pm$ 8.53  \\
        \multicolumn{ 1 }{|l|}{ ACKTR } & 89.91 $\pm$ 0.55 & 19.78 $\pm$ 0.79 & 84.39 $\pm$ 0.42 & 19.03 $\pm$ 0.98  \\
        \multicolumn{ 1 }{|l|}{ TRPO } & 49.49 $\pm$ 21.41 & 12.55 $\pm$ 6.30 & 48.76 $\pm$ 18.00 & 37.28 $\pm$ 32.93  \\
        \multicolumn{ 1 }{|l|}{ PPO } & 84.01 $\pm$ 4.54 & 18.04 $\pm$ 1.05 & 81.72 $\pm$ 3.49 & 21.36 $\pm$ 2.40  \\
        \multicolumn{ 1 }{|l|}{ DDPG } & 59.65 $\pm$ 10.08 & 14.10 $\pm$ 2.06 & 63.11 $\pm$ 8.46 & 34.02 $\pm$ 7.79  \\
        \multicolumn{ 1 }{|l|}{ TD3 } & 75.76 $\pm$ 6.54 & 13.30 $\pm$ 2.35 & 79.36 $\pm$ 4.59 & 26.35 $\pm$ 10.84  \\
        \multicolumn{ 1 }{|l|}{ SAC } & 75.27 $\pm$ 1.24 & 14.66 $\pm$ 1.15 & 78.32 $\pm$ 1.12 & 8.09 $\pm$ 0.26  \\
        \multicolumn{ 1 }{|l|}{ SAC Auto } & 83.32 $\pm$ 0.75 & 15.46 $\pm$ 0.74 & 83.56 $\pm$ 0.70 & 11.69 $\pm$ 0.55  \\ \hline

        \end{tabularx}       

        \caption{Tractometer scores for agents trained on the FiberCup dataset and tested on the Flipped dataset using WM seeding.}
        \label{tab:agents_flipped}
        \renewcommand{\arraystretch}{1.1}
        \begin{tabularx}{\columnwidth}{>{\hsize=.1\hsize}X|*4{>{\centering\arraybackslash}X}|}
        \cline{2-5}
        
        & \multicolumn{1}{c }{VC $\% \uparrow$}
        & \multicolumn{1}{c}{VB $\uparrow$} 
        & \multicolumn{1}{c }{IC $\% \downarrow$}
        & \multicolumn{1}{c|}{IB $\downarrow$} \\ \hline
        
        \multicolumn{ 1 }{|l|}{ PFT } & 26.92 $\pm$ 0.19 & 7.00 $\pm$ 0.00 & 14.90 $\pm$ 0.10 & 1.00 $\pm$ 0.00 \\ \hline
        \multicolumn{ 1 }{|l|}{ VPG } & 18.82 $\pm$ 15.57 & 4.20 $\pm$ 3.43 & 15.18 $\pm$ 12.46 & 1.20 $\pm$ 0.75  \\
        \multicolumn{ 1 }{|l|}{ A2C } & 17.00 $\pm$ 5.67 & 6.00 $\pm$ 1.10 & 20.67 $\pm$ 3.55 & 1.60 $\pm$ 0.49  \\
        \multicolumn{ 1 }{|l|}{ ACKTR } & 58.78 $\pm$ 3.24 & 7.00 $\pm$ 0.00 & 26.11 $\pm$ 2.92 & 2.40 $\pm$ 0.49  \\
        \multicolumn{ 1 }{|l|}{ TRPO } & 32.41 $\pm$ 19.50 & 5.40 $\pm$ 0.80 & 26.53 $\pm$ 11.70 & 2.40 $\pm$ 0.80  \\
        \multicolumn{ 1 }{|l|}{ PPO } & 58.20 $\pm$ 2.71 & 7.00 $\pm$ 0.00 & 22.25 $\pm$ 2.56 & 1.80 $\pm$ 0.75  \\
        \multicolumn{ 1 }{|l|}{ DDPG } & 27.86 $\pm$ 5.71 & 5.60 $\pm$ 1.02 & 30.35 $\pm$ 6.37 & 2.00 $\pm$ 0.63  \\
        \multicolumn{ 1 }{|l|}{ TD3 } & 40.67 $\pm$ 9.17 & 7.00 $\pm$ 0.00 & 37.06 $\pm$ 6.96 & 2.60 $\pm$ 0.80  \\
        \multicolumn{ 1 }{|l|}{ SAC } & 66.18 $\pm$ 5.09 & 7.00 $\pm$ 0.00 & 24.95 $\pm$ 4.76 & 2.40 $\pm$ 0.49  \\
        \multicolumn{ 1 }{|l|}{ SAC Auto } & 78.90 $\pm$ 1.54 & 7.00 $\pm$ 0.00 & 8.72 $\pm$ 1.41 & 2.00 $\pm$ 0.00  \\
        \hline
        
        &\multicolumn{1}{c}{OL $\% \uparrow$}
        &\multicolumn{1}{c}{OR $\% \downarrow$}
        &\multicolumn{1}{c}{F1 $\% \uparrow$}
        &\multicolumn{1}{c|}{NC $\downarrow$} \\ \hline   
         
        \multicolumn{ 1 }{|l|}{ PFT } & 99.51 $\pm$ 0.07 & 33.15 $\pm$ 0.37 & 79.84 $\pm$ 0.27 & 58.17 $\pm$ 0.12 \\ \hline
        \multicolumn{ 1 }{|l|}{ VPG } & 50.69 $\pm$ 41.47 & 10.50 $\pm$ 8.57 & 48.96 $\pm$ 40.04 & 66.00 $\pm$ 27.93  \\
        \multicolumn{ 1 }{|l|}{ A2C } & 59.28 $\pm$ 13.79 & 16.54 $\pm$ 2.92 & 59.66 $\pm$ 13.72 & 62.33 $\pm$ 7.08  \\
        \multicolumn{ 1 }{|l|}{ ACKTR } & 90.38 $\pm$ 0.81 & 18.22 $\pm$ 0.68 & 85.57 $\pm$ 0.72 & 15.11 $\pm$ 0.41  \\
        \multicolumn{ 1 }{|l|}{ TRPO } & 53.22 $\pm$ 20.29 & 13.31 $\pm$ 5.48 & 54.00 $\pm$ 16.05 & 41.06 $\pm$ 30.05  \\
        \multicolumn{ 1 }{|l|}{ PPO } & 85.69 $\pm$ 1.81 & 17.28 $\pm$ 0.67 & 83.84 $\pm$ 1.04 & 19.55 $\pm$ 2.30  \\
        \multicolumn{ 1 }{|l|}{ DDPG } & 55.50 $\pm$ 9.44 & 11.59 $\pm$ 3.17 & 59.34 $\pm$ 9.89 & 41.79 $\pm$ 10.10  \\
        \multicolumn{ 1 }{|l|}{ TD3 } & 76.76 $\pm$ 5.69 & 12.30 $\pm$ 1.19 & 80.80 $\pm$ 3.53 & 22.27 $\pm$ 7.74  \\
        \multicolumn{ 1 }{|l|}{ SAC } & 78.40 $\pm$ 1.60 & 13.93 $\pm$ 0.85 & 79.90 $\pm$ 1.01 & 8.88 $\pm$ 0.41  \\
        \multicolumn{ 1 }{|l|}{ SAC Auto } & 87.40 $\pm$ 0.87 & 17.75 $\pm$ 1.87 & 84.29 $\pm$ 0.59 & 12.37 $\pm$ 0.46  \\
        \hline
        
        \end{tabularx}       
        \end{table}           
         
        \begin{table}[!thp] 
        \caption{Tractometer scores for agents trained and tested on the ISMRM2015 dataset using WM seeding.}
        \label{tab:agents_ismrm}
        \renewcommand{\arraystretch}{1.1}
        \begin{tabularx}{\columnwidth}{>{\hsize=.1\hsize}X|*4{>{\centering\arraybackslash}X}|}
        \cline{2-5}
        
        & \multicolumn{1}{c }{VC $\% \uparrow$}
        & \multicolumn{1}{c}{VB $\uparrow$} 
        & \multicolumn{1}{c }{IC $\% \downarrow$}
        & \multicolumn{1}{c|}{IB $\downarrow$} \\ \hline
        
        
        \multicolumn{1}{|l|}{PFT} & 61.00 $\pm$ 0.13 & 24.00 $\pm$ 0.00 & 30.10 $\pm$ 0.11 & 127.60 $\pm$ 3.01 \\ \hline
        \multicolumn{ 1 }{|l|}{ VPG } & 40.87 $\pm$ 14.31 & 22.80 $\pm$ 0.40 & 49.01 $\pm$ 8.99 & 133.80 $\pm$ 14.92 \\
        \multicolumn{ 1 }{|l|}{ A2C } & 10.33 $\pm$ 7.29 & 15.80 $\pm$ 7.25 & 47.37 $\pm$ 12.72 & 78.40 $\pm$ 34.60 \\
        \multicolumn{ 1 }{|l|}{ ACKTR } & 68.78 $\pm$ 1.09 & 23.00 $\pm$ 0.00 & 27.47 $\pm$ 1.12 & 166.00 $\pm$ 3.03 \\
        \multicolumn{ 1 }{|l|}{ TRPO } & 63.51 $\pm$ 2.18 & 23.00 $\pm$ 0.00 & 31.76 $\pm$ 2.37 & 147.00 $\pm$ 3.41 \\
        \multicolumn{ 1 }{|l|}{ PPO } & 55.26 $\pm$ 27.46 & 19.20 $\pm$ 7.60 & 26.73 $\pm$ 2.12 & 108.60 $\pm$ 46.59 \\
        \multicolumn{ 1 }{|l|}{ DDPG } & 60.47 $\pm$ 2.69 & 23.00 $\pm$ 0.00 & 33.99 $\pm$ 2.43 & 140.40 $\pm$ 4.13 \\
        \multicolumn{ 1 }{|l|}{ TD3 } & 61.19 $\pm$ 1.03 & 23.00 $\pm$ 0.00 & 35.20 $\pm$ 0.97 & 167.60 $\pm$ 8.80 \\
        \multicolumn{ 1 }{|l|}{ SAC } & 69.46 $\pm$ 1.12 & 23.00 $\pm$ 0.00 & 28.72 $\pm$ 1.10 & 137.80 $\pm$ 7.73 \\
        \multicolumn{ 1 }{|l|}{ SAC Auto } & 74.23 $\pm$ 0.86 & 23.00 $\pm$ 0.00 & 23.04 $\pm$ 0.84 & 151.60 $\pm$ 3.01 \\ \hline
        
        &\multicolumn{1}{c}{OL $\% \uparrow$}
        &\multicolumn{1}{c}{OR $\% \downarrow$}
        &\multicolumn{1}{c}{F1 $\% \uparrow$}
        &\multicolumn{1}{c|}{NC $\downarrow$} \\ \hline   
        
        \multicolumn{1}{|l|}{PFT} & 78.54 $\pm$ 0.14 & 40.70 $\pm$ 0.18 & 63.70 $\pm$ 0.09 & 8.90 $\pm$ 0.03 \\ \hline
        \multicolumn{ 1 }{|l|}{ VPG } & 53.11 $\pm$ 9.97 & 32.99 $\pm$ 1.76 & 52.54 $\pm$ 8.71 & 10.12 $\pm$ 5.50 \\
        \multicolumn{ 1 }{|l|}{ A2C } & 16.64 $\pm$ 11.60 & 19.94 $\pm$ 7.41 & 21.02 $\pm$ 13.87 & 42.30 $\pm$ 18.64 \\
        \multicolumn{ 1 }{|l|}{ ACKTR } & 58.84 $\pm$ 0.20 & 30.55 $\pm$ 0.36 & 59.36 $\pm$ 0.14 & 3.75 $\pm$ 0.08 \\
        \multicolumn{ 1 }{|l|}{ TRPO } & 57.88 $\pm$ 0.97 & 30.32 $\pm$ 0.66 & 58.73 $\pm$ 0.44 & 4.73 $\pm$ 0.35 \\
        \multicolumn{ 1 }{|l|}{ PPO } & 50.38 $\pm$ 24.88 & 25.19 $\pm$ 9.69 & 49.59 $\pm$ 24.22 & 18.01 $\pm$ 28.09 \\
        \multicolumn{ 1 }{|l|}{ DDPG } & 54.97 $\pm$ 1.65 & 28.75 $\pm$ 1.46 & 57.35 $\pm$ 0.86 & 5.54 $\pm$ 0.34 \\
        \multicolumn{ 1 }{|l|}{ TD3 } & 53.45 $\pm$ 0.56 & 29.11 $\pm$ 0.64 & 56.36 $\pm$ 0.53 & 3.61 $\pm$ 0.47 \\
        \multicolumn{ 1 }{|l|}{ SAC } & 54.52 $\pm$ 0.51 & 27.92 $\pm$ 0.21 & 57.44 $\pm$ 0.36 & 1.83 $\pm$ 0.03 \\
        \multicolumn{ 1 }{|l|}{ SAC Auto } & 56.81 $\pm$ 0.34 & 27.74 $\pm$ 0.32 & 59.59 $\pm$ 0.30 & 2.73 $\pm$ 0.13 \\ \hline

        \end{tabularx}       
        \end{table}

\section{Hyperparameter search for experiment 2 - Interface seeding}\label{app:exp2_hyperparameters}
      
    Tables \ref{tab:exp2_reinforce_hyperparams}, \ref{tab:exp2_a2c_hyperparams}, \ref{tab:exp2_trpo_hyperparams}, \ref{tab:exp2_acktr_hyperparams}, \ref{tab:exp2_ppo_hyperparams}, \ref{tab:exp2_ddpg_hyperparams}, \ref{tab:exp2_td3_hyperparams}, \ref{tab:exp2_sac_hyperparams}, \ref{tab:exp2_sac_auto_hyperparams}  present the hyperparameters selected for each agent in experiment 2 - interface seeding. 
     
    \begin{table}[!tbhp] 
        \caption{Hyperparameters chosen for REINFORCE in experiment 2}
        \label{tab:exp2_reinforce_hyperparams}
        \renewcommand{\arraystretch}{1.2}
        \begin{tabularx}{\linewidth}{X|*3{>{\arraybackslash}X}}
            Hyperparameter                  & FiberCup & ISMRM2015 \\ \hline 
            Learning rate $\eta$            & 0.0005   & 0.0005   \\ 
            Discount ($\gamma$)                & 0.75     &  0.5     \\ \hline
        \end{tabularx}       

        \caption{Hyperparameters selected for A2C in experiment 2}
        \label{tab:exp2_a2c_hyperparams}
        \renewcommand{\arraystretch}{1.2}
        \begin{tabularx}{\linewidth}{X|*4{>{\arraybackslash}X}}
            Hyperparameter           & FiberCup & ISMRM2015 \\ \hline 
            Learning rate ($\eta$)   & 0.00001 & 0.00001    \\ 
            Discount ($\gamma$)         & 0.5     & 0.75       \\ \hline
        \end{tabularx}       

        \caption{Hyperparameters selected for TRPO in experiment 2}
        \label{tab:exp2_trpo_hyperparams}
        \renewcommand{\arraystretch}{1.2}
        \begin{tabularx}{\linewidth}{X|*4{>{\arraybackslash}X}}
            Hyperparameter            & FiberCup & ISMRM2015 \\ \hline 
            Learning rate ($\eta$)    & 0.001  & 0.0005     \\ 
            Discount ($\gamma$)          & 0.5    & 0.5        \\
            Delta ($\delta$)          & 0.001  & 0.01       \\ \hline
        \end{tabularx}       

        \caption{Hyperparameters selected for ACKTR in experiment 2}
        \label{tab:exp2_acktr_hyperparams}
        \renewcommand{\arraystretch}{1.2}
        \begin{tabularx}{\linewidth}{X|*4{>{\arraybackslash}X}}
            Hyperparameter            & FiberCup & ISMRM2015 \\ \hline 
            Learning rate ($\eta$)    & 0.1  & 0.01    \\ 
            Discount ($\gamma$)          & 0.9  & 0.85    \\
            Delta ($\delta$)          & 0.01 & 0.001   \\ \hline
        \end{tabularx}       

        \caption{Hyperparameters selected for PPO in experiment 2}
        \label{tab:exp2_ppo_hyperparams}
        \renewcommand{\arraystretch}{1.2}
        \begin{tabularx}{\linewidth}{X|*4{>{\arraybackslash}X}}
            Hyperparameter                  & FiberCup          & ISMRM2015   \\ \hline 
            Learning rate ($\eta$)          & 0.00001 & 0.00005 \\ 
            Discount ($\gamma$)                & 0.75    & 0.5 \\
            Clipping parameter ($\epsilon$) & 0.1     & 0.1  \\ \hline
        \end{tabularx}       
    \end{table}
    
    \begin{table}[!tbhp]

        \caption{Hyperparameters selected for DDPG in experiment 2}
        \label{tab:exp2_ddpg_hyperparams}
        \renewcommand{\arraystretch}{1.2}
        \begin{tabularx}{\linewidth}{X|*4{>{\arraybackslash}X}}
            Hyperparameter            & FiberCup  & ISMRM2015 \\ \hline 
            Learning rate ($\eta$)    & 0.0005  & 0.00005 \\ 
            Discount ($\gamma$)          & 0.5     & 0.5     \\
            Action noise ($\sigma$)   & 0.2     & 0.2     \\ \hline
        \end{tabularx}


        \caption{Hyperparameters selected for TD3 in experiment 2}
        \label{tab:exp2_td3_hyperparams}
        \renewcommand{\arraystretch}{1.2}
        \begin{tabularx}{\linewidth}{X|*4{>{\arraybackslash}X}}
            Hyperparameter            & FiberCup & ISMRM2015\\ \hline 
            Learning rate ($\eta$)    & 0.00001 & 0.00005 \\ 
            Discount ($\gamma$)          & 0.5 & 0.5 \\
            Action noise ($\sigma$)   & 0.2 & 0.2 \\ \hline
        \end{tabularx}       

        
        \caption{Hyperparameters selected for SAC in experiment 2}
        \label{tab:exp2_sac_hyperparams}
        \renewcommand{\arraystretch}{1.2}
        \begin{tabularx}{\linewidth}{X|*4{>{\arraybackslash}X}}
            Hyperparameter            & FiberCup & ISMRM2015 \\ \hline 
            Learning rate ($\eta$)    & 0.00005 & 0.0001 \\ 
            Discount ($\gamma$)          & 0.85    & 0.95 \\
            Alpha ($\alpha$)          & 0.1     &  0.2 \\ \hline
        \end{tabularx}      
        

        \caption{Hyperparameters selected for SAC Auto in experiment 2}
        \label{tab:exp2_sac_auto_hyperparams}
        \renewcommand{\arraystretch}{1.2}
        \begin{tabularx}{\linewidth}{X|*4{>{\arraybackslash}X}}
            Hyperparameter            & FiberCup & ISMRM2015 \\ \hline 
            Learning rate ($\eta$)    & 0.00005 & 0.00005 \\ 
            Discount ($\gamma$)          & 0.75   & 0.75 \\ \hline
        \end{tabularx}       
        
        
    \end{table}
    
\section{Full results for experiment 2 - Interface seeding}\label{app:exp2_results}

         Table \ref{tab:agents_fibercup_interface} presents the Tractometer scores for the agents trained and tested on the FiberCup dataset. Table \ref{tab:agents_flipped_interface} presents the Tractometer scores for the agents trained on the FiberCup dataset and tested on the Flipped dataset. Table \ref{tab:agents_ismrm_interface} presents the Tractometer scores for the agents trained and tested on the ISMRM2015 dataset.

            \begin{table}[th] 
            \caption{Tractometer scores for agents trained and tested on the FiberCup dataset using interface seeding. \emph{Italics} indicate $p < 0.005$ that the mean is lower than the reference results from experiment 1 using a one-sided Welch t-test. \textbf{Bold} indicates $p < 0.005$ that the mean is higher than the reference experiment using the same test.}
            \label{tab:agents_fibercup_interface}
            \renewcommand{\arraystretch}{1.0}
            \begin{tabularx}{\columnwidth}{>{\hsize=.1\hsize}X|*4{>{\centering\arraybackslash}X}|}
            \cline{2-5}
            
            & \multicolumn{1}{c }{VC $\% \uparrow$}
            & \multicolumn{1}{c}{VB $\uparrow$} 
            & \multicolumn{1}{c }{IC $\% \downarrow$}
            & \multicolumn{1}{c|}{IB $\downarrow$} \\ \hline
            
            \multicolumn{ 1 }{|l|}{ PFT } & \textbf{29.44} $\pm$ 0.53& 7.00 $\pm$ 0.00& \emph{11.32} $\pm$ 0.25& 1.20 $\pm$ 0.40 \\ \hline 
            \multicolumn{ 1 }{|l|}{ VPG } & 35.31 $\pm$ 11.93& 6.20 $\pm$ 0.98& 24.34 $\pm$ 8.89& 1.60 $\pm$ 0.49 \\
            \multicolumn{ 1 }{|l|}{ A2C } & 38.86 $\pm$ 12.92& \emph{3.20} $\pm$ 0.75& 8.82 $\pm$ 11.17& 1.20 $\pm$ 0.40 \\
            \multicolumn{ 1 }{|l|}{ ACKTR } & 24.99 $\pm$ 23.81& 2.80 $\pm$ 1.72& 20.29 $\pm$ 26.88& 1.20 $\pm$ 0.75 \\
            \multicolumn{ 1 }{|l|}{ TRPO } & \textbf{70.71} $\pm$ 2.58& 6.80 $\pm$ 0.40& 12.98 $\pm$ 3.24& 1.40 $\pm$ 0.49 \\
            \multicolumn{ 1 }{|l|}{ PPO } & 39.43 $\pm$ 16.20& 5.40 $\pm$ 1.20& 28.25 $\pm$ 12.18& 1.60 $\pm$ 0.49 \\
            \multicolumn{ 1 }{|l|}{ DDPG } & \textbf{54.92} $\pm$ 10.44& 6.80 $\pm$ 0.40& 28.08 $\pm$ 8.72& 2.00 $\pm$ 0.00 \\
            \multicolumn{ 1 }{|l|}{ TD3 } & \textbf{73.32} $\pm$ 5.66& 7.00 $\pm$ 0.00& 13.40 $\pm$ 2.68& 2.20 $\pm$ 0.75 \\
            \multicolumn{ 1 }{|l|}{ SAC } & \textbf{84.61} $\pm$ 5.86& 7.00 $\pm$ 0.00& 14.92 $\pm$ 6.08& 1.80 $\pm$ 0.40 \\
            \multicolumn{ 1 }{|l|}{ SAC Auto } & 80.72 $\pm$ 3.93& 7.00 $\pm$ 0.00& 17.41 $\pm$ 3.65& 1.80 $\pm$ 0.40 \\
            \hline
            
            &\multicolumn{1}{c}{OL $\% \uparrow$}
            &\multicolumn{1}{c}{OR $\% \downarrow$}
            &\multicolumn{1}{c}{F1 $\% \uparrow$}
            &\multicolumn{1}{c|}{NC $\downarrow$} \\ \hline   
            
            \multicolumn{ 1 }{|l|}{ PFT } & \textbf{98.74} $\pm$ 0.26& \textbf{25.36} $\pm$ 0.77& 84.81 $\pm$ 0.43& \textbf{59.24} $\pm$ 0.45 \\ \hline
            \multicolumn{ 1 }{|l|}{ VPG } & 58.83 $\pm$ 12.43& 10.03 $\pm$ 2.14& 64.49 $\pm$ 11.98& 40.35 $\pm$ 9.14 \\
            \multicolumn{ 1 }{|l|}{ A2C } & \emph{27.31} $\pm$ 7.11& \emph{6.00} $\pm$ 2.94& \emph{30.87} $\pm$ 7.14& 52.33 $\pm$ 10.17 \\
            \multicolumn{ 1 }{|l|}{ ACKTR } & \emph{22.31} $\pm$ 13.29& \emph{3.27} $\pm$ 2.94& \emph{27.20} $\pm$ 16.12& 54.73 $\pm$ 35.09 \\
            \multicolumn{ 1 }{|l|}{ TRPO } & 75.91 $\pm$ 4.21& 14.36 $\pm$ 0.65& 78.13 $\pm$ 4.66& 16.31 $\pm$ 1.02 \\
            \multicolumn{ 1 }{|l|}{ PPO } & \emph{41.89} $\pm$ 13.13& \emph{9.24} $\pm$ 2.76& 49.70 $\pm$ 13.71& 32.32 $\pm$ 16.42 \\
            \multicolumn{ 1 }{|l|}{ DDPG } & 76.81 $\pm$ 5.16& 11.91 $\pm$ 1.74& \textbf{80.14} $\pm$ 4.98& \emph{17.00} $\pm$ 6.19 \\
            \multicolumn{ 1 }{|l|}{ TD3 } & \textbf{77.73} $\pm$ 1.41& 11.99 $\pm$ 1.25& \textbf{81.94} $\pm$ 0.89& \emph{13.28} $\pm$ 4.14 \\
            \multicolumn{ 1 }{|l|}{ SAC } & \emph{67.20} $\pm$ 0.95& \emph{7.09} $\pm$ 1.11& \emph{77.35} $\pm$ 0.65& \emph{0.47} $\pm$ 0.27 \\
            \multicolumn{ 1 }{|l|}{ SAC Auto } & \emph{77.18} $\pm$ 0.82& \emph{7.28} $\pm$ 0.99& 84.06 $\pm$ 0.58& \emph{1.87} $\pm$ 0.38 \\
            \hline

            \end{tabularx}       

            \caption{Tractometer scores for agents trained on the FiberCup dataset and tested on the Flipped dataset using interface seeding. \emph{Italics} indicate $p < 0.005$ that the mean is lower than the reference results from experiment 1 using a one-sided Welch t-test. \textbf{Bold} indicates $p < 0.005$ that the mean is higher than the reference experiment using the same test.}
            \label{tab:agents_flipped_interface}
            \renewcommand{\arraystretch}{1.0}
            \begin{tabularx}{\columnwidth}{>{\hsize=.1\hsize}X|*4{>{\centering\arraybackslash}X}|}
            \cline{2-5}
            
            & \multicolumn{1}{c }{VC $\% \uparrow$}
            & \multicolumn{1}{c}{VB $\uparrow$} 
            & \multicolumn{1}{c }{IC $\% \downarrow$}
            & \multicolumn{1}{c|}{IB $\downarrow$} \\ \hline
            \multicolumn{ 1 }{|l|}{ PFT } & \textbf{29.90} $\pm$ 0.34& 7.00 $\pm$ 0.00& \emph{12.15} $\pm$ 0.17& 1.60 $\pm$ 0.49 \\ \hline
            \multicolumn{ 1 }{|l|}{ VPG } & 29.42 $\pm$ 10.65& 6.00 $\pm$ 1.26& 29.77 $\pm$ 11.96& 1.80 $\pm$ 0.40 \\
            \multicolumn{ 1 }{|l|}{ A2C } & 32.93 $\pm$ 12.41& 3.20 $\pm$ 1.60& 12.02 $\pm$ 9.40& 1.20 $\pm$ 0.40 \\
            \multicolumn{ 1 }{|l|}{ ACKTR } & 31.11 $\pm$ 19.49& 3.20 $\pm$ 1.94& 17.51 $\pm$ 10.14& 1.20 $\pm$ 0.40 \\
            \multicolumn{ 1 }{|l|}{ TRPO } & \textbf{63.04} $\pm$ 3.02& 6.80 $\pm$ 0.40& 18.61 $\pm$ 3.01& 1.60 $\pm$ 0.80 \\
            \multicolumn{ 1 }{|l|}{ PPO } & \emph{41.96} $\pm$ 3.98& 4.80 $\pm$ 1.17& 27.14 $\pm$ 13.34& 1.60 $\pm$ 0.49 \\
            \multicolumn{ 1 }{|l|}{ DDPG } & \textbf{60.69} $\pm$ 11.51& 7.00 $\pm$ 0.00& 16.71 $\pm$ 3.27& 1.80 $\pm$ 0.40 \\
            \multicolumn{ 1 }{|l|}{ TD3 } & \textbf{66.48} $\pm$ 5.09& \textbf{7.00} $\pm$ 0.00& 18.21 $\pm$ 2.72& 1.20 $\pm$ 0.40 \\
            \multicolumn{ 1 }{|l|}{ SAC } & \textbf{83.58} $\pm$ 5.45& 7.00 $\pm$ 0.00& 14.55 $\pm$ 5.71& 1.40 $\pm$ 0.49 \\
            \multicolumn{ 1 }{|l|}{ SAC Auto } & \textbf{87.83} $\pm$ 5.45& 7.00 $\pm$ 0.00& 6.89 $\pm$ 4.55& 2.20 $\pm$ 0.40 \\
            \hline
            
            &\multicolumn{1}{c}{OL $\% \uparrow$}
            &\multicolumn{1}{c}{OR $\% \downarrow$}
            &\multicolumn{1}{c}{F1 $\% \uparrow$}
            &\multicolumn{1}{c|}{NC $\downarrow$} \\ \hline   
            
            \multicolumn{ 1 }{|l|}{ PFT } & \textbf{98.71} $\pm$ 0.23& \textbf{24.03} $\pm$ 0.50& 85.74 $\pm$ 0.35& \textbf{57.95} $\pm$ 0.35 \\ \hline
            \multicolumn{ 1 }{|l|}{ VPG } & 56.64 $\pm$ 17.77& 9.95 $\pm$ 3.04& 62.57 $\pm$ 17.46& 40.81 $\pm$ 14.03 \\
            \multicolumn{ 1 }{|l|}{ A2C } & 28.91 $\pm$ 11.34& \emph{6.43} $\pm$ 3.44& 32.31 $\pm$ 13.68& 55.05 $\pm$ 20.40 \\
            \multicolumn{ 1 }{|l|}{ ACKTR } & \emph{21.59} $\pm$ 11.74& \emph{4.81} $\pm$ 3.97& \emph{27.36} $\pm$ 15.21& 51.38 $\pm$ 28.26 \\
            \multicolumn{ 1 }{|l|}{ TRPO } & 76.44 $\pm$ 5.42& 13.89 $\pm$ 1.59& 78.17 $\pm$ 4.86& \emph{18.35} $\pm$ 5.06 \\
            \multicolumn{ 1 }{|l|}{ PPO } & \emph{37.73} $\pm$ 8.28& \emph{5.42} $\pm$ 2.48& \emph{46.03} $\pm$ 9.59& 30.90 $\pm$ 15.87 \\
            \multicolumn{ 1 }{|l|}{ DDPG } & 79.74 $\pm$ 4.63& 12.16 $\pm$ 1.87& 82.30 $\pm$ 4.10& \emph{22.60} $\pm$ 11.34 \\
            \multicolumn{ 1 }{|l|}{ TD3 } & \textbf{73.92} $\pm$ 5.29& 11.56 $\pm$ 0.94& \textbf{78.57} $\pm$ 4.73& \emph{15.31} $\pm$ 6.70 \\
            \multicolumn{ 1 }{|l|}{ SAC } & \emph{66.43} $\pm$ 2.01& \emph{10.23} $\pm$ 0.74& \emph{74.05} $\pm$ 2.25& \emph{1.87} $\pm$ 0.42 \\
            \multicolumn{ 1 }{|l|}{ SAC Auto } & \emph{79.07} $\pm$ 0.98& \emph{10.26} $\pm$ 1.02& 83.52 $\pm$ 0.65& \emph{5.28} $\pm$ 1.72 \\
            \hline
            
            \end{tabularx}       
            \end{table}           
              
            \begin{table}[!tbhp] 
            \caption{Tractometer scores for agents trained and tested on the ISMRM2015 dataset using interface seeding. \emph{Italics} indicate $p < 0.005$ that the mean is lower than the reference results from experiment 1 using a one-sided Welch t-test. \textbf{Bold} indicates $p < 0.005$ that the mean is higher than the reference experiment using the same test.}
            \label{tab:agents_ismrm_interface}
            \renewcommand{\arraystretch}{1.2}
            \begin{tabularx}{\columnwidth}{>{\hsize=.1\hsize}X|*4{>{\centering\arraybackslash}X}|}
            \cline{2-5}
            
            & \multicolumn{1}{c }{VC $\% \uparrow$}
            & \multicolumn{1}{c}{VB $\uparrow$} 
            & \multicolumn{1}{c }{IC $\% \downarrow$}
            & \multicolumn{1}{c|}{IB $\downarrow$} \\ \hline
            
            \multicolumn{ 1 }{|l|}{ PFT } & \emph{54.29} $\pm$ 0.09& 24.20 $\pm$ 0.40& \emph{28.79} $\pm$ 0.08& \emph{108.60} $\pm$ 2.15 \\ \hline
            \multicolumn{ 1 }{|l|}{ VPG } & 42.55 $\pm$ 7.89& 22.60 $\pm$ 0.80& 41.28 $\pm$ 5.14& 138.40 $\pm$ 13.66 \\
            \multicolumn{ 1 }{|l|}{ A2C } & 26.59 $\pm$ 9.61& 20.40 $\pm$ 1.50& 52.07 $\pm$ 6.39& 67.20 $\pm$ 6.18 \\
            \multicolumn{ 1 }{|l|}{ ACKTR } & \emph{35.85} $\pm$ 5.06& 21.40 $\pm$ 0.80& \textbf{50.86} $\pm$ 4.49& \emph{88.80} $\pm$ 10.93 \\
            \multicolumn{ 1 }{|l|}{ TRPO } & \textbf{64.35} $\pm$ 3.92& \textbf{23.20} $\pm$ 0.40& 24.53 $\pm$ 4.06& \textbf{115.60} $\pm$ 9.85 \\
            \multicolumn{ 1 }{|l|}{ PPO } & 67.01 $\pm$ 2.21& 23.00 $\pm$ 0.00& 23.21 $\pm$ 2.31& 150.40 $\pm$ 5.92 \\
            \multicolumn{ 1 }{|l|}{ DDPG } & 54.29 $\pm$ 1.25& 23.00 $\pm$ 0.00& 36.74 $\pm$ 1.41& \textbf{203.80} $\pm$ 10.85 \\
            \multicolumn{ 1 }{|l|}{ TD3 } & 59.07 $\pm$ 2.50& 23.00 $\pm$ 0.00& \emph{32.88} $\pm$ 1.65& 191.80 $\pm$ 12.16 \\
            \multicolumn{ 1 }{|l|}{ SAC } & 66.38 $\pm$ 3.56& 23.00 $\pm$ 0.00& 30.73 $\pm$ 3.54& \emph{127.60} $\pm$ 10.09 \\
            \multicolumn{ 1 }{|l|}{ SAC Auto } & \emph{65.97} $\pm$ 0.82& 23.00 $\pm$ 0.00& \textbf{27.10} $\pm$ 0.83& 161.60 $\pm$ 7.55 \\
            \hline
            
            &\multicolumn{1}{c}{OL $\% \uparrow$}
            &\multicolumn{1}{c}{OR $\% \downarrow$}
            &\multicolumn{1}{c}{F1 $\% \uparrow$}
            &\multicolumn{1}{c|}{NC $\downarrow$} \\ \hline   
            
            \multicolumn{ 1 }{|l|}{ PFT } & \textbf{72.78} $\pm$ 0.22& \textbf{34.86} $\pm$ 0.47& \textbf{64.89} $\pm$ 0.35& \textbf{16.92} $\pm$ 0.14 \\ \hline
            \multicolumn{ 1 }{|l|}{ VPG } & 40.59 $\pm$ 6.37& \emph{28.44} $\pm$ 1.26& 45.71 $\pm$ 4.93& 16.17 $\pm$ 4.50 \\
            \multicolumn{ 1 }{|l|}{ A2C } & 19.67 $\pm$ 3.15& 25.22 $\pm$ 1.16& 25.95 $\pm$ 3.82& 21.34 $\pm$ 6.39 \\
            \multicolumn{ 1 }{|l|}{ ACKTR } & \emph{27.07} $\pm$ 1.64& \emph{25.11} $\pm$ 1.90& \emph{34.55} $\pm$ 2.01& \textbf{13.30} $\pm$ 2.15 \\
            \multicolumn{ 1 }{|l|}{ TRPO } & \textbf{41.55} $\pm$ 3.12& \textbf{25.96} $\pm$ 1.43& \textbf{49.06} $\pm$ 3.37& \emph{11.12} $\pm$ 1.75 \\
            \multicolumn{ 1 }{|l|}{ PPO } & 44.94 $\pm$ 1.21& 25.62 $\pm$ 0.25& 52.48 $\pm$ 0.86& 9.79 $\pm$ 0.87 \\
            \multicolumn{ 1 }{|l|}{ DDPG } & \emph{51.94} $\pm$ 1.46& \emph{29.09} $\pm$ 0.93& \emph{55.42} $\pm$ 1.19& 8.97 $\pm$ 2.17 \\
            \multicolumn{ 1 }{|l|}{ TD3 } & \emph{51.26} $\pm$ 1.39& \emph{29.32} $\pm$ 0.68& \emph{55.24} $\pm$ 1.08& \textbf{8.05} $\pm$ 1.07 \\
            \multicolumn{ 1 }{|l|}{ SAC } & \emph{39.70} $\pm$ 1.07& \emph{26.04} $\pm$ 1.13& \emph{46.96} $\pm$ 1.23& \textbf{2.89} $\pm$ 0.09 \\
            \multicolumn{ 1 }{|l|}{ SAC Auto } & \emph{49.25} $\pm$ 0.35& \emph{27.35} $\pm$ 0.38& \emph{54.99} $\pm$ 0.34& \textbf{6.93} $\pm$ 0.06 \\
            \hline

            \end{tabularx}       
            \end{table}              
            
\section{Hyperparameter search for experiment 3 - WM seeding with no re-tracking}\label{app:exp3_hyperparameters}

    Tables \ref{tab:exp3_reinforce_hyperparams}, \ref{tab:exp3_a2c_hyperparams}, \ref{tab:exp3_trpo_hyperparams}, \ref{tab:exp3_acktr_hyperparams}, \ref{tab:exp3_ppo_hyperparams}, \ref{tab:exp3_ddpg_hyperparams}, \ref{tab:exp3_td3_hyperparams}, \ref{tab:exp3_sac_hyperparams}, \ref{tab:exp3_sac_auto_hyperparams}  present the hyperparameters selected for each agent in experiment 3 - interface seeding. 
     
    \begin{table}[!tbhp] 
        \caption{Hyperparameters chosen for REINFORCE in experiment 3}
        \label{tab:exp3_reinforce_hyperparams}
        \renewcommand{\arraystretch}{1.2}
        \begin{tabularx}{\linewidth}{X|*3{>{\arraybackslash}X}}
            Hyperparameter                  & FiberCup & ISMRM2015 \\ \hline 
            Learning rate $\eta$            & 0.001   &  0.0005  \\ 
            Discount ($\gamma$)                & 0.9     &  0.85     \\ \hline
        \end{tabularx}       

        \caption{Hyperparameters selected for A2C in experiment 3}
        \label{tab:exp3_a2c_hyperparams}
        \renewcommand{\arraystretch}{1.2}
        \begin{tabularx}{\linewidth}{X|*4{>{\arraybackslash}X}}
            Hyperparameter           & FiberCup & ISMRM2015 \\ \hline 
            Learning rate $\eta$     & 0.00001   & 0.00001   \\ 
            Discount ($\gamma$)         & 0.75     &  0.95     \\ \hline
        \end{tabularx}       

        \caption{Hyperparameters selected for TRPO in experiment 3}
        \label{tab:exp3_trpo_hyperparams}
        \renewcommand{\arraystretch}{1.2}
        \begin{tabularx}{\linewidth}{X|*4{>{\arraybackslash}X}}
            Hyperparameter            & FiberCup & ISMRM2015 \\ \hline 
            Learning rate ($\eta$)    & 0.0005  & 0.0001    \\ 
            Discount ($\gamma$)          & 0.75    & 0.5        \\
            Delta ($\delta$)          & 0.001   & 0.01       \\ \hline
        \end{tabularx}       

        \caption{Hyperparameters selected for ACKTR in experiment 3}
        \label{tab:exp3_acktr_hyperparams}
        \renewcommand{\arraystretch}{1.2}
        \begin{tabularx}{\linewidth}{X|*4{>{\arraybackslash}X}}
            Hyperparameter            & FiberCup & ISMRM2015 \\ \hline 
            Learning rate ($\eta$)    & 0.01  & 0.01    \\ 
            Discount ($\gamma$)          & 0.75  & 0.5    \\
            Delta ($\delta$)          & 0.01 & 0.005   \\ \hline
        \end{tabularx}       

        \caption{Hyperparameters selected for PPO in experiment 3}
        \label{tab:exp3_ppo_hyperparams}
        \renewcommand{\arraystretch}{1.2}
        \begin{tabularx}{\linewidth}{X|*4{>{\arraybackslash}X}}
            Hyperparameter                  & FiberCup          & ISMRM2015   \\ \hline 
            Learning rate ($\eta$)          & 0.00005 & 0.00005 \\ 
            Discount ($\gamma$)                & 0.85    & 0.5     \\
            Clipping parameter ($\epsilon$) & 0.05    & 0.05    \\ \hline
        \end{tabularx}       
    \end{table}
    
    \begin{table}[!tbhp]

        \caption{Hyperparameters selected for DDPG in experiment 3}
        \label{tab:exp3_ddpg_hyperparams}
        \renewcommand{\arraystretch}{1.2}
        \begin{tabularx}{\linewidth}{X|*4{>{\arraybackslash}X}}
            Hyperparameter            & FiberCup  & ISMRM2015 \\ \hline 
            Learning rate ($\eta$)    & 0.0001  & 0.00001 \\ 
            Discount ($\gamma$)          & 0.5     & 0.5     \\
            Action noise ($\sigma$)   & 0.2     & 0.2     \\ \hline
        \end{tabularx}


        \caption{Hyperparameters selected for TD3 in experiment 3}
        \label{tab:exp3_td3_hyperparams}
        \renewcommand{\arraystretch}{1.2}
        \begin{tabularx}{\linewidth}{X|*4{>{\arraybackslash}X}}
            Hyperparameter            & FiberCup & ISMRM2015\\ \hline 
            Learning rate ($\eta$)    & 0.0005 & 0.0005 \\ 
            Discount ($\gamma$)          & 0.75 & 0.5 \\
            Action noise ($\sigma$)   & 0.2 & 0.25 \\ \hline
        \end{tabularx}       

        
        \caption{Hyperparameters selected for SAC in experiment 3}
        \label{tab:exp3_sac_hyperparams}
        \renewcommand{\arraystretch}{1.2}
        \begin{tabularx}{\linewidth}{X|*4{>{\arraybackslash}X}}
            Hyperparameter            & FiberCup & ISMRM2015 \\ \hline 
            Learning rate ($\eta$)    & 0.00005 & 0.0005 \\ 
            Discount ($\gamma$)          & 0.85    & 0.75 \\
            Alpha ($\alpha$)          & 0.075   & 0.1 \\ \hline
        \end{tabularx}      
        

        \caption{Hyperparameters selected for SAC Auto in experiment 3}
        \label{tab:exp3_sac_auto_hyperparams}
        \renewcommand{\arraystretch}{1.2}
        \begin{tabularx}{\linewidth}{X|*4{>{\arraybackslash}X}}
            Hyperparameter            & FiberCup & ISMRM2015 \\ \hline 
            Learning rate ($\eta$)    & 0.001 & 0.001 \\ 
            Discount ($\gamma$)          & 0.75  & 0.5 \\ \hline
        \end{tabularx}       
        
        
    \end{table}
\section{Full results for experiment 3 - Retracking}\label{app:exp3_full_results}

     Complete results for experiment 3 - retracking. Table \ref{tab:diff_fibercup_noretrack} presents the Tractometer scores for the agents trained and tested on the FiberCup dataset. Table \ref{tab:diff_flipped_noretrack} presents the Tractometer scores for the agents trained on the FiberCup dataset and tested on the Flipped dataset. Table \ref{tab:diff_ismrm2015_noretrack} presents the Tractometer scores for the agents trained and tested on the ISMRM2015 dataset.   
    \begin{table}[!thbp] 
    \caption{Difference between Tractometer scores for experiments 3 and 1 for agents trained and tested on the FiberCup dataset. \textbf{Bold} indicates a statistically different mean using a Welch t-test with p < 0.005.}
    \label{tab:diff_fibercup_noretrack}
    \renewcommand{\arraystretch}{1.2}
    \begin{tabularx}{\columnwidth}{>{\hsize=.1\hsize}X|*5{>{\centering\arraybackslash}X}|}
    \cline{2-6}
    
    & \multicolumn{1}{c }{VC $\% \uparrow$}
    & \multicolumn{1}{c }{IC $\% \downarrow$}
    & \multicolumn{1}{c }{NC $\downarrow$}
    & \multicolumn{1}{c }{OL $\uparrow$}
    & \multicolumn{1}{c|}{OR $\downarrow$} \\ \hline
    
    \multicolumn{ 1 }{|l|}{ VPG } & -17.28 $\pm$ 1.55& -3.79 $\pm$ 7.06& 21.07 $\pm$ 8.51& -21.43 $\pm$ 12.39& -0.95 $\pm$ 5.34\\
    \multicolumn{ 1 }{|l|}{ A2C } & \textbf{-17.19} $\pm$ 1.44& -0.89 $\pm$ 2.49& 18.09 $\pm$ 1.47& \textbf{-32.65} $\pm$ 5.64& \textbf{-10.77} $\pm$ 2.16\\
    \multicolumn{ 1 }{|l|}{ ACKTR } & \textbf{-54.45} $\pm$ 1.12& 14.04 $\pm$ 13.10& \textbf{40.41} $\pm$ 13.29& \textbf{-37.62} $\pm$ 6.08& \textbf{-8.55} $\pm$ 3.04\\
    \multicolumn{ 1 }{|l|}{ TRPO } & -15.98 $\pm$ 1.09& -5.44 $\pm$ 9.55& 21.42 $\pm$ 10.60& 2.50 $\pm$ 7.28& 0.42 $\pm$ 2.72\\
    \multicolumn{ 1 }{|l|}{ PPO } & \textbf{-57.23} $\pm$ 2.01& -11.50 $\pm$ 11.33& \textbf{68.73} $\pm$ 13.30& \textbf{-57.55} $\pm$ 2.10& -6.40 $\pm$ 3.15\\
    \multicolumn{ 1 }{|l|}{ DDPG } & \textbf{-31.09} $\pm$ 2.15& 2.02 $\pm$ 6.76& \textbf{29.07} $\pm$ 8.36& 1.06 $\pm$ 4.56& -2.03 $\pm$ 3.18\\
    \multicolumn{ 1 }{|l|}{ TD3 } & -20.15 $\pm$ 0.99& \textbf{26.71} $\pm$ 6.33& -6.56 $\pm$ 6.17& \textbf{-15.75} $\pm$ 5.72& -0.76 $\pm$ 3.17\\
    \multicolumn{ 1 }{|l|}{ SAC } & \textbf{-50.78} $\pm$ 2.31& 6.72 $\pm$ 4.77& \textbf{44.06} $\pm$ 3.23& \textbf{-36.23} $\pm$ 7.03& -3.38 $\pm$ 1.91\\
    \multicolumn{ 1 }{|l|}{ SAC Auto } & \textbf{-60.90} $\pm$ 2.36& \textbf{16.24} $\pm$ 4.62& \textbf{44.66} $\pm$ 2.65& \textbf{-27.97} $\pm$ 2.59& -4.17 $\pm$ 2.72\\ \hline
    \end{tabularx}       

    \caption{Difference between tractometer scores for experiment 1 and 3 for agents trained on the FiberCup dataset and tested on the Flipped dataset. \textbf{Bold} indicates a statistically different mean using a Welch t-test with p < 0.005.}
    \label{tab:diff_flipped_noretrack}
    \renewcommand{\arraystretch}{1.2}
    \begin{tabularx}{\columnwidth}{>{\hsize=.1\hsize}X|*5{>{\centering\arraybackslash}X}|}
    \cline{2-6}
    
    & \multicolumn{1}{c }{VC $\% \uparrow$}
    & \multicolumn{1}{c }{IC $\% \downarrow$}
    & \multicolumn{1}{c }{NC $\downarrow$}
    & \multicolumn{1}{c }{OL $\uparrow$}
    & \multicolumn{1}{c|}{OR $\downarrow$} \\ \hline

    \multicolumn{ 1 }{|l|}{ VPG } & -16.10 $\pm$ 1.67& -11.18 $\pm$ 3.14& 27.28 $\pm$ 4.12& -28.88 $\pm$ 11.50& -5.16 $\pm$ 3.23\\
    \multicolumn{ 1 }{|l|}{ A2C } & -12.90 $\pm$ 1.57& -5.07 $\pm$ 3.97& \textbf{17.97} $\pm$ 4.93& -30.05 $\pm$ 6.32& \textbf{-12.24} $\pm$ 2.62\\
    \multicolumn{ 1 }{|l|}{ ACKTR } & \textbf{-51.35} $\pm$ 1.05& -4.82 $\pm$ 5.08& \textbf{56.17} $\pm$ 4.69& \textbf{-42.77} $\pm$ 4.81& \textbf{-6.86} $\pm$ 0.77\\
    \multicolumn{ 1 }{|l|}{ TRPO } & -24.62 $\pm$ 1.27& -2.33 $\pm$ 3.34& 26.95 $\pm$ 3.12& -1.27 $\pm$ 7.56& -1.09 $\pm$ 1.06\\
    \multicolumn{ 1 }{|l|}{ PPO } & \textbf{-55.16} $\pm$ 2.43& -10.96 $\pm$ 7.97& \textbf{66.12} $\pm$ 10.33& \textbf{-63.66} $\pm$ 9.03& -6.76 $\pm$ 4.90\\
    \multicolumn{ 1 }{|l|}{ DDPG } & \textbf{-17.78} $\pm$ 0.97& -8.40 $\pm$ 2.16& \textbf{26.17} $\pm$ 3.05& 2.88 $\pm$ 4.95& -1.12 $\pm$ 2.22\\
    \multicolumn{ 1 }{|l|}{ TD3 } & -19.71 $\pm$ 1.69& 4.40 $\pm$ 7.92& 15.32 $\pm$ 7.92& \textbf{-8.93} $\pm$ 3.04& 3.73 $\pm$ 1.22\\
    \multicolumn{ 1 }{|l|}{ SAC } & \textbf{-57.50} $\pm$ 3.51& \textbf{-14.59} $\pm$ 3.74& \textbf{72.09} $\pm$ 6.92& \textbf{-28.07} $\pm$ 0.75& \textbf{-5.50} $\pm$ 0.62\\
    \multicolumn{ 1 }{|l|}{ SAC Auto } & \textbf{-69.44} $\pm$ 3.85& 2.09 $\pm$ 3.67& \textbf{67.35} $\pm$ 7.22& \textbf{-28.86} $\pm$ 3.88& \textbf{-7.86} $\pm$ 1.06\\ \hline          

    \end{tabularx}       
      
    \caption{Difference between tractometer scores for experiment 1 and 3 for agents trained and tested on the ISMRM2015 dataset. \textbf{Bold} indicates a statistically different mean using a Welch t-test with p < 0.005.}
    \label{tab:diff_ismrm2015_noretrack}
    \renewcommand{\arraystretch}{1.2}
    \begin{tabularx}{\columnwidth}{>{\hsize=.1\hsize}X|*5{>{\centering\arraybackslash}X}|}
    \cline{2-6}
    
    & \multicolumn{1}{c }{VC $\% \uparrow$}
    & \multicolumn{1}{c }{IC $\% \downarrow$}
    & \multicolumn{1}{c }{NC $\downarrow$}
    & \multicolumn{1}{c }{OL $\uparrow$}
    & \multicolumn{1}{c|}{OR $\downarrow$} \\ \hline
    
    \multicolumn{ 1 }{|l|}{ VPG } & -23.38 $\pm$ 1.34& 13.13 $\pm$ 3.10& 10.25 $\pm$ 2.96& -24.57 $\pm$ 3.11& -5.29 $\pm$ 3.12\\
    \multicolumn{ 1 }{|l|}{ A2C } & 11.02 $\pm$ 0.89& 1.67 $\pm$ 1.04& -12.69 $\pm$ 0.49& 4.26 $\pm$ 0.73& 0.64 $\pm$ 1.00\\
    \multicolumn{ 1 }{|l|}{ ACKTR } & \textbf{-37.06} $\pm$ 5.77& \textbf{20.34} $\pm$ 3.54& \textbf{16.71} $\pm$ 3.28& \textbf{-19.87} $\pm$ 1.31& -3.11 $\pm$ 2.26\\
    \multicolumn{ 1 }{|l|}{ TRPO } & \textbf{-24.38} $\pm$ 3.37& \textbf{10.52} $\pm$ 2.01& \textbf{13.86} $\pm$ 3.53& \textbf{-13.39} $\pm$ 1.82& \textbf{-4.10} $\pm$ 0.95\\
    \multicolumn{ 1 }{|l|}{ PPO } & -47.75 $\pm$ 2.31& \textbf{21.93} $\pm$ 3.49& 25.82 $\pm$ 3.60& -36.92 $\pm$ 4.17& -5.77 $\pm$ 2.34\\
    \multicolumn{ 1 }{|l|}{ DDPG } & \textbf{-23.40} $\pm$ 3.99& \textbf{13.17} $\pm$ 1.17& \textbf{10.23} $\pm$ 3.38& \textbf{-14.49} $\pm$ 2.19& \textbf{-4.44} $\pm$ 0.65\\
    \multicolumn{ 1 }{|l|}{ TD3 } & \textbf{-27.61} $\pm$ 2.30& \textbf{14.73} $\pm$ 1.38& \textbf{12.88} $\pm$ 2.07& \textbf{-14.38} $\pm$ 1.63& -3.22 $\pm$ 1.37\\
    \multicolumn{ 1 }{|l|}{ SAC } & \textbf{-29.47} $\pm$ 3.66& \textbf{19.19} $\pm$ 1.99& \textbf{10.28} $\pm$ 1.87& \textbf{-18.25} $\pm$ 1.84& -1.91 $\pm$ 0.84\\
    \multicolumn{ 1 }{|l|}{ SAC Auto } & \textbf{-31.23} $\pm$ 4.12& \textbf{18.15} $\pm$ 1.45& \textbf{13.07} $\pm$ 3.01& \textbf{-14.84} $\pm$ 1.10& \textbf{-2.47} $\pm$ 0.78\\ \hline

    \end{tabularx}       
    \end{table}      
\section{Full results for experiment 3 - No retracking}\label{app:exp3_results}

         Table \ref{tab:agents_fibercup_noretrack} presents the Tractometer scores for the agents trained and tested on the FiberCup dataset. Table \ref{tab:agents_flipped_noretrack} presents the Tractometer scores for agents trained on the FiberCup dataset and tested on the Flipped dataset without retracking. Table \ref{tab:agents_ismrm2015_noretrack} presents the Tractometer scores for the agents trained and tested on the ISMRM2015 dataset without retracking.

        \begin{table}[!thbp] 
        \caption{Tractometer scores from experiment 3 for agents trained on the FiberCup dataset.  \emph{Italics} indicate $p < 0.005$ that the mean is lower than the reference results from experiment 1 using a one-sided Welch t-test. \textbf{Bold} indicates $p < 0.005$ that the mean is higher than the reference experiment using the same test.}
        \label{tab:agents_fibercup_noretrack}
        \renewcommand{\arraystretch}{1.2}
        \begin{tabularx}{\columnwidth}{>{\hsize=.1\hsize}X|*4{>{\centering\arraybackslash}X}|}
        \cline{2-5}
        
        & \multicolumn{1}{c }{VC $\% \uparrow$}
        & \multicolumn{1}{c}{VB $\uparrow$} 
        & \multicolumn{1}{c }{IC $\% \downarrow$}
        & \multicolumn{1}{c|}{IB $\downarrow$} \\ \hline
        
        \multicolumn{ 1 }{|l|}{ VPG } & 4.03 $\pm$ 1.55& 4.00 $\pm$ 1.67& 11.64 $\pm$ 7.06& 1.60 $\pm$ 0.80 \\
        \multicolumn{ 1 }{|l|}{ A2C } & \emph{6.32} $\pm$ 1.44& \emph{3.40} $\pm$ 0.80& 14.75 $\pm$ 2.49& 1.40 $\pm$ 0.49 \\
        \multicolumn{ 1 }{|l|}{ ACKTR } & \emph{8.87} $\pm$ 1.12& \emph{5.60} $\pm$ 0.49& 31.68 $\pm$ 13.10& 1.40 $\pm$ 0.49 \\
        \multicolumn{ 1 }{|l|}{ TRPO } & 11.05 $\pm$ 1.09& 5.40 $\pm$ 0.80& 30.25 $\pm$ 9.55& 1.80 $\pm$ 0.75 \\
        \multicolumn{ 1 }{|l|}{ PPO } & \emph{2.76} $\pm$ 2.01& \emph{4.60} $\pm$ 0.49& 7.15 $\pm$ 11.33& 1.60 $\pm$ 0.49 \\
        \multicolumn{ 1 }{|l|}{ DDPG } & \emph{10.08} $\pm$ 2.15& 7.00 $\pm$ 0.00& 26.83 $\pm$ 6.76& 2.20 $\pm$ 0.75 \\
        \multicolumn{ 1 }{|l|}{ TD3 } & 7.83 $\pm$ 0.99& 5.40 $\pm$ 1.02& \textbf{55.51} $\pm$ 6.33& 2.20 $\pm$ 0.75 \\
        \multicolumn{ 1 }{|l|}{ SAC } & \emph{11.95} $\pm$ 2.31& 6.20 $\pm$ 0.40& 35.90 $\pm$ 4.77& 2.00 $\pm$ 0.00 \\
        \multicolumn{ 1 }{|l|}{ SAC Auto } & \emph{13.82} $\pm$ 2.36& 6.80 $\pm$ 0.40& \textbf{29.83} $\pm$ 4.62& 1.80 $\pm$ 0.40 \\ \hline
                
        &\multicolumn{1}{c}{OL $\% \uparrow$}
        &\multicolumn{1}{c}{OR $\% \downarrow$}
        &\multicolumn{1}{c}{F1 $\% \uparrow$}
        &\multicolumn{1}{c|}{NC $\downarrow$} \\ \hline   
        
        \multicolumn{ 1 }{|l|}{ VPG } & 30.17 $\pm$ 12.39& 10.05 $\pm$ 5.34& 33.69 $\pm$ 14.15& 84.34 $\pm$ 8.51 \\
        \multicolumn{ 1 }{|l|}{ A2C } & \emph{29.55} $\pm$ 5.64& \emph{7.23} $\pm$ 2.16& \emph{33.70} $\pm$ 6.77& 78.93 $\pm$ 1.47 \\
        \multicolumn{ 1 }{|l|}{ ACKTR } & \emph{52.29} $\pm$ 6.08& \emph{11.23} $\pm$ 3.04& \emph{57.62} $\pm$ 5.03& \textbf{59.44} $\pm$ 13.29 \\
        \multicolumn{ 1 }{|l|}{ TRPO } & 52.00 $\pm$ 7.28& 12.97 $\pm$ 2.72& 56.09 $\pm$ 7.62& 58.70 $\pm$ 10.60 \\
        \multicolumn{ 1 }{|l|}{ PPO } & \emph{26.45} $\pm$ 2.10& 11.64 $\pm$ 3.15& \emph{33.69} $\pm$ 1.86& \textbf{90.09} $\pm$ 13.30 \\
        \multicolumn{ 1 }{|l|}{ DDPG } & 60.71 $\pm$ 4.56& 12.06 $\pm$ 3.18& 70.31 $\pm$ 3.25& \textbf{63.09} $\pm$ 8.36 \\
        \multicolumn{ 1 }{|l|}{ TD3 } & \emph{32.83} $\pm$ 5.72& 10.04 $\pm$ 3.17& 42.25 $\pm$ 7.01& 36.66 $\pm$ 6.17 \\
        \multicolumn{ 1 }{|l|}{ SAC } & \emph{39.03} $\pm$ 7.03& 11.28 $\pm$ 1.91& \emph{50.66} $\pm$ 7.70& \textbf{52.15} $\pm$ 3.23 \\
        \multicolumn{ 1 }{|l|}{ SAC Auto } & \emph{55.35} $\pm$ 2.59& 11.29 $\pm$ 2.72& \emph{65.97} $\pm$ 2.70& \textbf{56.35} $\pm$ 2.65 \\ \hline
    
        \end{tabularx}       
    
        \caption{Tractometer scores from experiment 3 for agents trained on the FiberCup dataset and tested on the Flipped dataset. \emph{Italics} indicate $p < 0.005$ that the mean is lower than the reference results from experiment 1 using a one-sided Welch t-test. \textbf{Bold} indicates $p < 0.005$ that the mean is higher than the reference experiment using the same test.}
        \label{tab:agents_flipped_noretrack}
        \renewcommand{\arraystretch}{1.2}
        \begin{tabularx}{\columnwidth}{>{\hsize=.1\hsize}X|*4{>{\centering\arraybackslash}X}|}
        \cline{2-5}
        
        & \multicolumn{1}{c }{VC $\% \uparrow$}
        & \multicolumn{1}{c}{VB $\uparrow$} 
        & \multicolumn{1}{c }{IC $\% \downarrow$}
        & \multicolumn{1}{c|}{IB $\downarrow$} \\ \hline
        
        \multicolumn{ 1 }{|l|}{ VPG } & 2.71 $\pm$ 1.67& 3.40 $\pm$ 1.74& 4.00 $\pm$ 3.14& 1.40 $\pm$ 0.49 \\
        \multicolumn{ 1 }{|l|}{ A2C } & 4.10 $\pm$ 1.57& 3.60 $\pm$ 0.80& 15.60 $\pm$ 3.97& 1.60 $\pm$ 0.49 \\
        \multicolumn{ 1 }{|l|}{ ACKTR } & \emph{7.43} $\pm$ 1.05& 5.40 $\pm$ 0.80& 21.29 $\pm$ 5.08& 2.20 $\pm$ 0.40 \\
        \multicolumn{ 1 }{|l|}{ TRPO } & 7.79 $\pm$ 1.27& 5.80 $\pm$ 0.40& 24.21 $\pm$ 3.34& 1.40 $\pm$ 0.49 \\
        \multicolumn{ 1 }{|l|}{ PPO } & \emph{3.04} $\pm$ 2.43& 4.00 $\pm$ 1.55& 11.29 $\pm$ 7.97& 1.40 $\pm$ 0.49 \\
        \multicolumn{ 1 }{|l|}{ DDPG } & \emph{10.09} $\pm$ 0.97& 7.00 $\pm$ 0.00& 21.95 $\pm$ 2.16& 2.00 $\pm$ 0.63 \\
        \multicolumn{ 1 }{|l|}{ TD3 } & 8.93 $\pm$ 1.69& 6.20 $\pm$ 0.75& 30.25 $\pm$ 7.92& 1.80 $\pm$ 0.40 \\
        \multicolumn{ 1 }{|l|}{ SAC } & \emph{8.68} $\pm$ 3.51& 7.00 $\pm$ 0.00& \emph{10.36} $\pm$ 3.74& 1.40 $\pm$ 0.49 \\
        \multicolumn{ 1 }{|l|}{ SAC Auto } & \emph{9.46} $\pm$ 3.85& 7.00 $\pm$ 0.00& 10.81 $\pm$ 3.67& 1.40 $\pm$ 0.49 \\ \hline
                    
        &\multicolumn{1}{c}{OL $\% \uparrow$}
        &\multicolumn{1}{c}{OR $\% \downarrow$}
        &\multicolumn{1}{c}{F1 $\% \uparrow$}
        &\multicolumn{1}{c|}{NC $\downarrow$} \\ \hline   
        
        \multicolumn{ 1 }{|l|}{ VPG } & 21.81 $\pm$ 11.50& 5.34 $\pm$ 3.23& 27.24 $\pm$ 14.11& 93.28 $\pm$ 4.12 \\
        \multicolumn{ 1 }{|l|}{ A2C } & 29.23 $\pm$ 6.32& \emph{4.31} $\pm$ 2.62& 34.76 $\pm$ 6.63& \textbf{80.30} $\pm$ 4.93 \\
        \multicolumn{ 1 }{|l|}{ ACKTR } & \emph{47.61} $\pm$ 4.81& \emph{11.35} $\pm$ 0.77& \emph{52.79} $\pm$ 6.36& \textbf{71.28} $\pm$ 4.69 \\
        \multicolumn{ 1 }{|l|}{ TRPO } & 51.95 $\pm$ 7.56& 12.22 $\pm$ 1.06& 57.42 $\pm$ 7.00& 68.00 $\pm$ 3.12 \\
        \multicolumn{ 1 }{|l|}{ PPO } & \emph{22.03} $\pm$ 9.03& 10.52 $\pm$ 4.90& \emph{28.55} $\pm$ 10.97& \textbf{85.67} $\pm$ 10.33 \\
        \multicolumn{ 1 }{|l|}{ DDPG } & 58.38 $\pm$ 4.95& 10.47 $\pm$ 2.22& 67.66 $\pm$ 4.17& \textbf{67.96} $\pm$ 3.05 \\
        \multicolumn{ 1 }{|l|}{ TD3 } & \emph{41.22} $\pm$ 3.04& 13.79 $\pm$ 1.22& 50.64 $\pm$ 4.63& 60.82 $\pm$ 7.92 \\
        \multicolumn{ 1 }{|l|}{ SAC } & \emph{50.32} $\pm$ 0.75& \emph{8.43} $\pm$ 0.62& \emph{63.06} $\pm$ 0.69& \textbf{80.96} $\pm$ 6.92 \\
        \multicolumn{ 1 }{|l|}{ SAC Auto } & \emph{58.53} $\pm$ 3.88& \emph{9.89} $\pm$ 1.06& \emph{69.59} $\pm$ 2.91& \textbf{79.73} $\pm$ 7.22 \\ \hline
        
        \end{tabularx}       
        \end{table}           
          
        \begin{table}[!tbhp] 
        \caption{Tractometer scores from experiment 3 for agents trained and tested on the ISMRM2015 dataset. \emph{Italics} indicate $p < 0.005$ that the mean is lower than the reference results from experiment 1 using a one-sided Welch t-test. \textbf{Bold} indicates $p < 0.005$ that the mean is higher than the reference experiment using the same test.}
        \label{tab:agents_ismrm2015_noretrack}
        \renewcommand{\arraystretch}{1.2}
        \begin{tabularx}{\columnwidth}{>{\hsize=.1\hsize}X|*4{>{\centering\arraybackslash}X}|}
        \cline{2-5}
        
        & \multicolumn{1}{c }{VC $\% \uparrow$}
        & \multicolumn{1}{c}{VB $\uparrow$} 
        & \multicolumn{1}{c }{IC $\% \downarrow$}
        & \multicolumn{1}{c|}{IB $\downarrow$} \\ \hline
        
        \multicolumn{ 1 }{|l|}{ VPG } & 17.49 $\pm$ 1.34& 21.20 $\pm$ 0.98& 62.14 $\pm$ 3.10& 115.00 $\pm$ 13.49 \\
        \multicolumn{ 1 }{|l|}{ A2C } & 21.35 $\pm$ 0.89& 20.40 $\pm$ 1.36& 49.03 $\pm$ 1.04& 68.40 $\pm$ 5.89 \\
        \multicolumn{ 1 }{|l|}{ ACKTR } & \emph{31.72} $\pm$ 5.77& 23.00 $\pm$ 0.00& \textbf{47.81} $\pm$ 3.54& \emph{111.00} $\pm$ 13.91 \\
        \multicolumn{ 1 }{|l|}{ TRPO } & \emph{39.13} $\pm$ 3.37& 23.00 $\pm$ 0.00& \textbf{42.28} $\pm$ 2.01& 136.40 $\pm$ 8.82 \\
        \multicolumn{ 1 }{|l|}{ PPO } & 7.51 $\pm$ 2.31& 17.40 $\pm$ 2.06& \textbf{48.65} $\pm$ 3.49& 76.60 $\pm$ 8.89 \\
        \multicolumn{ 1 }{|l|}{ DDPG } & \emph{37.07} $\pm$ 3.99& 23.00 $\pm$ 0.00& \textbf{47.16} $\pm$ 1.17& 142.80 $\pm$ 10.11 \\
        \multicolumn{ 1 }{|l|}{ TD3 } & \emph{33.58} $\pm$ 2.30& 22.80 $\pm$ 0.40& \textbf{49.93} $\pm$ 1.38& 147.20 $\pm$ 3.37 \\
        \multicolumn{ 1 }{|l|}{ SAC } & \emph{39.98} $\pm$ 3.66& 22.80 $\pm$ 0.40& \textbf{47.91} $\pm$ 1.99& \emph{100.40} $\pm$ 6.59 \\
        \multicolumn{ 1 }{|l|}{ SAC Auto } & \emph{43.00} $\pm$ 4.12& 23.00 $\pm$ 0.00& \textbf{41.20} $\pm$ 1.45& \emph{111.60} $\pm$ 4.41 \\ \hline
        
        &\multicolumn{1}{c}{OL $\% \uparrow$}
        &\multicolumn{1}{c}{OR $\% \downarrow$}
        &\multicolumn{1}{c}{F1 $\% \uparrow$}
        &\multicolumn{1}{c|}{NC $\downarrow$} \\ \hline   
        
        \multicolumn{ 1 }{|l|}{ VPG } & 28.54 $\pm$ 3.11& 27.69 $\pm$ 3.12& 34.85 $\pm$ 3.56& 20.37 $\pm$ 2.96 \\
        \multicolumn{ 1 }{|l|}{ A2C } & 20.90 $\pm$ 0.73& 20.59 $\pm$ 1.00& 27.59 $\pm$ 1.18& 29.61 $\pm$ 0.49 \\
        \multicolumn{ 1 }{|l|}{ ACKTR } & \emph{38.98} $\pm$ 1.31& 27.44 $\pm$ 2.26& \emph{45.93} $\pm$ 1.12& \textbf{20.47} $\pm$ 3.28 \\
        \multicolumn{ 1 }{|l|}{ TRPO } & \emph{44.49} $\pm$ 1.82& \emph{26.22} $\pm$ 0.95& \emph{51.30} $\pm$ 1.15& \textbf{18.59} $\pm$ 3.53 \\
        \multicolumn{ 1 }{|l|}{ PPO } & 13.46 $\pm$ 4.17& 19.42 $\pm$ 2.34& 18.77 $\pm$ 5.02& 43.84 $\pm$ 3.60 \\
        \multicolumn{ 1 }{|l|}{ DDPG } & \emph{40.48} $\pm$ 2.19& \emph{24.31} $\pm$ 0.65& \emph{48.64} $\pm$ 1.59& \textbf{15.77} $\pm$ 3.38 \\
        \multicolumn{ 1 }{|l|}{ TD3 } & \emph{39.07} $\pm$ 1.63& 25.88 $\pm$ 1.37& \emph{46.28} $\pm$ 1.64& \textbf{16.49} $\pm$ 2.07 \\
        \multicolumn{ 1 }{|l|}{ SAC } & \emph{36.28} $\pm$ 1.84& 26.01 $\pm$ 0.84& \emph{44.11} $\pm$ 1.25& \textbf{12.11} $\pm$ 1.87 \\
        \multicolumn{ 1 }{|l|}{ SAC Auto } & \emph{41.97} $\pm$ 1.10& \emph{25.27} $\pm$ 0.78& \emph{49.88} $\pm$ 0.81& \textbf{15.80} $\pm$ 3.01 \\ \hline
    
        \end{tabularx}       
        \end{table}              

 \section{Hyperparameter search for experiment 4 - State variations}\label{app:exp4_hyperparameters}

    Tables~\ref{tab:exp4_raw_hyperparams},~\ref{tab:exp4_0_hyperparams},~\ref{tab:exp4_2_hyperparams} and \ref{tab:exp4_nomask_hyperparams} present the hyperparameters selected for the SAC Auto agent in experiment 4. 

    \begin{table}[!tbhp] 
        \caption{Hyperparameters chosen for SAC Auto in experiment 4 - raw diffusion signal.}
        \label{tab:exp4_raw_hyperparams}
        \renewcommand{\arraystretch}{1.2}
        \begin{tabularx}{\linewidth}{X|*6{>{\arraybackslash}X}}
            Hyperparameter            & FiberCup & ISMRM2015 \\ \hline 
            Learning rate ($\eta$)    & 0.0001 & 0.00005 \\ 
            Discount ($\gamma$)          & 0.5   & 0.5 \\ \hline
        \end{tabularx}       
        \caption{Hyperparameters chosen for SAC Auto in experiment 4 - 0 directions.}
        \label{tab:exp4_0_hyperparams}
        \renewcommand{\arraystretch}{1.2}
        \begin{tabularx}{\linewidth}{X|*6{>{\arraybackslash}X}}
            Hyperparameter            & FiberCup & ISMRM2015 \\ \hline 
            Learning rate ($\eta$)    & 0.00001 & 0.00005 \\ 
            Discount ($\gamma$)          & 0.9 & 0.5 \\ \hline
        \end{tabularx}       
                \caption{Hyperparameters chosen for SAC Auto in experiment 4 - 2 directions.}
        \label{tab:exp4_2_hyperparams}
        \renewcommand{\arraystretch}{1.2}
        \begin{tabularx}{\linewidth}{X|*6{>{\arraybackslash}X}}
            Hyperparameter            & FiberCup & ISMRM2015 \\ \hline 
            Learning rate ($\eta$)    & 0.001 & 0.0001 \\ 
            Discount ($\gamma$)          & 0.5 & 0.5 \\ \hline
        \end{tabularx}       
                \caption{Hyperparameters chosen for SAC Auto in experiment 4 - no WM mask.}
        \label{tab:exp4_nomask_hyperparams}
        \renewcommand{\arraystretch}{1.2}
        \begin{tabularx}{\linewidth}{X|*6{>{\arraybackslash}X}}
            Hyperparameter            & FiberCup & ISMRM2015 \\ \hline 
            Learning rate ($\eta$)    & 0.00001 & 0.00001 \\ 
            Discount ($\gamma$)          & 0.75 & 0.75 \\ \hline
        \end{tabularx}       
    \end{table}   
      
\section{Full results for experiment 4 - State variations}\label{app:exp4_results}

    Tables~\ref{tab:sac_auto_raw_signal_full},~\ref{tab:sac_auto_0dirs_signal_full},~\ref{tab:sac_auto_2dirs_signal_full} and~\ref{tab:sac_auto_nowm_signal_full} present the full Tractometer scores for agents trained during experiment 4 - state variations.
        
    \begin{table}[!thbp] 
        \caption{Tractometer scores for the SAC Auto agents trained using the raw diffusion signal. Arrow indicates training $\rightarrow$ testing datasets. \emph{Italics} indicate $p < 0.005$ that the mean is lower than the reference results from experiment 1 using a one-sided Welch t-test. \textbf{Bold} indicates $p < 0.005$ that the mean is higher than the reference experiment using the same test.}
        \label{tab:sac_auto_raw_signal_full}
        \renewcommand{\arraystretch}{1.1}
        \begin{tabularx}{\columnwidth}{>{\hsize=.1\hsize}X|*4{>{\centering\arraybackslash}X}|}
        \cline{2-5}
        
        & \multicolumn{1}{c }{VC $\% \uparrow$}
        & \multicolumn{1}{c}{VB $\uparrow$} 
        & \multicolumn{1}{c }{IC $\% \downarrow$}
        & \multicolumn{1}{c|}{IB $\downarrow$} \\ \hline
        
        \multicolumn{0}{|l|}{FiberCup $\rightarrow$ FiberCup} & 76.36 $\pm$ 4.35& 7.00 $\pm$ 0.00& 11.57 $\pm$ 3.67& 1.60 $\pm$ 0.66 \\
        \multicolumn{0}{|l|}{FiberCup $\rightarrow$ Flipped} & 76.94 $\pm$ 2.43& 7.00 $\pm$ 0.00& 10.89 $\pm$ 1.89& 1.60 $\pm$ 0.49 \\
        \multicolumn{0}{|l|}{ISMRM2015 $\rightarrow$ ISMRM2015} & 72.27 $\pm$ 0.99& 23.00 $\pm$ 0.00& 24.65 $\pm$ 0.91& 146.00 $\pm$ 8.88 \\ \hline
         
        &\multicolumn{1}{c}{OL $\% \uparrow$}
        &\multicolumn{1}{c}{OR $\% \downarrow$}
        &\multicolumn{1}{c}{F1 $\% \uparrow$}
        &\multicolumn{1}{c|}{NC $\downarrow$} \\ \hline   

        \multicolumn{0}{|l|}{FiberCup $\rightarrow$ FiberCup} & 85.68 $\pm$ 1.98& 17.50 $\pm$ 0.86& 83.58 $\pm$ 0.88& 12.07 $\pm$ 1.18 \\
        \multicolumn{0}{|l|}{FiberCup $\rightarrow$ Flipped} & 87.39 $\pm$ 0.82& 17.80 $\pm$ 1.09& 84.19 $\pm$ 0.57& 12.17 $\pm$ 1.02 \\
        \multicolumn{0}{|l|}{ISMRM2015 $\rightarrow$ ISMRM2015} & \textbf{58.00} $\pm$ 0.32& 28.18 $\pm$ 0.32& 60.01 $\pm$ 0.29& 3.08 $\pm$ 0.16 \\ \hline
        
        \end{tabularx}       
        
        \caption{Tractometer scores for the SAC Auto agents trained using no previous directions in the state. Arrow indicates training $\rightarrow$ testing datasets. \emph{Italics} indicate $p < 0.005$ that the mean is lower than the reference results from experiment 1 using a one-sided Welch t-test. \textbf{Bold} indicates $p < 0.005$ that the mean is higher than the reference experiment using the same test.}
        \label{tab:sac_auto_0dirs_signal_full}
        \renewcommand{\arraystretch}{1.1}
        \begin{tabularx}{\columnwidth}{>{\hsize=.1\hsize}X|*4{>{\centering\arraybackslash}X}|}
        \cline{2-5}
        
        & \multicolumn{1}{c }{VC $\% \uparrow$}
        & \multicolumn{1}{c}{VB $\uparrow$} 
        & \multicolumn{1}{c }{IC $\% \downarrow$}
        & \multicolumn{1}{c|}{IB $\downarrow$} \\ \hline
        
        \multicolumn{0}{|l|}{FiberCup $\rightarrow$ FiberCup} & \emph{2.76} $\pm$ 1.01& \emph{1.60} $\pm$ 0.66& 7.27 $\pm$ 2.44& \emph{1.10} $\pm$ 0.30 \\
        \multicolumn{0}{|l|}{FiberCup $\rightarrow$ Flipped} & \emph{2.06} $\pm$ 0.48& \emph{1.40} $\pm$ 0.49& 8.20 $\pm$ 2.77& \emph{1.00} $\pm$ 0.00 \\
        \multicolumn{0}{|l|}{ISMRM2015 $\rightarrow$ ISMRM2015} & \emph{14.57} $\pm$ 2.42& 17.20 $\pm$ 2.56& \textbf{43.31} $\pm$ 0.98& \emph{43.00} $\pm$ 3.74 \\ \hline
         
        &\multicolumn{1}{c}{OL $\% \uparrow$}
        &\multicolumn{1}{c}{OR $\% \downarrow$}
        &\multicolumn{1}{c}{F1 $\% \uparrow$}
        &\multicolumn{1}{c|}{NC $\downarrow$} \\ \hline   

        \multicolumn{0}{|l|}{FiberCup $\rightarrow$ FiberCup} & \emph{11.58} $\pm$ 3.92& \emph{1.05} $\pm$ 1.44& \emph{14.91} $\pm$ 5.17& \textbf{89.97} $\pm$ 2.37 \\
        \multicolumn{0}{|l|}{FiberCup $\rightarrow$ Flipped} & \emph{10.68} $\pm$ 3.53& \emph{0.31} $\pm$ 0.15& \emph{13.83} $\pm$ 4.66& \textbf{89.75} $\pm$ 2.53 \\
        \multicolumn{0}{|l|}{ISMRM2015 $\rightarrow$ ISMRM2015} & \emph{12.87} $\pm$ 2.57& \emph{15.11} $\pm$ 1.45& \emph{19.01} $\pm$ 3.36& \textbf{42.12} $\pm$ 2.58 \\ \hline
        
        \end{tabularx}       
         
        \caption{Tractometer scores for the SAC Auto agents trained using two previous directions in the state. Arrow indicates training $\rightarrow$ testing datasets. \emph{Italics} indicate $p < 0.005$ that the mean is lower than the reference results from experiment 1 using a one-sided Welch t-test. \textbf{Bold} indicates $p < 0.005$ that the mean is higher than the reference experiment using the same test.}
        \label{tab:sac_auto_2dirs_signal_full}
        \renewcommand{\arraystretch}{1.1}
        \begin{tabularx}{\columnwidth}{>{\hsize=.1\hsize}X|*4{>{\centering\arraybackslash}X}|}
        \cline{2-5}
        
        & \multicolumn{1}{c }{VC $\% \uparrow$}
        & \multicolumn{1}{c}{VB $\uparrow$} 
        & \multicolumn{1}{c }{IC $\% \downarrow$}
        & \multicolumn{1}{c|}{IB $\downarrow$} \\ \hline
        
        \multicolumn{0}{|l|}{FiberCup $\rightarrow$ FiberCup} & 76.71 $\pm$ 1.94& 7.00 $\pm$ 0.00& 10.99 $\pm$ 2.01& 1.40 $\pm$ 0.49 \\
        \multicolumn{0}{|l|}{FiberCup $\rightarrow$ Flipped} & 78.10 $\pm$ 1.72& 7.00 $\pm$ 0.00& 9.41 $\pm$ 1.08& 1.80 $\pm$ 0.40 \\
        \multicolumn{0}{|l|}{ISMRM2015 $\rightarrow$ ISMRM2015} & 75.10 $\pm$ 1.06& 23.00 $\pm$ 0.00& 22.17 $\pm$ 1.10& 151.80 $\pm$ 5.27 \\ \hline
         
        &\multicolumn{1}{c}{OL $\% \uparrow$}
        &\multicolumn{1}{c}{OR $\% \downarrow$}
        &\multicolumn{1}{c}{F1 $\% \uparrow$}
        &\multicolumn{1}{c|}{NC $\downarrow$} \\ \hline   

        \multicolumn{0}{|l|}{FiberCup $\rightarrow$ FiberCup} & 85.35 $\pm$ 2.70& 17.38 $\pm$ 1.18& 83.53 $\pm$ 1.23& 12.29 $\pm$ 0.83 \\
        \multicolumn{0}{|l|}{FiberCup $\rightarrow$ Flipped} & 87.66 $\pm$ 1.23& 17.93 $\pm$ 1.22& 84.35 $\pm$ 0.59& 12.50 $\pm$ 0.86 \\
        \multicolumn{0}{|l|}{ISMRM2015 $\rightarrow$ ISMRM2015} & 57.10 $\pm$ 0.37& 27.56 $\pm$ 0.11& 59.82 $\pm$ 0.27& 2.73 $\pm$ 0.06 \\ \hline
        
        \end{tabularx}      
        
         \caption{Tractometer scores for the SAC Auto agents trained without the WM mask as part of the state. Arrow indicates training $\rightarrow$ testing datasets. \emph{Italics} indicate $p < 0.005$ that the mean is lower than the reference results from experiment 1 using a one-sided Welch t-test. \textbf{Bold} indicates $p < 0.005$ that the mean is higher than the reference experiment using the same test.}
        \label{tab:sac_auto_nowm_signal_full}
        \renewcommand{\arraystretch}{1.1}
        \begin{tabularx}{\columnwidth}{>{\hsize=.1\hsize}X|*4{>{\centering\arraybackslash}X}|}
        \cline{2-5}
        
        & \multicolumn{1}{c }{VC $\% \uparrow$}
        & \multicolumn{1}{c}{VB $\uparrow$} 
        & \multicolumn{1}{c }{IC $\% \downarrow$}
        & \multicolumn{1}{c|}{IB $\downarrow$} \\ \hline
        
        \multicolumn{0}{|l|}{FiberCup $\rightarrow$ FiberCup} & 78.60 $\pm$ 5.04& 7.00 $\pm$ 0.00& 14.11 $\pm$ 4.99& 1.80 $\pm$ 0.60 \\
        \multicolumn{0}{|l|}{FiberCup $\rightarrow$ Flipped} & 75.75 $\pm$ 2.49& 7.00 $\pm$ 0.00& \textbf{16.54} $\pm$ 2.41& 1.80 $\pm$ 0.75 \\
        \multicolumn{0}{|l|}{ISMRM2015 $\rightarrow$ ISMRM2015} & 74.09 $\pm$ 1.22& 23.00 $\pm$ 0.00& 23.73 $\pm$ 1.21& \emph{121.00} $\pm$ 7.56 \\ \hline
         
        &\multicolumn{1}{c}{OL $\% \uparrow$}
        &\multicolumn{1}{c}{OR $\% \downarrow$}
        &\multicolumn{1}{c}{F1 $\% \uparrow$}
        &\multicolumn{1}{c|}{NC $\downarrow$} \\ \hline   

        \multicolumn{0}{|l|}{FiberCup $\rightarrow$ FiberCup} & 82.86 $\pm$ 1.72& 14.78 $\pm$ 0.71& 83.37 $\pm$ 1.11& \emph{7.29} $\pm$ 0.70 \\
        \multicolumn{0}{|l|}{FiberCup $\rightarrow$ Flipped} & \emph{83.47} $\pm$ 1.06& 14.73 $\pm$ 0.90& 83.57 $\pm$ 1.17& \emph{7.71} $\pm$ 0.55 \\
        \multicolumn{0}{|l|}{ISMRM2015 $\rightarrow$ ISMRM2015} & \emph{54.09} $\pm$ 0.78& \emph{24.99} $\pm$ 0.36& 58.48 $\pm$ 0.68& \emph{2.18} $\pm$ 0.12 \\ \hline
        
        \end{tabularx}              
    \end{table}

 \section{Hyperparameter search for experiment 5 - Reward variations}\label{app:exp5_hyperparameters}

    Tables~\ref{tab:exp5_length_hyperparams} and \ref{tab:exp5_gm_hyperparams} present the hyperparameters selected for the SAC Auto agent in experiment 5. 

    \begin{table}[!tbhp] 
        \caption{Hyperparameters chosen for SAC Auto in experiment 5 - length bonus.}
        \label{tab:exp5_length_hyperparams}
        \renewcommand{\arraystretch}{1.2}

        \begin{tabularx}{\linewidth}{X|*{11}{>{\arraybackslash}X}}
        \multicolumn{1}{l|}{} & \multicolumn{5}{c}{FiberCup} & \multicolumn{5}{c}{ISMRM2015}\\ \cline{2-11}
        \multicolumn{1}{l|}{Bonus coefficient} & 0.01 & 0.1 & 0.5 & 1 & 5 & 0.01 & 0.1 & 0.5 & 1 & 5   \\
        \hline
        \multicolumn{1}{l|}{ Learning rate ($\eta$)} & 0.00001 & 0.0005 & 0.00005 & 0.00001 & 0.00001 & 0.001 & 0.00005 & 0.00005 & 0.0001 & 0.0001\\ 
        \multicolumn{1}{l|}{ Discount ($\gamma$)}    & 0.75    & 0.5    & 0.5     & 0.5     & 0.75    & 0.5   & 0.5     & 0.5     & 0.5    & 0.5 \\ \hline
        \end{tabularx}       
        
        \caption{Hyperparameters chosen for SAC Auto in experiment 5 - GM bonus.}
        \label{tab:exp5_gm_hyperparams}
        \renewcommand{\arraystretch}{1.2}

        \begin{tabularx}{\linewidth}{X|*{7}{>{\arraybackslash}X}}
        \multicolumn{1}{l|}{} & \multicolumn{3}{c}{FiberCup} & \multicolumn{3}{c}{ISMRM2015}\\ \cline{2-7}
        \multicolumn{1}{l|}{Bonus coefficient} & 1 & 10 & 100 & 1 & 10 & 100   \\
        \hline
        \multicolumn{1}{l|}{ Learning rate ($\eta$)} & 0.00001 & 0.00001 & 0.00005 & 0.0001 & 0.0001 & 0.0001 \\ 
        \multicolumn{1}{l|}{ Discount ($\gamma$)}    & 0.5     & 0.9     & 0.95    & 0.5    & 0.75   & 0.95 \\ \hline
        \end{tabularx}       
    \end{table}   
          
\section{Full results for experiment 5 - Reward variations}\label{app:exp5_results}

    Tables~\ref{tab:sac_auto_length_0.01_full},~\ref{tab:sac_auto_length_0.1_full},~\ref{tab:sac_auto_length_0.5_full},~\ref{tab:sac_auto_length_1_full} and~\ref{tab:sac_auto_length_5_full} present the full Tractometer scores for agents trained during experiment 5 with a length bonus added to the reward function. Tables~\ref{tab:sac_auto_target_1_full},~\ref{tab:sac_auto_target_10_full} and~\ref{tab:sac_auto_target_100_full} present the full Tractometer scores for agents trained during experiment 5 with a GM bonus added to the reward function.
        
    \begin{table}[!thbp] 
        \caption{Tractometer scores for the SAC Auto agents trained with a length bonus of 0.01. Arrow indicates training $\rightarrow$ testing datasets. \emph{Italics} indicate $p < 0.005$ that the mean is lower than the reference results from experiment 1 using a one-sided Welch t-test. \textbf{Bold} indicates $p < 0.005$ that the mean is higher than the reference experiment using the same test.}
        \label{tab:sac_auto_length_0.01_full}
        \renewcommand{\arraystretch}{1.1}
        \begin{tabularx}{\columnwidth}{>{\hsize=.1\hsize}X|*4{>{\centering\arraybackslash}X}|}
        \cline{2-5}
        
        & \multicolumn{1}{c }{VC $\% \uparrow$}
        & \multicolumn{1}{c}{VB $\uparrow$} 
        & \multicolumn{1}{c }{IC $\% \downarrow$}
        & \multicolumn{1}{c|}{IB $\downarrow$} \\ \hline
        
        \multicolumn{0}{|l|}{FiberCup $\rightarrow$ FiberCup} & 82.71 $\pm$ 4.07& 7.00 $\pm$ 0.00& 10.76 $\pm$ 4.03& 2.10 $\pm$ 0.70 \\
        \multicolumn{0}{|l|}{FiberCup $\rightarrow$ Flipped} & 79.09 $\pm$ 1.68& 7.00 $\pm$ 0.00& \textbf{14.36} $\pm$ 1.77& 2.00 $\pm$ 0.63 \\
        \multicolumn{0}{|l|}{ISMRM2015 $\rightarrow$ ISMRM2015} & 72.33 $\pm$ 0.69& 23.00 $\pm$ 0.00& 25.02 $\pm$ 0.75& 150.00 $\pm$ 3.85 \\ \hline
         
        &\multicolumn{1}{c}{OL $\% \uparrow$}
        &\multicolumn{1}{c}{OR $\% \downarrow$}
        &\multicolumn{1}{c}{F1 $\% \uparrow$}
        &\multicolumn{1}{c|}{NC $\downarrow$} \\ \hline   

        \multicolumn{0}{|l|}{FiberCup $\rightarrow$ FiberCup} & 84.54 $\pm$ 1.33& 14.55 $\pm$ 1.13& 84.35 $\pm$ 0.73& \emph{6.52} $\pm$ 0.36 \\
        \multicolumn{0}{|l|}{FiberCup $\rightarrow$ Flipped} & \emph{83.60} $\pm$ 1.20& 14.12 $\pm$ 1.23& 83.98 $\pm$ 0.79& \emph{6.55} $\pm$ 0.30 \\
        \multicolumn{0}{|l|}{ISMRM2015 $\rightarrow$ ISMRM2015} & 57.39 $\pm$ 0.29& 27.79 $\pm$ 0.31& 59.82 $\pm$ 0.17& 2.65 $\pm$ 0.23 \\ \hline
        
        \end{tabularx}       
        
        \caption{Tractometer scores for the SAC Auto agents trained with a length bonus of 0.1. Arrow indicates training $\rightarrow$ testing datasets. \emph{Italics} indicate $p < 0.005$ that the mean is lower than the reference results from experiment 1 using a one-sided Welch t-test. \textbf{Bold} indicates $p < 0.005$ that the mean is higher than the reference experiment using the same test.}
        \label{tab:sac_auto_length_0.1_full}
        \renewcommand{\arraystretch}{1.1}
        \begin{tabularx}{\columnwidth}{>{\hsize=.1\hsize}X|*4{>{\centering\arraybackslash}X}|}
        \cline{2-5}
        
        & \multicolumn{1}{c }{VC $\% \uparrow$}
        & \multicolumn{1}{c}{VB $\uparrow$} 
        & \multicolumn{1}{c }{IC $\% \downarrow$}
        & \multicolumn{1}{c|}{IB $\downarrow$} \\ \hline
        
        \multicolumn{0}{|l|}{FiberCup $\rightarrow$ FiberCup} & 76.03 $\pm$ 2.78& 7.00 $\pm$ 0.00& 12.11 $\pm$ 2.78& 1.80 $\pm$ 0.60 \\
        \multicolumn{0}{|l|}{FiberCup $\rightarrow$ Flipped} & 77.49 $\pm$ 1.90& 7.00 $\pm$ 0.00& 10.33 $\pm$ 1.62& 2.00 $\pm$ 0.00 \\
        \multicolumn{0}{|l|}{ISMRM2015 $\rightarrow$ ISMRM2015} & 74.77 $\pm$ 0.49& 23.00 $\pm$ 0.00& 22.37 $\pm$ 0.48& 156.40 $\pm$ 6.12 \\ \hline
         
        &\multicolumn{1}{c}{OL $\% \uparrow$}
        &\multicolumn{1}{c}{OR $\% \downarrow$}
        &\multicolumn{1}{c}{F1 $\% \uparrow$}
        &\multicolumn{1}{c|}{NC $\downarrow$} \\ \hline   

        \multicolumn{0}{|l|}{FiberCup $\rightarrow$ FiberCup} & 85.54 $\pm$ 2.13& 16.67 $\pm$ 1.03& 84.03 $\pm$ 1.01& 11.87 $\pm$ 0.89 \\
        \multicolumn{0}{|l|}{FiberCup $\rightarrow$ Flipped} & 87.38 $\pm$ 0.56& 17.09 $\pm$ 1.26& 84.64 $\pm$ 0.66& 12.18 $\pm$ 0.97 \\
        \multicolumn{0}{|l|}{ISMRM2015 $\rightarrow$ ISMRM2015} & 57.30 $\pm$ 0.24& 27.88 $\pm$ 0.18& 59.73 $\pm$ 0.22& 2.86 $\pm$ 0.11 \\ \hline
        
        \end{tabularx}       
         
        \caption{Tractometer scores for the SAC Auto agents trained with a length bonus of 0.5. Arrow indicates training $\rightarrow$ testing datasets. \emph{Italics} indicate $p < 0.005$ that the mean is lower than the reference results from experiment 1 using a one-sided Welch t-test. \textbf{Bold} indicates $p < 0.005$ that the mean is higher than the reference experiment using the same test.}
        \label{tab:sac_auto_length_0.5_full}
        \renewcommand{\arraystretch}{1.1}
        \begin{tabularx}{\columnwidth}{>{\hsize=.1\hsize}X|*4{>{\centering\arraybackslash}X}|}
        \cline{2-5}
        
        & \multicolumn{1}{c }{VC $\% \uparrow$}
        & \multicolumn{1}{c}{VB $\uparrow$} 
        & \multicolumn{1}{c }{IC $\% \downarrow$}
        & \multicolumn{1}{c|}{IB $\downarrow$} \\ \hline
        
        \multicolumn{0}{|l|}{FiberCup $\rightarrow$ FiberCup} & 74.88 $\pm$ 5.88& 7.00 $\pm$ 0.00& 14.23 $\pm$ 5.12& 2.20 $\pm$ 0.40 \\
        \multicolumn{0}{|l|}{FiberCup $\rightarrow$ Flipped} & 78.90 $\pm$ 3.90& 7.00 $\pm$ 0.00& 10.30 $\pm$ 1.69& 2.00 $\pm$ 0.00 \\
        \multicolumn{0}{|l|}{ISMRM2015 $\rightarrow$ ISMRM2015} & 71.95 $\pm$ 1.69& 23.00 $\pm$ 0.00& 25.40 $\pm$ 1.58& 159.40 $\pm$ 9.97 \\ \hline
         
        &\multicolumn{1}{c}{OL $\% \uparrow$}
        &\multicolumn{1}{c}{OR $\% \downarrow$}
        &\multicolumn{1}{c}{F1 $\% \uparrow$}
        &\multicolumn{1}{c|}{NC $\downarrow$} \\ \hline   

        \multicolumn{0}{|l|}{FiberCup $\rightarrow$ FiberCup} & 84.51 $\pm$ 2.59& 16.09 $\pm$ 0.68& 83.85 $\pm$ 1.41& 10.88 $\pm$ 1.89 \\
        \multicolumn{0}{|l|}{FiberCup $\rightarrow$ Flipped} & 86.81 $\pm$ 0.64& 16.01 $\pm$ 0.85& 85.02 $\pm$ 0.67& 10.80 $\pm$ 2.35 \\
        \multicolumn{0}{|l|}{ISMRM2015 $\rightarrow$ ISMRM2015} & 57.20 $\pm$ 0.61& 28.08 $\pm$ 0.49& 59.31 $\pm$ 0.47& 2.65 $\pm$ 0.15 \\ \hline
        
        \end{tabularx}      
        
         \caption{Tractometer scores for the SAC Auto agents trained with a length bonus of 1. Arrow indicates training $\rightarrow$ testing datasets. \emph{Italics} indicate $p < 0.005$ that the mean is lower than the reference results from experiment 1 using a one-sided Welch t-test. \textbf{Bold} indicates $p < 0.005$ that the mean is higher than the reference experiment using the same test.}
        \label{tab:sac_auto_length_1_full}
        \renewcommand{\arraystretch}{1.1}
        \begin{tabularx}{\columnwidth}{>{\hsize=.1\hsize}X|*4{>{\centering\arraybackslash}X}|}
        \cline{2-5}
        
        & \multicolumn{1}{c }{VC $\% \uparrow$}
        & \multicolumn{1}{c}{VB $\uparrow$} 
        & \multicolumn{1}{c }{IC $\% \downarrow$}
        & \multicolumn{1}{c|}{IB $\downarrow$} \\ \hline
        
        \multicolumn{0}{|l|}{FiberCup $\rightarrow$ FiberCup} & \emph{68.13} $\pm$ 3.91& 7.00 $\pm$ 0.00& 17.21 $\pm$ 5.01& 1.70 $\pm$ 0.46 \\
        \multicolumn{0}{|l|}{FiberCup $\rightarrow$ Flipped} & \emph{65.32} $\pm$ 2.74& 7.00 $\pm$ 0.00& \textbf{21.01} $\pm$ 2.62& 1.80 $\pm$ 0.40 \\
        \multicolumn{0}{|l|}{ISMRM2015 $\rightarrow$ ISMRM2015} & 64.76 $\pm$ 4.24& 23.00 $\pm$ 0.00& 32.69 $\pm$ 4.27& 157.60 $\pm$ 3.72 \\ \hline
         
        &\multicolumn{1}{c}{OL $\% \uparrow$}
        &\multicolumn{1}{c}{OR $\% \downarrow$}
        &\multicolumn{1}{c}{F1 $\% \uparrow$}
        &\multicolumn{1}{c|}{NC $\downarrow$} \\ \hline   

        \multicolumn{0}{|l|}{FiberCup $\rightarrow$ FiberCup} & 83.19 $\pm$ 1.40& \textbf{19.96} $\pm$ 1.09& \emph{80.83} $\pm$ 1.09& \textbf{14.65} $\pm$ 1.85 \\
        \multicolumn{0}{|l|}{FiberCup $\rightarrow$ Flipped} & \emph{83.24} $\pm$ 1.21& 19.09 $\pm$ 0.69& \emph{81.34} $\pm$ 0.61& 13.66 $\pm$ 1.25 \\
        \multicolumn{0}{|l|}{ISMRM2015 $\rightarrow$ ISMRM2015} & 53.52 $\pm$ 2.87& 27.15 $\pm$ 0.95& 56.03 $\pm$ 2.56& 2.55 $\pm$ 0.22 \\ \hline
        
        \end{tabularx}                       
    \end{table}                    
    \begin{table}[!thbp]      
         \caption{Tractometer scores for the SAC Auto agents trained with a length bonus of 5. Arrow indicates training $\rightarrow$ testing datasets. \emph{Italics} indicate $p < 0.005$ that the mean is lower than the reference results from experiment 1 using a one-sided Welch t-test. \textbf{Bold} indicates $p < 0.005$ that the mean is higher than the reference experiment using the same test.}
        \label{tab:sac_auto_length_5_full}
        \renewcommand{\arraystretch}{1.1}
        \begin{tabularx}{\columnwidth}{>{\hsize=.1\hsize}X|*4{>{\centering\arraybackslash}X}|}
        \cline{2-5}
        
        & \multicolumn{1}{c }{VC $\% \uparrow$}
        & \multicolumn{1}{c}{VB $\uparrow$} 
        & \multicolumn{1}{c }{IC $\% \downarrow$}
        & \multicolumn{1}{c|}{IB $\downarrow$} \\ \hline
        
        \multicolumn{0}{|l|}{FiberCup $\rightarrow$ FiberCup} & 67.54 $\pm$ 8.44& 7.00 $\pm$ 0.00& \textbf{20.68} $\pm$ 7.11& 1.80 $\pm$ 0.40 \\
        \multicolumn{0}{|l|}{FiberCup $\rightarrow$ Flipped} & \emph{61.92} $\pm$ 5.31& 7.00 $\pm$ 0.00& \textbf{25.51} $\pm$ 3.55& 1.60 $\pm$ 0.49 \\
        \multicolumn{0}{|l|}{ISMRM2015 $\rightarrow$ ISMRM2015} & \emph{50.59} $\pm$ 5.24& 22.80 $\pm$ 0.40& \textbf{44.92} $\pm$ 4.46& 140.80 $\pm$ 7.36 \\ \hline
         
        &\multicolumn{1}{c}{OL $\% \uparrow$}
        &\multicolumn{1}{c}{OR $\% \downarrow$}
        &\multicolumn{1}{c}{F1 $\% \uparrow$}
        &\multicolumn{1}{c|}{NC $\downarrow$} \\ \hline   

        \multicolumn{0}{|l|}{FiberCup $\rightarrow$ FiberCup} & \emph{79.77} $\pm$ 3.55& 16.71 $\pm$ 1.00& \emph{80.16} $\pm$ 2.35& 11.78 $\pm$ 3.06 \\
        \multicolumn{0}{|l|}{FiberCup $\rightarrow$ Flipped} & \emph{78.17} $\pm$ 2.13& 16.28 $\pm$ 0.99& \emph{79.40} $\pm$ 1.73& 12.57 $\pm$ 3.31 \\
        \multicolumn{0}{|l|}{ISMRM2015 $\rightarrow$ ISMRM2015} & 47.26 $\pm$ 4.38& 27.79 $\pm$ 1.29& 51.13 $\pm$ 3.53& 4.49 $\pm$ 0.90 \\ \hline
        
        \end{tabularx}                     
        
         \caption{Tractometer scores for the SAC Auto agents trained with a GM bonus of 1. Arrow indicates training $\rightarrow$ testing datasets. \emph{Italics} indicate $p < 0.005$ that the mean is lower than the reference results from experiment 1 using a one-sided Welch t-test. \textbf{Bold} indicates $p < 0.005$ that the mean is higher than the reference experiment using the same test.}
        \label{tab:sac_auto_target_1_full}
        \renewcommand{\arraystretch}{1.1}
        \begin{tabularx}{\columnwidth}{>{\hsize=.1\hsize}X|*4{>{\centering\arraybackslash}X}|}
        \cline{2-5}
        
        & \multicolumn{1}{c }{VC $\% \uparrow$}
        & \multicolumn{1}{c}{VB $\uparrow$} 
        & \multicolumn{1}{c }{IC $\% \downarrow$}
        & \multicolumn{1}{c|}{IB $\downarrow$} \\ \hline
        
        \multicolumn{0}{|l|}{FiberCup $\rightarrow$ FiberCup} & 78.39 $\pm$ 1.51& 7.00 $\pm$ 0.00& 9.57 $\pm$ 1.59& 1.60 $\pm$ 0.49 \\
        \multicolumn{0}{|l|}{FiberCup $\rightarrow$ Flipped} & 78.06 $\pm$ 1.16& 7.00 $\pm$ 0.00& 9.67 $\pm$ 1.48& 1.80 $\pm$ 0.40 \\
        \multicolumn{0}{|l|}{ISMRM2015 $\rightarrow$ ISMRM2015} & 75.16 $\pm$ 0.67& 23.00 $\pm$ 0.00& 22.15 $\pm$ 0.62& 149.40 $\pm$ 1.20 \\ \hline
         
        &\multicolumn{1}{c}{OL $\% \uparrow$}
        &\multicolumn{1}{c}{OR $\% \downarrow$}
        &\multicolumn{1}{c}{F1 $\% \uparrow$}
        &\multicolumn{1}{c|}{NC $\downarrow$} \\ \hline   

        \multicolumn{0}{|l|}{FiberCup $\rightarrow$ FiberCup} & 86.53 $\pm$ 1.93& 17.40 $\pm$ 0.72& 84.14 $\pm$ 0.75& 12.03 $\pm$ 0.89 \\
        \multicolumn{0}{|l|}{FiberCup $\rightarrow$ Flipped} & 87.77 $\pm$ 0.81& 17.26 $\pm$ 0.31& 84.73 $\pm$ 0.33& 12.26 $\pm$ 1.19 \\
        \multicolumn{0}{|l|}{ISMRM2015 $\rightarrow$ ISMRM2015} & 56.82 $\pm$ 0.34& 27.04 $\pm$ 0.14& 59.88 $\pm$ 0.21& 2.69 $\pm$ 0.14 \\ \hline
        
        \end{tabularx}                            
         
         \caption{Tractometer scores for the SAC Auto agents trained with a GM bonus of 10. Arrow indicates training $\rightarrow$ testing datasets. \emph{Italics} indicate $p < 0.005$ that the mean is lower than the reference results from experiment 1 using a one-sided Welch t-test. \textbf{Bold} indicates $p < 0.005$ that the mean is higher than the reference experiment using the same test.}
        \label{tab:sac_auto_target_10_full}
        \renewcommand{\arraystretch}{1.1}
        \begin{tabularx}{\columnwidth}{>{\hsize=.1\hsize}X|*4{>{\centering\arraybackslash}X}|}
        \cline{2-5}
        
        & \multicolumn{1}{c }{VC $\% \uparrow$}
        & \multicolumn{1}{c}{VB $\uparrow$} 
        & \multicolumn{1}{c }{IC $\% \downarrow$}
        & \multicolumn{1}{c|}{IB $\downarrow$} \\ \hline
        
        \multicolumn{0}{|l|}{FiberCup $\rightarrow$ FiberCup} & 77.49 $\pm$ 5.86& 7.00 $\pm$ 0.00& 12.86 $\pm$ 5.52& 2.40 $\pm$ 0.66 \\
        \multicolumn{0}{|l|}{FiberCup $\rightarrow$ Flipped} & 75.66 $\pm$ 5.64& 7.00 $\pm$ 0.00& 14.38 $\pm$ 5.15& 2.20 $\pm$ 0.40 \\
        \multicolumn{0}{|l|}{ISMRM2015 $\rightarrow$ ISMRM2015} & \textbf{78.45} $\pm$ 0.51& 23.00 $\pm$ 0.00& \emph{20.07} $\pm$ 0.53& 144.40 $\pm$ 7.09 \\ \hline
         
        &\multicolumn{1}{c}{OL $\% \uparrow$}
        &\multicolumn{1}{c}{OR $\% \downarrow$}
        &\multicolumn{1}{c}{F1 $\% \uparrow$}
        &\multicolumn{1}{c|}{NC $\downarrow$} \\ \hline   

        \multicolumn{0}{|l|}{FiberCup $\rightarrow$ FiberCup} & \emph{78.48} $\pm$ 3.50& \emph{12.62} $\pm$ 1.77& 80.92 $\pm$ 3.18& \emph{9.65} $\pm$ 0.80 \\
        \multicolumn{0}{|l|}{FiberCup $\rightarrow$ Flipped} & 79.02 $\pm$ 4.55& \emph{12.77} $\pm$ 1.67& 80.97 $\pm$ 4.09& \emph{9.95} $\pm$ 0.90 \\
        \multicolumn{0}{|l|}{ISMRM2015 $\rightarrow$ ISMRM2015} & \emph{55.16} $\pm$ 0.51& \emph{23.97} $\pm$ 0.26& 60.02 $\pm$ 0.46& \emph{1.48} $\pm$ 0.08 \\ \hline
        
        \end{tabularx}                                   
          
         \caption{Tractometer scores for the SAC Auto agents trained with a GM bonus of 100. Arrow indicates training $\rightarrow$ testing datasets. \emph{Italics} indicate $p < 0.005$ that the mean is lower than the reference results from experiment 1 using a one-sided Welch t-test. \textbf{Bold} indicates $p < 0.005$ that the mean is higher than the reference experiment using the same test.}
        \label{tab:sac_auto_target_100_full}
        \renewcommand{\arraystretch}{1.1}
        \begin{tabularx}{\columnwidth}{>{\hsize=.1\hsize}X|*4{>{\centering\arraybackslash}X}|}
        \cline{2-5}
        
        & \multicolumn{1}{c }{VC $\% \uparrow$}
        & \multicolumn{1}{c}{VB $\uparrow$} 
        & \multicolumn{1}{c }{IC $\% \downarrow$}
        & \multicolumn{1}{c|}{IB $\downarrow$} \\ \hline
        
        \multicolumn{0}{|l|}{FiberCup $\rightarrow$ FiberCup} & \emph{43.48} $\pm$ 12.28& 6.40 $\pm$ 0.66& \textbf{28.71} $\pm$ 5.58& 2.30 $\pm$ 0.64 \\
        \multicolumn{0}{|l|}{FiberCup $\rightarrow$ Flipped} & \emph{37.06} $\pm$ 10.73& 6.80 $\pm$ 0.40& \textbf{28.66} $\pm$ 5.97& 2.40 $\pm$ 0.49 \\
        \multicolumn{0}{|l|}{ISMRM2015 $\rightarrow$ ISMRM2015} & \emph{63.67} $\pm$ 3.62& 22.80 $\pm$ 0.40& \textbf{33.64} $\pm$ 3.20& 131.40 $\pm$ 10.59 \\ \hline
         
        &\multicolumn{1}{c}{OL $\% \uparrow$}
        &\multicolumn{1}{c}{OR $\% \downarrow$}
        &\multicolumn{1}{c}{F1 $\% \uparrow$}
        &\multicolumn{1}{c|}{NC $\downarrow$} \\ \hline   

        \multicolumn{0}{|l|}{FiberCup $\rightarrow$ FiberCup} & \emph{60.00} $\pm$ 6.16& \emph{9.10} $\pm$ 2.56& \emph{66.53} $\pm$ 5.03& \textbf{27.82} $\pm$ 12.58 \\
        \multicolumn{0}{|l|}{FiberCup $\rightarrow$ Flipped} & \emph{59.00} $\pm$ 6.63& \emph{10.34} $\pm$ 2.56& \emph{66.99} $\pm$ 5.20& 34.29 $\pm$ 13.27 \\
        \multicolumn{0}{|l|}{ISMRM2015 $\rightarrow$ ISMRM2015} & \emph{41.50} $\pm$ 2.13& 23.95 $\pm$ 1.95& \emph{47.71} $\pm$ 1.87& 2.69 $\pm$ 0.49 \\ \hline
        
        \end{tabularx}                                          
    \end{table}                                 